**TRABAJO FIN DE MASTER / TRABAJO FIN DE GRADO**

UNIVERSIDAD CATÓLICA SAN ANTONIO

UCAM

ESCUELA UNIVERSITARIA POLITÉCNICA

Departamento de Ingeniería Informática

GRADO EN INGENIERÍA INFORMÁTICA

# "ANÁLISIS E IMPLEMENTACIÓN DE ALGORITMOS EVOLUTIVOS PARA LA OPTIMIZACIÓN DE SIMULACIONES EN INGENIERÍA CIVIL."

**Alumno: José Alberto García Gutiérrez**

**Directores: Dr. D. José María Cecilia Canales, Dr. D. Alejandro Mateo Hernández Díaz.**

Esta página ha sido dejada intencionalmente en blanco.

**TRABAJO FIN DE MASTER / TRABAJO FIN DE GRADO**

UNIVERSIDAD CATÓLICA SAN ANTONIO

UCAM

ESCUELA UNIVERSITARIA POLITÉCNICA

Departamento de Ingeniería Informática

GRADO EN INGENIERÍA INFORMÁTICA

# "ANÁLISIS E IMPLEMENTACIÓN DE ALGORITMOS EVOLUTIVOS PARA LA OPTIMIZACIÓN DE SIMULACIONES EN INGENIERÍA CIVIL."

**Alumno: José Alberto García Gutiérrez**

**Directores: Dr. D. José María Cecilia Canales, Dr. D. Alejandro Mateo Hernández Díaz.**

# Agradecimientos

Este trabajo ha sido realizado en colaboración con el departamento de Ingeniería Civil de la Universidad Católica de Murcia, a quienes agradezco la ayuda que me han prestado y en especial a mis tutores, Alejandro y José María por su paciencia en las explicaciones, así como por haberme confiado este trabajo a pesar de que no siempre he podido dedicarle todo el tiempo que me hubiese gustado. En último lugar, me gustaría agradecer también a la unidad departamental de Física Básica de la Universidad de La Laguna de Tenerife el permitirme hacer uso de los recursos de cálculo de dicha universidad. Parte de los textos e ilustraciones referidas al análisis y el desarrollo matemático de los problemas tratados fueron aportados por sus autores para la elaboración del presente trabajo siendo por tanto suyos los derechos de cesión o reproducción de esos apartados.

# Resumen

En este trabajo, estudiamos la posible aplicación de algoritmos evolutivos, concretamente de la familia de las Estrategias de Evolución al problema de la estimación de un parámetro que modelize la degradación por efecto del cortante en elementos de hormigón armado, un problema costoso computacionalmente y de gran relevancia en el ámbito de la ingeniería de estructuras, cuya resolución algorítmica no ha sido sin embargo abordada de forma extensa en la literatura.

# Abstract

This paper studies the applicability of evolutionary algorithms, particularly, the *evolution strategies* family in order to estimate a degradation parameter in the shear design of reinforced concrete members. This problem represents a great computational task and is highly relevant in the framework of the structural engineering that for the first time is solved using genetic algorithms.

# Índice General

# Índice de Figuras

Capítulo 1.

# Introducción.

Cuando evaluamos la aplicabilidad de las técnicas algorítmicas de inspiración evolutiva a la optimización de sistemas complejos encontramos que éstas vienen avaladas por un enorme número de casos de uso y resultados de investigación abarcando estudios de todo tipo. Por separado o conjuntamente a otros métodos, los algoritmos evolutivos se han abierto paso y aparecen ligados a disciplinas muy dispares como: la Genética, la Robótica, la Física Experimental, la Ingeniería del Software, la Ingeniería Civil, el Control de Sistemas Críticos, el Diseño Industrial o la Ingeniería de Materiales. En los siguientes capítulos, tratamos de dar una perspectiva amplia sobre qué es la computación evolutiva, el conjunto de técnicas y variantes algorítmicas que comprende y cómo estas pueden ayudar en el trabajo de campo de la modelización numérica y experimental en Ingeniería Civil.

## 1.1 Algoritmos evolutivos: Visión general

Los algoritmos evolutivos son estrategias de optimización y búsqueda de soluciones que toman como inspiración la evolución en distintos sistemas biológicos. La idea fundamental de estos algoritmos es mantener un conjunto de individuos que representan una posible solución del problema. Estos individuos se mezclan y compiten entre sí, siguiendo el principio de selección natural por el cual sólo los mejor adaptados sobreviven al paso del tiempo. Esto redunda en una evolución hacia soluciones cada vez más aptas.

Los algoritmos evolutivos son una familia de métodos de optimización, y como tal, tratan de hallar una *tupla* de valores *(xi,...,xn)* tales que se minimice una determinada función *F(xi,...,xn)*. En un algoritmo evolutivo, tras parametrizar el problema en una serie de variables, *(xi,...,xn)* se codifican en una población de cromosomas. Sobre esta población se aplican uno o varios operadores genéticos y se fuerza una presión selectiva (los operadores utilizados se aplicarán sobre estos cromosomas, o sobre poblaciones de ellos). Esta forma de funcionamiento les confiere su característica más destacable: un algoritmo evolutivo puede ser implementado de forma independiente del problema, o a lo sumo, con un conocimiento básico de éste, lo cual los hace algoritmos robustos, por ser útil para cualquier problema, pero a la vez débiles, pues no están especializados en ningún problema concreto siendo los operadores genéticos empleados los que en gran parte confieren la *especificabilidad* al método empleado.

En los últimos años son muchos los esfuerzos dedicados por investigadores de todo el mundo al desarrollo y la aplicación de nuevos operadores y nuevas variantes algorítmicas evolutivas especializadas en los más diversos problemas. Sin embargo, no deben considerarse a las técnicas evolutivas como técnicas aisladas ni pensar que son adecuadas a todos los casos, si no entenderlas en el contexto de la que hoy se conoce como técnicas algorítmicas de *Soft-Computing* [69, 70, 71, 72, 79, 80], una rama de investigación muy activa en la actualidad que viene a recoger el testigo de los avances producidos en el campo de la Inteligencia Artificial después de que este resurgiera con fuerza a mediados de los años 80 del pasado siglo.

Debido a su alta aplicabilidad, los algoritmos evolutivos han tenido una adopción muy rápida tanto en la industria como en el ámbito civil o militar. Actualmente, muchos centros de investigación dedican ya importantes inversiones dentro de sus presupuestos a la aplicación de técnicas de inteligencia artificial en general y de computación evolutiva en particular [81, 82, 83] y cada vez existen más profesionales formados y especializados en su aplicación y uso. La Inteligencia Artificial se perfila por tanto como una herramienta imprescindible para el trabajo de ingeniería y que se encuentra ya embebida en multitud de dispositivos y paquetes comerciales de software permitiendo aportar una solución de caja negra a muchos problemas, haciendo posible hallar soluciones realistas y computacionalmente tratables en áreas como:

a. Optimización numérica, real o simbólica, en situaciones donde existe una alta dimensionalidad, varios objetivos enfrentados, o no es posible conocer a priori la forma del espacio objetivo [61, 62, 73].

b. Aprendizaje automático, clustering, clasificación y reconocimiento de patrones [74, 75].

c. Implementación de sistemas robustos, capaces de reaccionar ante situaciones anómalas o inesperadas. [51, 52, 76]

d. Conseguir comportamiento emergente, es decir, sistemas con respuesta adaptativa, capaces de lidiar con problemas computacionalmente difíciles y de obtener soluciones válidas ante cambios en los parámetros del problema. [49, 59, 77]

e. En sistemas empotrados o tiempo real, capaces de dar una respuesta rápida de calidad aceptable en tiempos acotados. [5, 6, 7, 8, 78]

f. En el tratamiento de flujos de información, compresión de bloques o procesamiento de grandes volúmenes de datos, en usos como la extracción de conocimiento, eliminación de ruido o la predicción de secuencias. [10, 33, 37, 38, 39]

De entre estas aplicaciones, nos interesaremos especialmente en aquella dedicada a optimización numérica (ya que en ella esta englobado este trabajo), donde se trata, de extrapolar el potencial de la computación evolutiva como una poderosa herramienta de optimización de diseños a la resolución de problemas generales de optimización de naturaleza heterogénea.

## 1.2 Alcance y objetivos del proyecto

En ámbitos como la ingeniería o el diseño industrial se presentan con frecuencia problemas de optimización de distinto grado de dificultad. De entre ellos destaca un subconjunto especial de problemas denominado conjunto NP. Un problema perteneciente a este grupo es aquel para el cual no se conoce un algoritmo exacto de resolución cuyo coste computacional guarde una relación polinomial respecto al tamaño de la entrada, esto es, que no podemos abarcar el problema simplemente aumentando nuestra capacidad de cálculo, siendo esto lo que los hace de difícil resolución. La tarea se vuelve aún más difícil cuando el problema a resolver tiene una alta dimensionalidad por la presencia de un gran número de características [1,2] o variables de entrada [3,4], obteniéndose un crecimiento cercano a exponencial en el tiempo de cómputo conforme se incrementa de forma lineal el número de variables consideradas. Una dificultad adicional a considerar al lidiar con un problema de optimización es que, a veces, pequeñas variaciones en las variables de entrada del problema pueden ocasionar grandes cambios en el espacio de soluciones, (esto da lugar a los llamados sistemas caóticos), lo que hace necesario que el algoritmo ofrezca una capacidad manejable de re-parametrización [5,6]. En este ámbito las metaheurísticas aparecen como una nueva e innovadora manera de encontrar buenos ajustes y parámetros de calibración a modelos matemáticos que por su complejidad no admiten el análisis analítico, siendo además muy tolerantes a la necesitad de re-parametrizar el problema o realizar ajuste fino cuando aparece un nuevo conjunto de condiciones en el entorno [7,8].

La familia de algoritmos analíticos o exactos, usados tradicionalmente como herramientas para abordar problemas de optimización, garantizan encontrar el óptimo global en muchos problemas, pero por contra, tienen el grave inconveniente de que en problemas reales pueden elevar el tiempo de ejecución necesario hasta hacer inabarcables los costes de su implementación siendo por tanto inasumibles y obligando a considerar técnicas diferentes al análisis matemático exhaustivo.

En contraste, los algoritmos heurísticos adhoc son normalmente bastante rápidos [84, 85, 99, 100], pero la calidad de las soluciones encontradas puede ser a veces insuficiente y estar lejos de ser óptima, además de presentar el inconveniente adicional de determinar los discriminantes heurísticos, difíciles de definir en determinados problemas.

Las metaheurísticas ofrecen un equilibrio adecuado entre ambos extremos: son métodos genéricos que ofrecen soluciones de buena calidad (el óptimo global en muchos casos) en un tiempo moderado [11]. Además, la naturaleza de los problemas en el ámbito científico (por ejemplo simulaciones físicas, modelado de materiales, diseño industrial, etc.) hace que en ocasiones el proceso de cómputo se realice de manera descentralizada, bien en grandes computadores especializados [9, 10], bien mediante redes dedicadas para el cálculo distribuido, y por este motivo, los algoritmos evolutivos se adaptan, por lo general, muy bien a las infraestructuras de cálculo disponibles mostrándose como herramientas de paralelización eficaces en un campo de investigación notablemente reclamado en el ámbito científico actual.

En las siguientes secciones se discutirá las posibilidades de aplicación de varias clases de algoritmos evolutivos a problemas de diseño avanzado en ingeniería civil difícilmente abarcables mediante computación exhaustiva por su alta dimensionalidad y su carácter multimodal. La realización del presente trabajo se enmarca dentro de los objetivos académicos para la obtención del título de **Grado en Ingeniería Informática** mención en Ingeniería de Software. En él, abordaremos un problema de optimización elegido por su alta dificultad y su trascendencia en el ámbito de estudio y llevaremos a cabo su resolución empleando técnicas evolutivas donde esperamos encontrar soluciones de calidad comparables a las conseguidas a través de las técnicas empleadas habitualmente y recogidas en la bibliografía pero con un menor costo computacional. Los objetivos que se marcan como mínimo exigible para este proyecto fin de grado incluyen:

- Entender el problema propuesto, realizar una búsqueda bibliográfica que nos permita situar el problema y comprender su contexto así como cuál ha sido el abordaje tradicional y que trabajos conforma el estado de la cuestión.
- Comprender, compilar y testar los fuentes de código que constituyen la implementación del problema y valorar las posibles formas de mejora y optimización.
- Proponer, analizar, e implementar una o varias propuestas de resolución mediante técnicas meta-heurísticas.
- Analizar los resultados obtenidos por cada uno de los métodos cuantificando las mejoras conseguidas por cada uno y realizando un análisis crítico sobre indicadores mesurables como el tiempo de cómputo, la calidad de la solución, o la complejidad algorítmica.

## 1.3 Estructura y desarrollo del proyecto

Para el presente trabajo decidimos abordar dos partes bien diferenciadas: en una primera parte llevaremos a cabo un acercamiento a la problemática que queremos abordar y analizaremos las características generales de las meta-heurísticas como herramientas de parametrización de modelos numéricos y el estado del arte de la investigación en este campo. En una segunda parte, introduciremos el problema de estimación del parámetro kappa para el hormigón armado solicitado a cortante y trataremos de proponer un nuevo enfoque para su resolución basado en técnicas meta-heurísticas.

En el capítulo 2, daremos algunas nociones básicas sobre los diferentes enfoques algorítmicos que existen para la optimización de funciones continuas y realizaremos un recorrido bibliográfico sobre el desarrollo en este campo y las vertientes donde se centra la investigación en estas áreas en la actualidad.

En el capítulo 3, propondremos la resolución del problema objeto de estudio mediante Estrategias de Evolución y propondremos distintas variantes y su aplicabilidad.

En el capítulo 4, expondremos la metodología empleada en la gestión del proyecto y en la construcción del software que lo acompaña desde el punto de vista de la ingeniería del software, los diferentes criterios que se han seguido en relación a la organización de su desarrollo y la organización de las tareas. Ideas, que desarrollaremos en el capítulo 5 donde propondremos una planificación temporal para el proyecto y daremos una primera estimación de costes.

En el capítulo 6 haremos un breve recorrido por *Wolfram Mathematica,* sus fortalezas y sus ventajas sobre otras herramientas software para modelización matemática.

Finalmente el capítulo 7 revisará de forma analítica los resultados obtenidos por las diferentes propuestas y mostrará los resultados de forma gráfica y comparada.

La última parte del trabajo consistirá en la extracción de conclusiones, exposición de limitaciones del estudio y planteamiento de posibles líneas de ampliación futuras.

Capítulo 2.

# Estado de la cuestión.

Hay muchas formas de abordar un problema de optimización. Probablemente la más directa, máxime cuando los detalles del problema pueden formularse de forma clara y concisa, sea la mera aproximación analítica. Sin embargo el abordaje matemático directo no es siempre posible, y los modelos que son analíticamente tratables habitualmente son también demasiado generales para proveer la precisión que requiere un sistema físico. A esto debemos sumar que el tratamiento analítico resulta difícil y por tanto, costoso computacionalmente sobre todo cuando trabajamos con funciones de varias variables. Una familia de algoritmos ampliamente utilizada son los algoritmos de escalado de colinas (del inglés *hillclimbing*) que basan su estrategia en intentar encontrar el valor mínimo de una función mediante evaluaciones sucesivas. El mecanismo general es bastante simple y su implementación sencilla, pero el éxito de la búsqueda en este tipo de métodos no depende sólo de la función objetivo, sino también de la política de escalado que escojamos o dicho de otra forma de si escogemos apropiadamente la función paso (iteración) y el punto de inicio de la búsqueda (ver Figura 1).

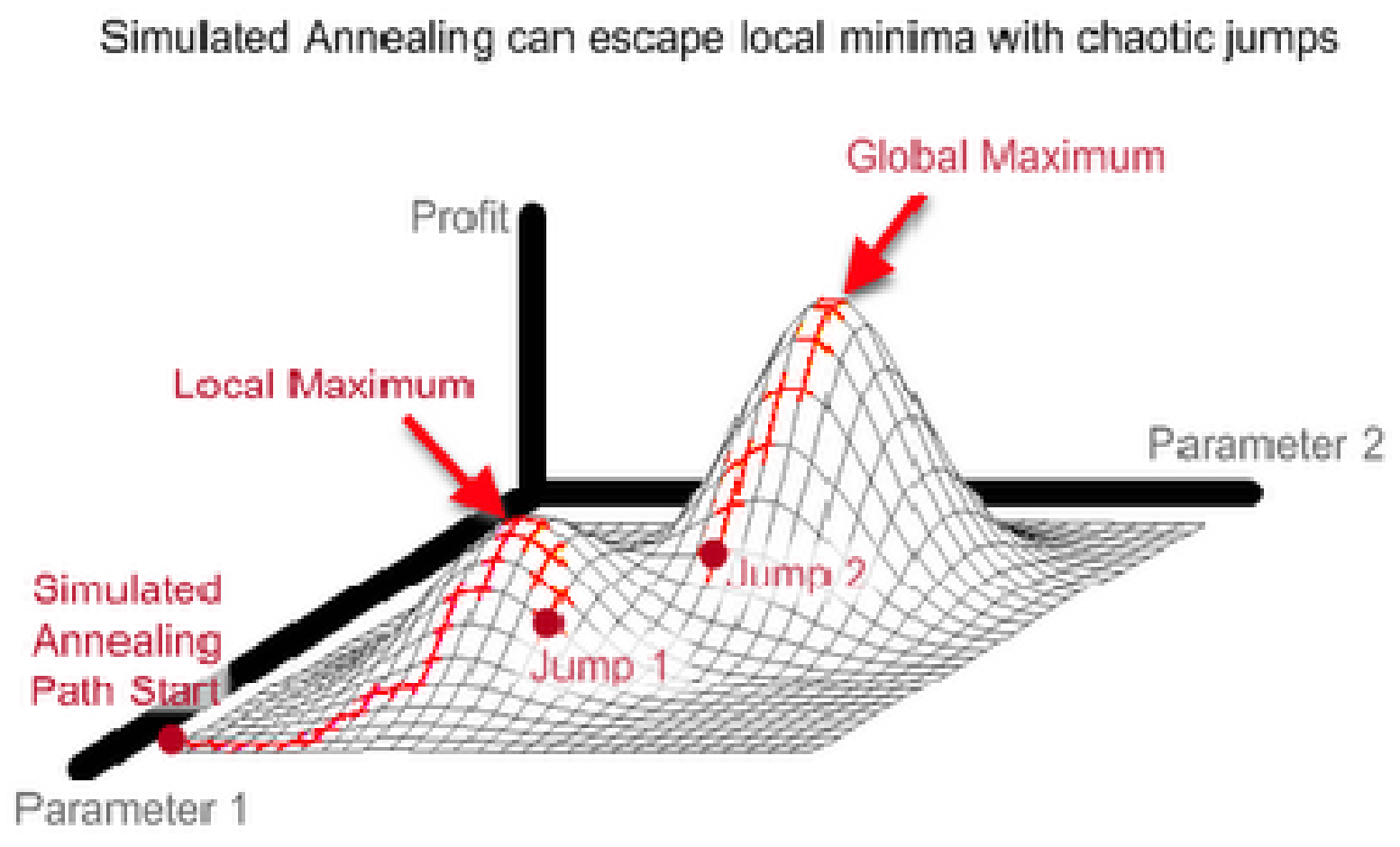

*Figura 1. Algoritmo escalador, en este caso un algoritmo de recocido simulado (Stanford University Information Theory Group at at https://itservices.stanford.edu).*

Aun así se trata de algoritmos pensados para realizar un refinamiento de una solución que ya era en inicio razonablemente buena y por ello son sobretodo técnicas de búsqueda local, centrándose en encontrar el máximo más cercano al punto de inicio. Un algoritmo escalador simplemente evaluará la función en uno o más puntos e irá moviéndose por ella en pequeñas variaciones buscando el punto que maximiza la evaluación de la función objetivo. Por ello, frecuentemente este tipo de métodos no son adecuados para paisajes de búsqueda complejos, pues suelen atascarse en mínimos locales.

Una variante interesante que intenta evadir los problemas de los algoritmos escaladores es el método de recocido simulado [114, 115] llamado así por su similitud a como se forman los metales en una forja. La técnica de recocido simulado intenta escapar de estos falsos puntos solución aceptando como valor siguiente algún vecino que no tiene necesariamente que ser mejor que las soluciones ya encontradas. Para decidir cuándo un valor, aun siendo peor, podría ser prometedor en el futuro se utiliza un factor de aceptación nombrado como temperatura. Así, de esta manera, el nuevo valor será candidato a solución siempre que la diferencia entre el punto en el que se encuentra el algoritmo y el siguiente punto a seleccionar sea menor que el valor actual de temperatura. A pesar de su aparente sencillez, esta idea permite que el algoritmo muestre una mayor tolerancia y sea más difícil el estancamiento. Sin embargo, el hecho de evitar quedar atascados e incluso alcanzar la condición de convergencia no nos asegura que estemos cerca de un valor óptimo ya que se sigue dependiendo demasiado del punto donde se inicia la búsqueda.

Parte de las problemáticas anteriores se pueden resolver usando una estrategia de multi-comienzo [116, 117]. Este tipo de estrategias, en realidad, no son más que la ejecución paralela de varios de estos algoritmos, y por tanto, aunque son mejores, tampoco nos garantizan que se encuentre, o incluso se aproximen al máximo global. En todo caso, tiene la ventaja de que, en cada iteración del algoritmo, se tiene una solución válida, aunque no tiene porqué ser la mejor posible.

Empleemos el método que empleemos, para no degradar en un proceso de búsqueda semi-ciega (búsqueda voraz) todos estos algoritmos necesitan una pista de hacia dónde deben avanzar para alcanzar la mejor solución. A esta función guía es a lo que en computación se le llama heurística. Una heurística consiste básicamente en usar una serie de reglas (un conocimiento previo mínimo) que permiten al algoritmo avanzar hacia la resolución de un tipo de problema.

Los métodos metaheurísticos fueron introducidos por vez primera por Fred Glover (también conocido por ser el creador del *método Tabú*) en [12] y surgieron al combinar diferentes métodos heurísticos con el objetivo de alcanzar una mayor eficiencia, robustez y eficacia en la exploración del espacio de búsqueda. Fueron diseñados para resolver problemas de optimización industrial (optimización combinatoria y continua), donde no existía un conocimiento profundo de la forma que tiene el espacio *n-dimensional* de soluciones. Los algoritmos evolutivos pertenecen a este grupo de técnicas.

## 2.1 Metaheurísticas

Una metaheurística se diferencia de un método heurístico tradicional en que en las primeras existe un mecanismo o estrategia de alto nivel que guía o supervisa la aplicación y la evolución de las diferentes técnicas heurísticas que se utilizan en la resolución de un problema de forma que se maximiza la eficacia de éstas. Actualmente existen diferentes formas de clasificar a los algoritmos meta-heurísticos [7]  encontrando una u otra dependiendo del texto y autor consultado, entre las más aceptadas y utilizadas encontramos:

a) Algoritmos bio-inspirados / no bio-inspirados: Según si se basan o no en mecanismos biológicos, por ejemplo, en la evolución de las especies, el vuelo de los pajaros en bandada, o el comportamiento de diversos patógenos.

b) Estáticos / Dinámicos, Monoobjetivo / Multiobjetivo: Si se utiliza la misma función objetivo o no durante todo el proceso evolutivo. Es decir, si son usadas una única función objetivo o varias.

c) Basados en la evolución de un conjunto de soluciones (población) / basados en una única solución, es decir, si existe una población de soluciones que convergen hacia la solución o una única solución que se acerca a la solución mediante un proceso de refinado sucesivo.

d) Algoritmos con memoria / sin memoria: Esta clasificación depende del uso que hacen de su historia de búsqueda, es decir, si utilizan algún tipo de memoria auxiliar a la exploración o no.

En este trabajo adoptaremos la clasificación de acuerdo al criterio bajo el epígrafe ‘c’, por ser esta es una terminología ampliamente aceptada en producción e investigación científica, y, porque esta nos permite realizar una clasificación flexible donde podemos generalizar lo suficiente al hablar de la familia de técnicas comúnmente denominadas técnicas de inspiración evolutiva (AE). Diremos por tanto, que las técnicas evolutivas, que desarrollaremos en profundidad, pertenecen al grupo de las metaheurísticas basadas en población.

### 2.1.1 Conceptos básicos de algoritmos evolutivos

Como hemos visto en las secciones anteriores, cuando modelamos problemas de la vida real uno de los problemas más frecuente es la alta dimensionalidad de espacio de búsqueda. También es habitual encontrarse con problemas en los que las condiciones para la optimalidad de la solución varían a lo largo del tiempo o no existen como tal. Este tipo de problemas se conocen como problemas multimodales [90, 118, 119].

No siempre las condiciones del problema permanecen estáticas durante la duración del mismo; puede ser que el espacio de búsqueda aumente o disminuya, que la valoración de una solución cambie, o simplemente que la forma más simple de resolver un problema consista en resolver previamente una serie de sub-problemas, que vayan acotando la solución cada vez más. El concepto de multi-modalidad ha sido usado en diferentes contextos: estrictamente, un problema multimodal es un problema que tiene varios máximos, todos ellos de la misma jerarquía, pero también se aplica a aquellos problemas que tienen varias soluciones posibles.

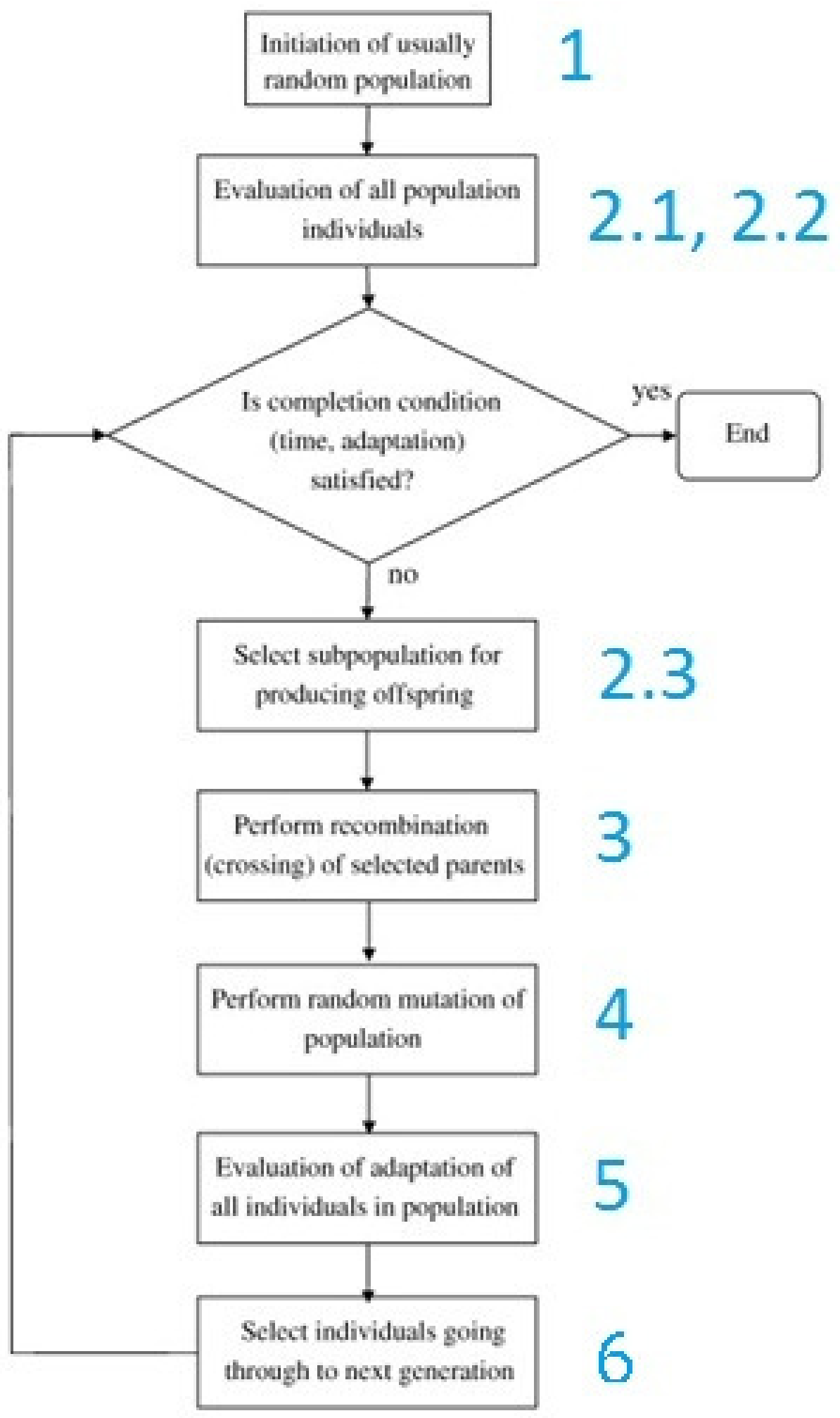

*Figura 2. Esquema general de un algoritmo evolutivo (Rogalska et al. 2008)*

En general, casi todo problema de búsqueda, formulado como un problema de optimización, suele tener varios máximos, uno de ellos es mejor que el resto, y este se le denomina máximo global. El resto son máximos locales; es decir, se puede definir una vecindad alrededor de ellos en la cual son máximos globales.

Los AE son, a grosso modo, un método de optimización basado en población, especialmente útiles cuando tratamos un problema donde es costoso realizar un gran número de iteraciones o donde existen dos o más funciones objetivo ya que se adaptan excepcionalmente bien a problemas multi-objetivo. Por lo tanto, los AE están indicados para resolver todo tipo de problemas que puedan ser expresados en forma de problema de optimización de una o varias funciones sujetas a un número variable de restricciones y a una o más restricciones de contorno. Los AE son tremendamente sensibles a la manera en que codifiquemos a los individuos de la población y la codificación utilizada puede influir sensiblemente en las posibilidades de convergencia. Tanto es así que algunas variantes de AE se diferencian precisamente en elegir una u otra forma de codificación interna. Por tanto, la tarea más importante en un AE será encontrar la representación adecuada para las soluciones. La segunda tarea crítica será elegir correctamente la función que guiará la búsqueda, y que puede ser única o formarse de la combinación ponderada de varias funciones. Esta función suele recibir el nombre de función objetivo ó función de *fitness* (adecuación). Básicamente, los algoritmos genéticos funcionan como sigue: dada una población de soluciones candidatas, y en base al valor de la función objetivo para cada uno de los individuos (soluciones) de esa población, se seleccionan los mejores (los que minimizan la función objetivo) y se combinan para generar otros nuevos. Este proceso se repite cíclicamente.

En primer lugar, debemos contar con un modelo matemático que permita evaluar un punto del espacio de soluciones en que se define el problema. En otros términos, tenemos que poder plantear el problema como un problema de minimización (o maximización) de una función objetivo, que representa la presión selectiva del medio. En segundo lugar, deberá especificarse la manera de codificar las soluciones. Las codificaciones más sencillas, son aquellas que están basadas en código binario (representación en forma de cadena de bits). La interpretación que sedé a esa cadena dependerá de la naturaleza de la solución (puede ser la codificación de un valor entero, de un real, un vector de valores boléanos... y cualquier otra estructura de datos).

La figura 2 muestra el esquema algorítmico general que puede encontrarse en cualquier algoritmo evolutivo con pequeñas modificaciones en lo esencial. En primer lugar (1) se procede a la inicialización de la población. Para cada individuo de la población se selecciona un valor que puede ser completamente aleatorio. También puede considerarse tomar como valores iníciales una aproximación a la solución. Después se aplica a cada individuo la función objetivo, es decir, la función objetivo nos dice cuan bueno es un individuo como solución, lo que da una medida de lo adaptado que está cada uno de ellos. En función del valor obtenido se ordena la población (2.2), quedando así en primer lugar los individuos más adaptados. Se seleccionan entonces los individuos que se van a cruzar (2.3).

Elegir uno u otro método de selección determinará la estrategia de búsqueda del Algoritmo. Si se opta por un método con una alta presión de selección se centra la búsqueda de las soluciones en un entorno próximo a las mejores soluciones actuales. Por el contrario, optando por una presión de selección menor se deja el camino abierto para la exploración de nuevas regiones del espacio de búsqueda. Entre los métodos de selección más usuales encontramos:

Selección basada en rango: Según este criterio se seleccionan los *k* individuos mejor adaptados.

Selección por ruleta: Consiste en dar a cada individuo una probabilidad de ser seleccionado proporcional a su fitness. En este tipo de selección un individuo puede seleccionarse dos veces y cruzarse consigo mismo.

Selección por torneo: La idea principal de este método consiste en realizar la selección en base a comparaciones directas entre individuos. Existen dos versiones de selección mediante torneo según esta elección se realice de forma determinista o probabilística.

Selección a medida (ad hoc): Muchas veces la naturaleza del problema o de los datos de entrada fuerzan a que la selección se realice siguiendo criterios específicos al problema.

En investigación frecuentemente es necesario definir operadores genéticos a medida, que se ajustan mejor a las características del problema. Un operador especializado suele ser una buena opción si es posible su implementación. El operador de cruce, normalmente representa el mecanismo más importante y está en la base de todos los AE ya que en este paso es donde se produce el intercambio de información genética. Aunque existen muchas variantes a la operación de cruce (crossover en lengua inglesa) entre las más usuales encontramos:

*Crossover* n-puntos: los dos cromosomas se cortan por n puntos, y el material genético situado entre ellos se intercambia. Lo más habitual es un crossover deun punto o de dos puntos.

*Crossover* uniforme: se genera un patrón aleatorio de 1s y 0s, y se intercambian los bits de los dos cromosomas que coincidan donde hay un 1 en el patrón. O bien se genera un número aleatorio para cada bit, y si supera una determinada probabilidad se intercambia ese bit entre los dos cromosomas.

*Crossover* especializados: en algunos problemas, aplicar aleatoriamente el crossover da lugar a cromosomas que codifican soluciones inválidas; en este caso hay que aplicar el crossover de forma que generen siempre soluciones válidas. Un ejemplo de estos son los operadores de crossover usados en el problema del viajante. También aparecen operadores de cruce especializados en codificaciones muy especializadas o en problemas donde se ha llevado un estudio exhaustivo y tenemos un amplio conocimiento del problema y queremos traspasarlo al algoritmo. En el paso 4 se genera en la descendencia una mutación en un alelo (en el caso más simple la mutación consiste en negar un bit) aleatorio. Esto sucede con una probabilidad baja (esto es, en la mayoría de los casos, el paso 4 no tiene efecto).

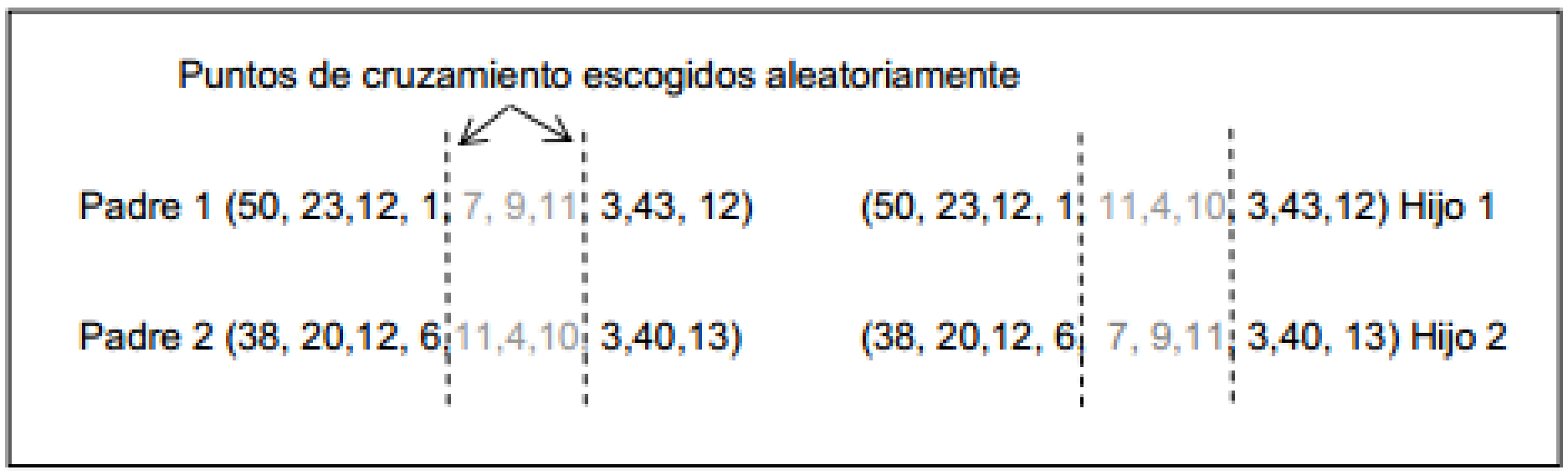

*Figura 3. Representación gráfica del operador de cruce a dos puntos.*

Una vez se ha generado la descendencia, deberá insertarse en la población (5). Para llevar a cabo la inserción existen diversas políticas:

- Eliminar a los individuos peor adaptados e insertar los recién generados.

- Eliminar al individuo peor y reemplazarlo.

- Reemplazar toda la población

Debemos elegir cuidadosamente cada uno de estos parámetros ya que de ellos dependerá el éxito de nuestro algoritmo y la política de selección y reemplazo adoptada es crucial en la convergencia del individuo.

Por último, en el paso 6 se comprueba si alguno de los individuos disponibles satisface los criterios establecidos y se puede considerar como solución al problema. En este paso también se puede comprobar si se ha excedido un número de iteraciones o un límite de tiempo. Si no es así y ningún individuo cumple los criterios de parada, se vuelve al paso 2.3.

Este sencillo algoritmo junto los mecanismos de cruce y selección se encuentran en sus distintas formas en todos los algoritmos genéticos que con el tiempo han ido incorporando a un amplio abanico de operadores genéticos más específicos y complejos.

### 2.1.2 Técnicas evolutivas. Clasificación y tipos

Existen varias aproximaciones a la idea de algoritmos evolutivos pero todos ellos son similares en su planteamiento básico y en su arquitectura básica, difiriendo principalmente en la forma en que representan y manipulan la información así como la importancia que dan a unos operadores genéticos sobre otros.

A pesar de que bajo la terminología de "técnicas de computación evolutiva" se agrupan un gran número de variantes algorítmicas [20] la mayoría de los autores están de acuerdo en que estas pueden ser clasificadas en cuatro subgrupos:

- Algoritmos genéticos (Genetic algorithms)
- Programación genética (Genetic programming)
- Programación evolutiva (Evolutionary programming)
- Estrategias de evolución (Evolutionary strategies)

Los Algoritmos Genéticos (GA) como caso particular, son un tipo de algoritmo evolutivo que ha demostrado ser muy efectivo en la optimización de procesos no lineales [28,29], con saturación de ruido, y en general poco conocidos. Además los AG son algoritmos que pueden abarcar y aplicarse con éxito a un amplio espectro de problemas y para su diseño es suficiente con tener un conocimiento mínimo a priori acerca del sistema. Esto convierte a los algoritmos genéticos en un paradigma de aplicación deseable en un gran número de escenarios donde la complejidad del problema hace desaconsejable otro tipo de metodologías. A veces, cuando el usuario desea hacer una implementación rápida (aunque esta no sea la más eficiente) se emplea una simplificación de este esquema llamada Simple Genetic Algorithm (SGA).

Los algoritmos de Estrategia de Evolución (ES) fueron introducidos por Rechenberg et al. [22] a principio de los 70 en diferentes aplicaciones industriales e hidráulicas. Destacan sobre las otras porque están pensados para trabajar sobre espacios continuos (números reales) y porque los parámetros de funcionamiento del algoritmo (tasa de mutación, probabilidad de cruce.) no son fijos sino que forman parte del proceso de optimización. Por ejemplo, en el algoritmo CMA-ES [90, 91, 92], el cálculo de dichos parámetros se calcula a partir de la obtención de las matrices de covarianza en el espacio de n dimensiones.

Una aproximación evolutiva distinta es la Programación Genética (GP). Este paradigma permite abordar problemas de optimización no lineal basada en un lenguaje simbólico. El paradigma usado en programación genética también utiliza principios de selección darwiniana como la selección basada en *fitness*, pero los operadores genéticos ahora actúan sobre árboles simbólicos [30]. Por ejemplo, cada uno de estos árboles podría estar formado por sentencias de un lenguaje de programación determinado. Sale de lo convencional y se diferencia de los GA principalmente en lo que respecta a su sistema de representación. Las estructuras sometidas a adaptación son comúnmente programas completos que son ejecutables o conjuntos jerárquicos de reglas evaluables de forma dinámica y con tamaños y formas distintos. Frecuentemente este tipo de sistemas son *sistemas de configuración híbrida*. A modo de ejemplo, podemos citar el caso de la *GP-Fuzzy* [31] (Abreviatura inglesa de Programación genética difusa), que comprende una población de reglas difusas / bases (estructuras simbólicas) que se construyen mediante un proceso evolutivo donde son los candidatos a ser soluciones al problema, y evolucionan en respuesta a una presión selectiva inducida por su relativo éxito en la implementación de la conducta deseada. Este no es un ejemplo aislado, en muchos casos las técnicas evolutivas son usadas dentro de soluciones híbridas. Los métodos híbridos han demostrado ser eficaces en el diseño de sistemas inteligentes [32]. En los últimos años han proliferado todo tipo de soluciones híbridas que transgreden las líneas de separación entre diferentes algoritmos para tomas las características buenas de uno y otro. La lógica borrosa, las redes neuronales y los paradigmas evolutivos pueden ser y son metodologías complementarias en los trabajos de diseño e implementación de sistemas inteligentes. Cada uno de esos enfoques tiene sus ventajas e inconvenientes. Para aprovechar las ventajas y eliminar sus desventajas, en aplicaciones operativas reales muchos trabajos proponen la integración de varias de estas metodologías. Estas técnicas incluyen la integración de redes neuronales y técnicas de lógica difusa, así como la combinación de estas dos tecnologías con técnicas de computación evolutiva.

### 2.1.3 Estrategias de evolución (EE).

Las Estrategias de evolución (*Evolutionary Strategies* en lengua inglesa) son una familia de algoritmos estocásticos de optimización numérica de funciones no-lineales o problemas de optimización continua no convexa donde no es posible conocer a priori la forma del espacio de soluciones ni es factible realizar el cálculo de las derivadas sucesivas. El algoritmo CMA-ES (Estrategia de evolución de adaptación mediante covarianzas) [90, 91, 92] es un ejemplo de un miembro de esta familia.

Las estrategias de evolución fueron desarrolladas por Rechenberg [22] en su intento de resolver problemas difíciles en el campo de la hidrodinámica. La primera versión del algoritmo, llamada (1+1)-EE o estrategia e evolución de dos miembros utilizaba únicamente un padre y un descendiente. El descendiente se mantiene en la población solo si resultaba mejor que su padre. En la siguiente generación el siguiente hijo es calculado a partir de valores normales (ecuación 1), donde t se refiere a la generación actual y N es un vector de números Gaussianos con media 0 y desviación estándar σ.

$$\overline{X}_{t+1} = \overline{X}_t + N(0,\sigma) \qquad (1)$$

En sucesivos trabajos, Rechenberg extendió el concepto de población y propuso otras variantes como la variante (μ+1) – EE [23], en la cual hay μ pares que generan solo un descendiente el cual puede reemplazar al peor padre de la población.

Algoritmo 1: *Evolutionary Strategies (De Jong, 2006)*

$t := 0;$
*initialize* $P(0) := \{\vec{a}_1(0), \ldots, \vec{a}_\mu(0)\} \in I^\mu$
    **where** $I = \mathbb{R}^{n+w}$
    **and** $\vec{a}_k = (x_i,\ \sigma_i,\ \alpha_j\ \forall i \in \{1, \ldots, n\}\ ,\ \forall j \in \{1, \ldots, n \cdot (n-1)/2\});$
*evaluate* $P(0) :\ \{\Phi(\vec{a}_1(0)), \ldots, \Phi(\vec{a}_\mu(0))\}$
    **where** $\Phi(\vec{a}_k(0)) = f(\vec{x}_k(0));$
**while** $(\iota(P(t)) \neq$ **true**) **do**     % while termination criterion not fulfilled
    *recombine:* $\vec{a}_k'(t) := r'(P(t))\ \forall k \in \{1, \ldots, \lambda\};$
    *mutate:* $\vec{a}_k''(t) := m'_{\{\tau,\tau',\beta\}}(\vec{a}_k'(t))\ \forall k \in \{1, \ldots, \lambda\};$
    *evaluate:* $P''(t) := \{\vec{a}_1''(t), \ldots, \vec{a}_\lambda''(t)\} :$
        $\{\Phi(\vec{a}_1''(t)), \ldots, \Phi(\vec{a}_\lambda''(t))\}$ **where** $\Phi(\vec{a}_k''(t)) = f(\vec{x}_k''(t));$
    *select:* $P(t+1) :=$ **if** $(\mu, \lambda)$*-selection*
        **then** $s_{(\mu,\lambda)}(P''(t));$
        **else** $s_{(\mu+\lambda)}(P(t) \cup P''(t));$
    $t := t + 1;$
**od** .

Con posterioridad, serian Schwefel et al. quienes mejorarían el concepto añadiendo el uso de múltiples hijos y una variante multi-generacional, respectivamente (μ+λ) – EE, y (μ,λ) - EE [24]. En el primer caso, el el proceso de selección la descendencia y los padres son tenidos en cuenta de forma equitativa; En el segundo caso, solo se tiene en cuenta la descendencia. El esquema general de un algoritmo de estrategia de evolución aparece recogido en el algoritmo 2.1.

De especial relevancia es el método de selección, que en las estrategias de evolución se realice de forma determinista, razón por la cual solo los mejores individuos pasan a la siguiente generación. El operador principal es la mutación, realizando el operador de recombinación un papel únicamente secundario que incluso se omite en algunos casos.

Según algunos autores [102,103] cuando nos enfrentamos a un problema de optimización complejo son buenas guías de cuando debería considerarse el uso de EE el que se den varias de las situaciones siguientes:

Funciones no-lineales
Funciones de variables no-separables
Espacios de búsqueda no convexos
Situaciones de multi-modalidad
Funciones caóticas o con gran cantidad de ruido
Media o alta dimensionalidad (desde 5 hasta 100 dimensiones)

Cuando una o más e estas características concurren en la naturaleza de la función a optimizar (ver figura 4), el uso de estrategias de evolución puede reportar importantes beneficios.

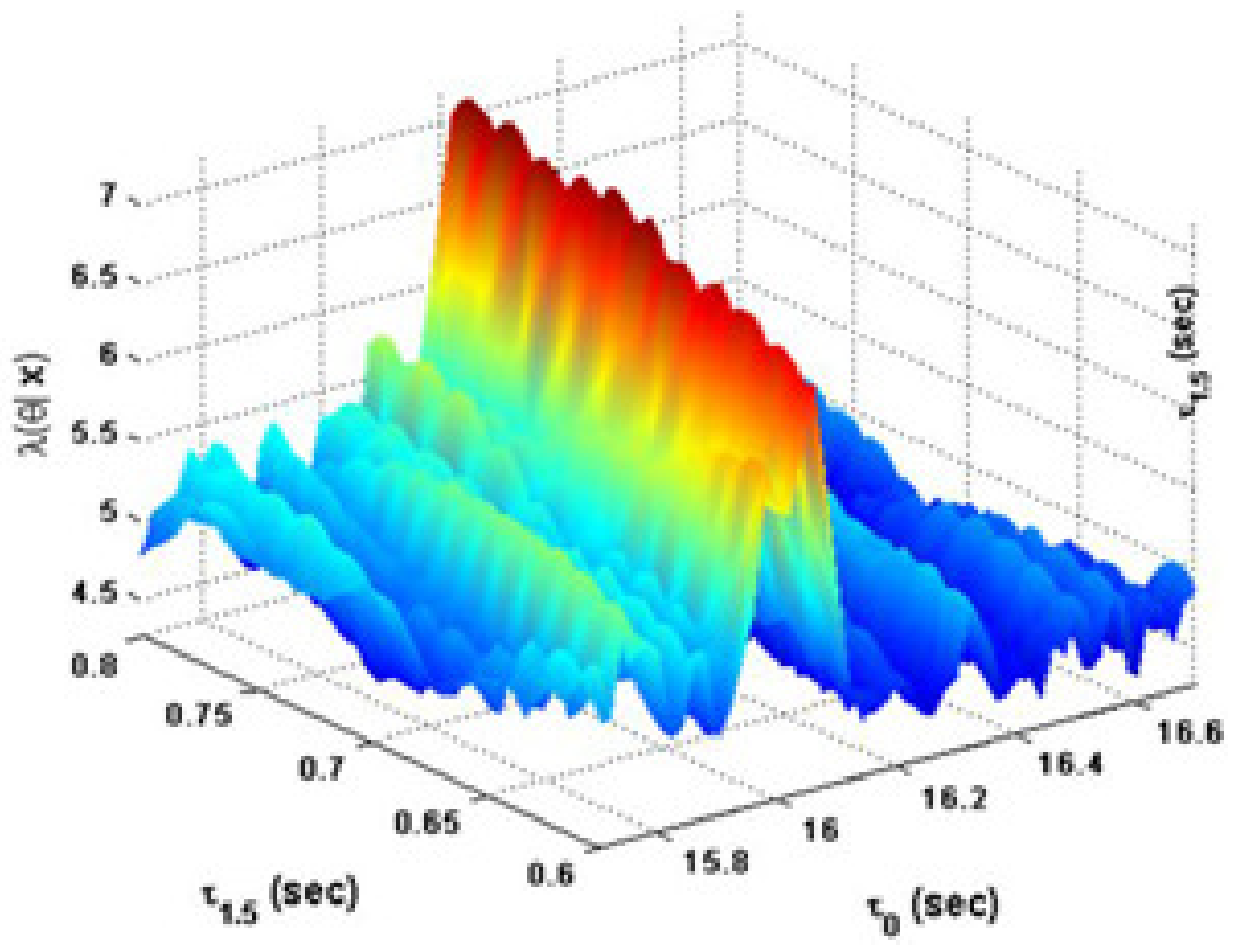

*Figura 4. Ejemplo de buena función candidata para optimización mediante estrategias de evolución en la que se observan numerosos máximos locales (Función de EggHolder).*

## 2.2 Algoritmos evolutivos en aplicaciones industriales e ingeniería

Los Algoritmos Evolutivos son una potente herramienta en optimización de formas. Debido a su relativa sencillez en comparación con otros métodos algorítmicos, su facilidad de uso, y su capacidad para adaptarse a los problemas de optimización multi-objetivo, los AE han sido aplicados a problemas de optimización de diseños en muchas áreas [21, 35, 40], entre estas aplicaciones se encuentran la resolución de problemas reales de diseño de materiales o todo tipo de conjuntos aerodinámicos, desde conducciones de aire, turbinas o compresores, hasta diseños de motores, diseños mecánicos completos o conjuntos de alerones.

Los AE han sido aplicados con éxito en proyectos de toda envergadura produciendo en ocasiones importantes reducciones de costes y mejoras sustanciales en los resultados conseguidos [93, 94, 95, 96, 97]. Revisando la bibliografía encontramos referencias muy tempranas, dando testimonio de la gran solidez teórica de la que disponen estas técnicas a día de hoy. Entre ellos, encontramos los trabajos de referencia de Fogel [21] quien estudió la aplicabilidad de lo que hoy conocemos como técnicas de programación evolutiva al diseño de autómatas; Rechenberg [22,23], que resolvió con éxito por primera vez problemas industriales de hidráulica utilizando algoritmos poblacionales que simulaban evolución, muy parecidos a las actuales estrategias evolutivas; y Schwefel [24], que estudió su aplicación a problemas de optimización numérica, todos ellos de principios de los 70.

Más tarde sería Holland [25], quien, basándose en el trabajo que venía desarrollando en el estudio de sistemas adaptativos sentaría la base teórica formal de todos los algoritmos evolutivos actuales dando forma a la teoría de los esquemas [25, 26].

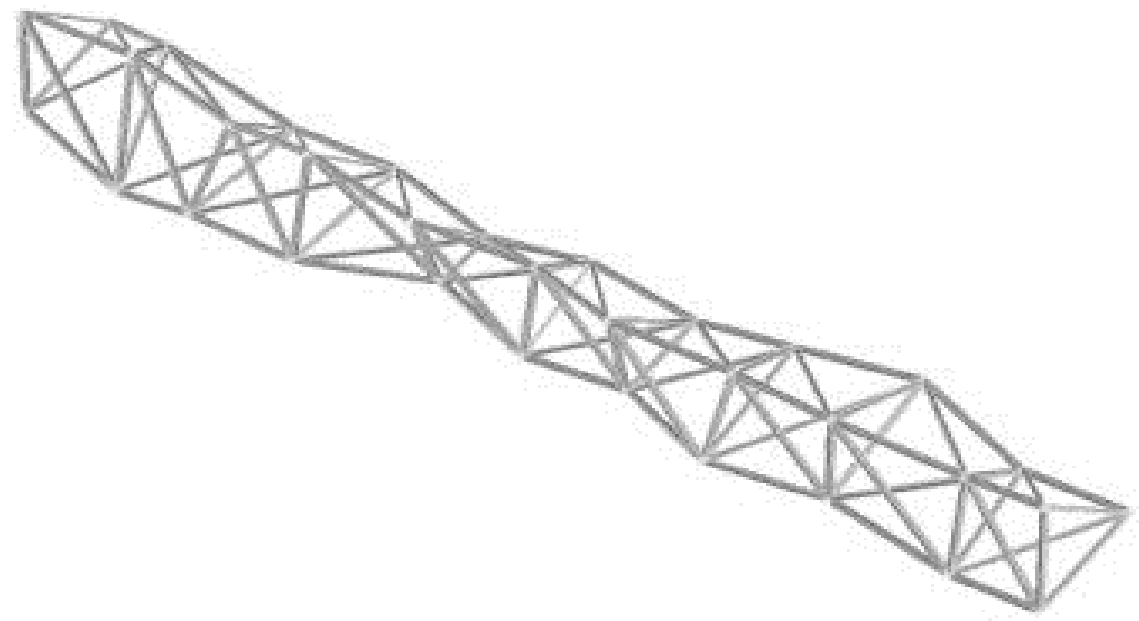

*Figura 5. Un brazo tridimensional optimizado genéticamente, con una respuesta mejorada a la frecuencia (Keane et al. 1995)*

Mucho más recientemente, queda patente al consultar la bibliografía, un resurgir del interés por los algoritmos evolutivos en la primera mitad de los años 90. Son muchos los trabajos que se publican en esos años, aunque podemos destacar algunos; Por ejemplo en el campo de la física encontramos el trabajo de Charbonneau et al. entre los años 1994 y 1996, centrados en la elaboración de una extensa toolbox para la utilización de algoritmos genéticos en aplicaciones científicas como el modelado del viento solar, la simulación de interacciones entre cuerpos masivamente pesados o la estimación de distancias basadas en el efecto Doppler [33]. En el campo de la acústica, y casi al mismo tiempo, cabe destacar el trabajo de Tang et at. [34], en que se analizan las posibles ventajas del uso de algoritmos genéticos en el análisis de ondas y el procesamiento de señales. Orientados a la industria aeroespacial, encontramos las publicaciones de Keane y Brown [35], que utilizaron un algoritmo genético para producir nuevos diseños para brazos o jirafas destinados a transportar carga pesada que pudiesen montarse en órbita y utilizarse con satélites, paseos espaciales y otros proyectos de construcción aeroespacial.

En un enfoque diferente, Altshuler y Linden [36] a finales de 1997, utilizaron un algoritmo genético para conseguir formas evolutivas de antenas de alambre con propiedades especificadas a priori por la parametrización dada. Y en 2000, Hughes y Leylanden [37] centraron su investigación en los problemas de optimización multi-objetivo, y aplicaron con éxito algoritmos genéticos modificados a problemas multi-objetivo como el de la clasificación de objetivos basándose en sus reflexiones radar.

Algo después, en 2002, Gurfil et al. utilizarían un algoritmo similar aplicándolo a la caracterización de orbitas geocéntricas [38], poniendo especial énfasis en su estudio en encontrar soluciones sub-óptimas, por la importancia que tienen este tipo de órbitas para el posicionamiento de satélites pues permiten una alta tasa de transferencia en comunicaciones mientras se mantienen dentro de un entorno operacional seguro y fuera de perturbaciones térmicas y de radiaciones e interferencias producidas por el campo magnético de la tierra.

A finales de ese año, encontramos los trabajos de Metcalfe et al. aplicados al ámbito de la física [39], donde se lleva a cabo la implementación de un algoritmo genético distribuido para la determinación de parámetros globalmente óptimos para el ajuste de modelos matemáticos experimentales en la inferencia de información física y estructural de cuerpos celestes a través del análisis de sus frecuencias de oscilación (estudio de sus periodos de pulsación) abriendo la puerta a la automatización masiva de este tipo de observaciones.

En el campo de la optimización de diseños, distinguimos los experimentos de Galvão et al. [40] donde los autores utilizaron algoritmos genéticos para diseñar polímeros conductores de electricidad basados en el carbono, conocidos como polianilinas. Y centrándonos en diseño industrial, cabe mencionar el notorio trabajo de Oyama et al., donde se aplican algoritmos evolutivos al rediseño de rotores transónicos de cohetes [44] así como los trabajos de Liou et al. [45] y Lian et al. [46,47] aplicados al rediseño de los pistones de motores de combustible líquido.

En el mismo campo, Kroo et al. [41] describen el uso de métodos de diseño evolutivo en aplicaciones en aeronáutica a través de ejemplos aplicados al diseño de aviones supersónicos (figura 6 izq.). En la aeronáutica, los procesos de optimización cobran vital importancia pues, un pequeño cambio en la geometría del diseño, puede producir grandes deferencias en el flujo del aire a altas velocidades, y una pequeña variación en el peso estructural de un diseño, puede repercutir enormemente en su rendimiento operativo. Las complejas simulaciones necesarias para llevar a cabo las mediciones que permiten validar los nuevos diseños basados en todo un conjunto de hipótesis de comportamiento a menudo representan un tiempo de cómputo elevado, requiriendo en ocasiones la solución de grandes sistemas de ecuaciones diferenciales no lineales a veces trabajando con millones de vértices o soluciones estructurales con cientos de miles de grados de libertad. En esta materia, los algoritmos evolutivos son una solución válida que permiten encontrar muchas buenas aproximaciones y hacerlo además de forma paralela, rebajando con ello los tiempos de cálculo.

La adaptabilidad y la simplicidad de las técnicas evolutivas permiten su aplicación en estos casos obteniendo en muchos casos soluciones más eficientes que la optimización numérica directa y ahorrando un considerable esfuerzo computacional. En este caso concreto el algoritmo propuesto por los autores fue un algoritmo PCGA, un algoritmo que está estrechamente relacionado con las estrategias de evolución [42,43]. El algoritmo PCGA utiliza una población generalmente pequeña con codificación real para las variables. En PCGA un descendiente se crea como combinación lineal de sus padres, concretamente mediante interpolación entre dos diseños anteriores o extrapolación en una dirección dada a partir de estos. El mejor hijo sustituye al peor padre preservando el mejor diseño en la población y requiriendo solo comparaciones locales que pueden ser fácilmente paralelas.

De nuevo en el estudio del tratamiento de señales y el filtrado de ruido es interesante el trabajo de Andreas M. Chwatal de 2008 [48], que en este caso propone la utilización de algoritmos evolutivos para el análisis de los datos obtenidos de la sonda espacial europea CAROT, en concreto como alternativa a la descomposión de ondas de interferencia mediante el método de Fourier.

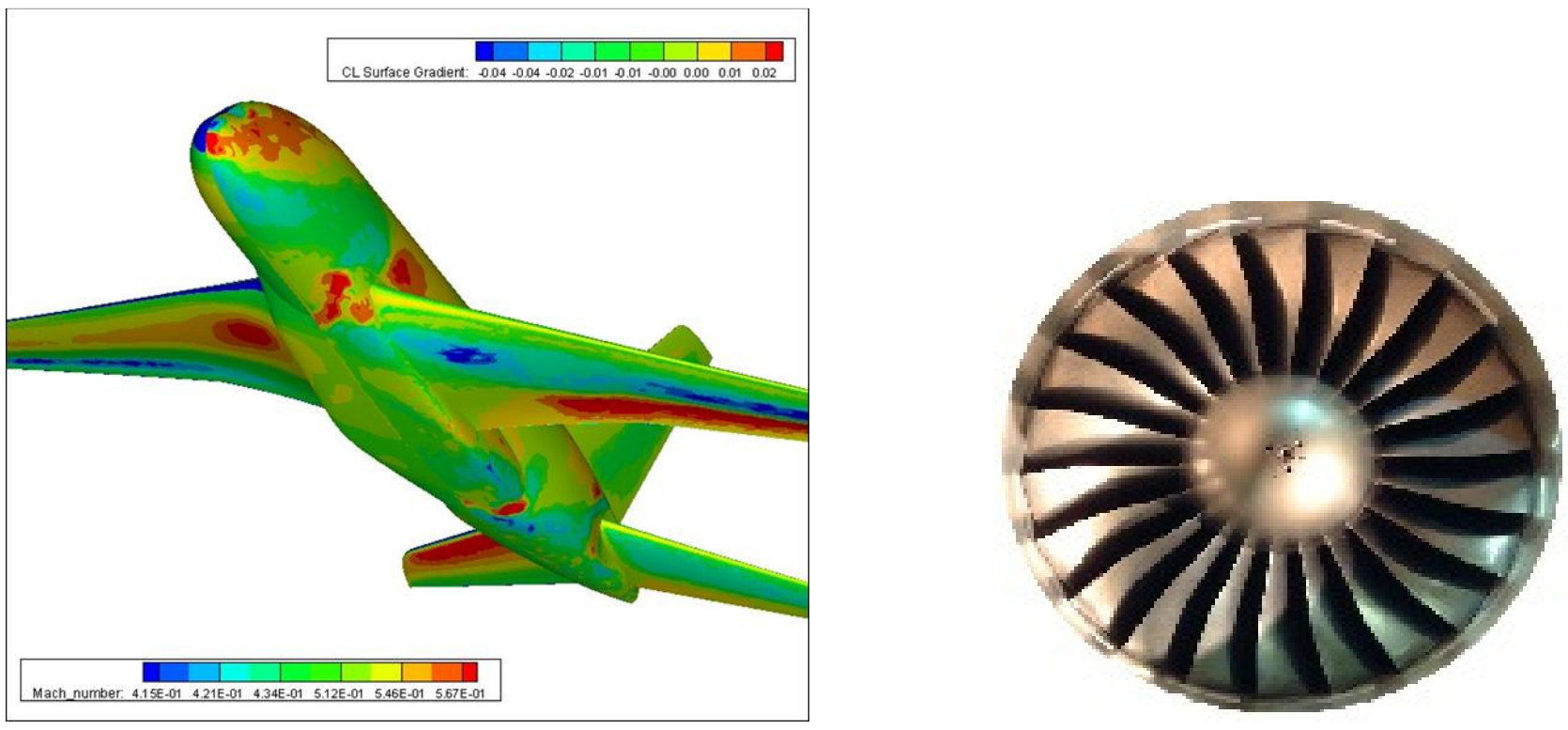

*Figura 6. Izq. Representación de la presión del aire sobre el fuselaje en un modelo de simulación (Kroo et al. 2004) Der. Primer plano del Nasa Rotor Transónico 67 de diseño evolutivo (Hansen et al. 2005).*

Más tarde, encontramos nuevas aplicaciones en ingeniería industrial en las publicaciones de Peniak y Cangelosi [49] quienes utilizaran un algoritmo genético para entrenar una red neuronal de capa oculta utilizada para enviar correcciones de rumbo en tiempo real a vehículos autónomos a partir de la información de las mediciones de los sensores a bordo, siendo un buen representante de una tendencia con grandes perspectivas en la actulidad, los algorítmos híbridos. Se han publicado numerosas propuestas para intentar mejorar el rendimiento de los EA. Una forma de conseguirlo es mediante la hibridación, como hemos explicado en las secciones anteriores las soluciones hibridas suelen darse con frecuencia, y entre estas soluciones, una fórmula de hibridación frecuente es la unión de un algoritmo evolutivo y un método analítico, por ejemplo el método de descenso del gradiente [50]. Los métodos de descenso de gradiente tienen una tasa de rapidez de convergencia muy elevada y la idea es ceder parte de esta propiedad al EA. La estrategia a seguir es normalmente la de utilizar el EA para descomponer el espacio de soluciones original en varias subregiones para luego dejar al algoritmo determinista la tarea de buscar cerca de esos sub-intervalos. Esta es la estrategia propuesta por Oyama et al. que utiliza un método híbrido basado en un EA y un método basado en gradiente, en este caso un resolutor por programación secuencial cuadrática (*SQP*).

En el último lustro, podemos citar como ejemplos representativos, como el trabajo de Dellnitz et al. [51] que implementa un algoritmo de programación genética con la finalidad de generar secuencias óptimas de comandos de control para satélites que se encuentran aparcados en órbitas periódicas; Kang et al. [52], los cuales hacen uso de un algoritmo de programación genética que les permite el estudio de las mejores rutas sobre mapas representados por grafos adaptativos obtenidos mediante el despliegue aleatorio de sensores; y Gosselin et al. [53] que publica un completo *review* sobre el uso de técnicas evolutivas en simulaciones de incendios y problemas de trasferencia de calor con frontera.

## 2.3 Algoritmos evolutivos en Ingeniería Civil.

En lo referente al campo concreto de la ingeniería civil, los primeros trabajos que encontramos donde hay un uso expreso de AE fueron desarrollados por Coello et al. [54, 54b] en 1997, y tenían como objetivo la optimización de del diseño de vigas de hormigón armado. Para tal fin, implementaron un AG simplificado (SGA) en diferentes sistemas de codificación, provisto de una tasa de mutación que se elegía de manera aleatoria en cada iteración (dentro del rango 0.0 - 0.9), un método de selección por torneo y un operador de cruce de 2 puntos. Los autores utilizan métodos empíricos para experimentar con diferentes sistemas de codificación y diferente número de variables de entrada entre las que consideraron el canto y ancho de la viga, el tipo de refuerzo, y el área de su armadura inferior. El resultado final fue la optimización del diseño de vigas de hormigón armado sometidas a un conjunto específico de restricciones, teniendo en cuenta factores como el coste de los materiales (hormigón, acero, encofrado, etc...), o el coste del proceso de producción. Los resultados obtenidos fueron comparados con los presentados algunos años antes por Chakrabarty, utilizando programación geométrica e involucrando un menor número de restricciones en el modelo. En esta ocasión, el algoritmo fue una variante de SGA (descrito con anterioridad) que representaba una clara alternativa al proceso tradicional de obtención de estos valores mediante métodos analíticos obteniendo resultados más realistas para los parámetros del modelo habida cuenta del mayor número de restricciones aplicadas y una convergencia más rápida. De manera concurrente Rafiq y Southcombe [55] aplican SGA al problema de optimizar el armado de pilares de hormigón sometidos a esfuerzo axil y flexión esviada. En este último caso, la geometría del pilar no es una variable del problema sino un parámetro de entrada.

En 1998, Kuomousis et al. [57] utilizan un algoritmo genético para decidir el número y la ubicación óptimos de barras de armado en secciones de hormigón en edificios de varias plantas. Y un año más tarde, Botello et al. [56] presentan un algoritmo hibrido que utiliza un AG con codificación real que integra fases de búsqueda local mediante el método de recocido simulado; en esta ocasión, el objetivo era, una vez más, la optimización del diseño de estructuras bajo diferentes condiciones de funcionamiento; por un lado, se estudió el diseño óptimo de pórticos de naves industriales sometidos a cargas laterales, y por otro, se analizó la respuesta de elementos estructurales de puentes sometidos a efectos de compresión.

Cuatro años más tarde, Chau y Albermani [58] crean una nueva aplicación informática para el diseño optimizado de depósitos rectangulares de hormigón armado mediante el uso del programa comercial abaqus. Los resultados fueron muy buenos, a pesar de llevar a cabo un análisis relativamente simple del problema basado en la consideración únicamente de tres variables geométricas: canto de las losa de hormigón, cuantía de la armadura y separación entre barras, repitiendo los mismos valores de dichas variables para todos los elementos del depósito. También en 2003, Leps y Sejnoha [59] aplican un algoritmo de 21 variables que combina SA (recocido simulado) y un AG para la optimización de una viga continua simétrica de hormigón de dos vanos con armadura de cortante y de flexión. La técnica en cuestión, conocida como *Augmented Simulted Annealing*, es similar a un SGA en el que: a) se trabaja con una población de soluciones en lugar de con una única solución y b) las nuevas soluciones se obtienen a partir de las existentes mediante la aplicación de los operadores genéticos y no mediante el concepto de movimiento (como se realiza en el SGA).

Mas recientemente, Lee y Ahn, proponen el optimizado de pórticos planos de hormigón armado [60] empleando también un algoritmo SGA al que incorporan una estrategia elitista, y, Fairnairn et al. [61], diseñan un procedimiento para optimizar la construcción de estructuras de hormigón en masa utilizando un SGA con codificación binaria y política elitista, y, tomando como variables de entrada, el tipo y coste de materiales, su resistencia térmica o la distribución interna de cada capa de componente; ello permitió estimar diferentes características de funcionamiento de la estructura como su permeabilidad, su distribución térmica, o su estado tensional. El algoritmo propuesto se enfrenta a los resultados obtenidos mediante la aplicación de un modelo termo-quimio-mecánico de Coussy al caso concreto de la construcción de una presa para una central hidroeléctrica. De forma muy similar, Lim et al. [62] estudian el balance óptimo de proporciones en la fabricación de hormigón de alta resistencia. En 2005, Sobolev et al. [63], utilizan un algoritmo genético para optimizar las proporciones de materiales compuestos de cemento y hormigón tomando en cuenta de su repercusión en las propiedades físicas del material (densidad, viscosidad, etc...) así como en el rendimiento (fortaleza, durabilidad, elasticidad, etc...) de la mezcla finalmente obtenida. El algoritmo utilizado es el algoritmo SAGA (Seft-Adaptative Genetic Algorithm) propuesto por Amirjanov un año antes [64]. Este algoritmo se basa en un algoritmo genético de evolución diferencial con aprendizaje adaptativo, donde los mejores individuos de la población actúan como focos atractores indicando al algoritmo las zonas prometedoras donde debe concentrarse la búsqueda.

En otro ámbito de aplicabilidad, Prendes et al. estúdian el uso de algoritmos genéticos en la evaluación del diseño de edificios de estructura metálica [65] centrando los objetivos de diseño en la resistencia a cargas estáticas, la seguridad y el coste de los materiales. El algoritmo utilizado es un AG de baja especificabilidad utilizando codificación real, un tamaño de población de entre 60 y 100 individuos y tasa de mutación aleatoria con valores entre 0.01 y 0.03, obteniendo mejoras moderadas de en torno al 10% en la estabilidad de la estructura. También cabe destacar el trabajo desarrollado por Dhyanjyoti et al. en 2006 [66], donde los autores se valieron de un algoritmo genético para llevar a cabo predicciones sobre el comportamiento elasto-plástico que experimentan los materiales de estructura cristalina cuando se les somete a cargas cíclicas. El artículo estudia además la resistencia de los distintos materiales frente a ratios de deformación constantes, buscando un aumentando la seguridad del diseño y evitando posibles procesos de fatiga. En este caso, la fuente consultada no proporciona mayor detalle acerca del tipo de AG empleado.

En los últimos 5 años encontramos las publicaciones de Nehdi et al. [67] en el que se presenta un modelo para medir la resistencia a esfuerzo a cortante en vigas de hormigón con refuerzo laminado externo de materiales FRP. Para parametrizar su modelo utilizan un AG simple que parte de un conjunto de 212 datos obtenidos de forma experimental obteniendo un ajuste mejor que el presentado hasta entonces por los modelos y normativas existentes (ACI 440,

*EuroCode2*, *Matthys model* y *Colotty model*). Los mismos autores [68] ya habían estudiado la aplicación de AG en la estimación de la resistencia a cortante en vigas de hormigón igualmente reforzadas internamente con FRP. En ambos casos, el algoritmo implementado utilizaba codificación real y un método de selección estocástica con una tasa de mutación baja de 0,005. La capacidad de obtener varias soluciones sub-optimas concurrentemente permitió a los autores experimentar con diferentes materiales y disposiciones.

# Capítulo 3.

# El problema de la estimación de parámetro de degradación del hormigón solicitado a cortante.

La teoría de campo de compresiones (CFT) comprende un conjunto de hipótesis de comportamiento que permiten, en el caso del hormigón armado, realizar predicciones sobre su deformación y evolución de ruptura del material cuando este es sometido a carga y a esfuerzo cortante como ocurre en las vigas y los pilares de hormigon. Hernández-Díaz en su tesis doctoral [101], lleva a cabo una revisión teórica de dichas hipótesis, proponiendo una formulación actualizada del modelo más cercana a como ocurre el deterioro del hormigón armado (en función de la deformación y del comportamiento de tenso-rigidez del acero) y que integra algunos de los últimos avances y postulados en el campo (Vecchio 1986; Collins 1991; Bentz 2000) siendo por tanto más coherente con el comportamiento de la armadura y con el fenómeno real de adherencia entre el hormigón y el acero, antes, y una vez se produce el agrietamiento de la estructura. En este contexto, Hernández-Díaz utiliza la estimación de un parámetro *kappa* que relaciona la deformación con la consistencia y degradación de la viga, lo que le permite realizar una notable simplificación en las ecuaciones, teniendo por el contrario el inconveniente, de ser un parámetro, que requiere un costoso ajuste experimental ya que resulta de la resolubilidad del sistema no lineal de ecuaciones que conforman las ecuaciones directoras al modelo de campo de compresiones, más las ecuaciones de equilibrio, y del modelo constitutivo del acero. Apoyándonos en sus resultados, tratamos de obtener una mejor estimación para el parámetro kappa utilizando algoritmos de optimización evolutiva que tratamos mejoren los resultados obtenidos por la estimación algebraica llevada a cabo por los autores.

### 3.1 Descripción del problema. Formalización. Restricciones

Durante los últimos años han aparecido varias teorías que tienen como objetivo estudiar la respuesta de elementos de hormigón armado sometidos a esfuerzos cortantes. La pieza fundamental dentro de este marco teórico son las <u>teorías de campo de compresiones</u>.

Son varias las hipótesis que se han ido incorporando por distintos autores a la predicción de deformaciones. Hernandez-Diaz en [101] realiza una revisión actualizada de las teorías de campo de compresiones integrando la denominada hipótesis de Wagner [112] (según la cual, la dirección del campo de tensiones coincide con la del campo principal de deformaciones) y el denominado "parámetro de degradación del hormigón a cortante" introducido por Gil Martín et al [113] que permite que el área efectiva de hormigón sometida a tracción ($A_c$) varíe conforme lo hace el agrietamiento. Dicho parámetro, resulta clave en las ecuaciones constitutivas del acero, siendo por tanto, merecedor de un estudio en profundidad.

El estudio de la fisurización del hormigón estructural es un proceso especialmente complejo por tratarse de un sistema vivo, en el que, conforme avanza la degradación, la fisura inicial se propaga y se ramifica en nuevas grietas en cuyo cálculo, una parte de las ecuaciones dependen de los ángulos y profundidad de las fisuras. La teoría unificada de compresiones ayuda a predecir la forma y severidad del agrietado midiendo la energía de deformación que soporta la armadura (tanto trasversal como longitudinal), y el hormigón, sujeto a ciertas condiciones de equilibrio manteniendo la coherencia con las relaciones de tensión-deformación entre la armadura y el hormigón agrietado. El estudio unificado de estos dos aspectos, nos permite tener una visión completa de como la rigidez tensional del hormigón (la contribución a tracción del hormigón) afecta a la respuesta tenso-deformacional del acero y por tanto al material compuesto.

El modelo descrito por Hernández-Díaz está formado por un sistema de once ecuaciones no lineales [101] que pasaremos a enumerar a continuación. Este trabajo, **no tiene por finalidad la justificación de dichas ecuaciones**, por lo que procedemos simplemente a enumerarlas brevemente agrupadas según su naturaleza:

3 condiciones de equilibrio

$$\sigma_1 + \sigma_2 = (\tan\theta + \cot\theta)\frac{V}{z \cdot b_w} \quad \text{(Ec. 1)}$$

$$A_{s;t}\sigma_{s;t} = (\sigma_2 \sin^2\theta - \sigma_1 \cos^2\theta) \cdot b_w s \quad \text{(Ec. 2)}$$

$$A_{s;x1}\sigma_{s;x1} + A_{s;x2}\sigma_{s;x2} + \sigma_1 b_w z = \frac{V}{\tan\theta} \quad \text{(Ec. 3)}$$

donde $\theta$ es el ángulo de inclinación de las tensiones de compresión diagonal, $V$ es el esfuerzo cortante, $\sigma_1$ es la tensión de tracción principal para el hormigón, , $\sigma_2$ es el esfuerzo de compresión principal en el hormigón, $b_w$ es el ancho efectivo de la sección a cortante, *s* es la separación entre estribos, y, $A_{s;x1}$, $A_{s;x2}$ y $A_{s;t}$ son las áreas de sección transversal de las barras inferiores longitudinales, barras superiores longitudinales y de estribos, respectivamente.

2 condiciones de compatibilidad

$$\tan^2\theta = \frac{\varepsilon_x - \varepsilon_2}{\varepsilon_t - \varepsilon_2} = \frac{\varepsilon_1 - \varepsilon_t}{\varepsilon_1 - \varepsilon_x} \quad \text{(Ec. 4)}$$

$$\varepsilon_1 = \varepsilon_x + \varepsilon_t - \varepsilon_2 \quad \text{(Ec. 5)}$$

donde $\varepsilon_t$ es la deformación transversal media, $\varepsilon_1$ es la deformación principal por tracción, $\varepsilon_x$ es la deformación longitudinal media y $\varepsilon_2$ es la tensión principal de compresión.

2 ecuaciones para el comportamiento del hormigón a compresión:

$$f_{2,max} = f_c \cdot \min\{1\ ,\ (0.8+170\varepsilon_1)^{-1}\} \quad \text{(Ec. 6)}$$

$$\sigma_2 = f_{2,max}\left(2\frac{\varepsilon_2}{\varepsilon_c} - \left(\frac{\varepsilon_2}{\varepsilon_c}\right)^2\right) \quad \text{(Ec. 7)}$$

Donde $f_c$ es resistencia a compresión del hormigón, $\varepsilon_c$ es la deformación correspondiente a $f_c$, y $f_{2,max}$ es una cota máxima para el estrés a compresión.

1 ecuación de comportamiento del hormigón a tracción

$$\sigma_1(\varepsilon_1) = \begin{cases} E_c\varepsilon_1 & \text{for} \quad \varepsilon_1 \le \varepsilon_{ct} \\ \dfrac{\alpha \cdot f_{ct}}{1+\sqrt{500\cdot\varepsilon_1}} & \text{for} \quad \varepsilon_1 > \varepsilon_{ct} \end{cases} \quad \text{(Ec. 8)}$$

Donde $E_c$ es el módulo de la elasticidad del hormigón y $\varepsilon_{ct}$ es la deformación correspondiente a la resistencia a la tracción del hormigón (*fct*).

Y por último, 3 ecuaciones correspondientes al modelo constitutivo del acero de las barras de refuerzo (una para la armadura longitudinal inferior, otra para la armadura longitudinal superior, y una para la armadura transversal)

$$\sigma_{s;x1} = \begin{cases} E_s\varepsilon_x, & \varepsilon_x \le \varepsilon_{max;x1} \\ f_{y;x1} - \dfrac{\kappa A_{c;x1}}{A_{s;x1}} \dfrac{f_{ct}}{1+\sqrt{500\varepsilon_x}}, & \varepsilon_x > \varepsilon_{max;x1} \end{cases}$$

where:

$$\varepsilon_{\max;x1} = \frac{f_{y;x1}}{E_s} - \frac{\dfrac{f_{ct}}{1+\sqrt{500\,\varepsilon_{\max;x1}}}}{E_s A_{s;x1}} \kappa A_{c;x1} \qquad \text{(Ec. 9)}$$

$$\sigma_{s;x2} = \begin{cases} E_s\varepsilon_x, & \varepsilon_x \le \varepsilon_{max;x2} \\ f_{y;x2} - \dfrac{\kappa A_{c;x2}}{A_{s;x2}} \dfrac{f_{ct}}{1+\sqrt{500\varepsilon_x}}, & \varepsilon_x > \varepsilon_{max;x2} \end{cases}$$

where:

$$\varepsilon_{\max;x2} = \frac{f_{y;x2}}{E_s} - \frac{\dfrac{f_{ct}}{1+\sqrt{500\,\varepsilon_{\max;x2}}}}{E_s A_{s;x2}} \kappa A_{c;x2} \qquad \text{(Ec. 10)}$$

$$\sigma_{s;t} = \begin{cases} E_s\varepsilon_x, & \varepsilon_t \le \varepsilon_{max;t} \\ f_{y;t} - \dfrac{\kappa A_{c;t}}{A_{s;t}} \dfrac{f_{ct}}{1+\sqrt{500\varepsilon_x}}, & \varepsilon_t > \varepsilon_{max;t} \end{cases}$$

where:

$$\varepsilon_{\max;t} = \frac{f_{y;t}}{E_s} - \frac{\dfrac{f_{ct}}{1+\sqrt{500\,\varepsilon_{\max;t}}}}{E_s A_{s;t}} \kappa A_{c;t} \qquad \text{(Ec. 11)}$$

En las ecuaciones (9), (10) y (11), los subíndices $x_1$, $x_2$ y $t$ hacen referencia al refuerzo longitudinal inferior, superior y transversal. En el caso más general, cada tipo de armadura presenta diferente límite elástico ($f_y$), diferente cuantía ($A_s$), diferente coeficiente de adherencia hormigón-acero ($\alpha_i$) y diferente área efectiva de hormigón a tracción ($A_c$), razón por la cual la deformación aparente de cedencia ($\varepsilon_{max}$) definida por la teoría unificada de compresiones variará de un tipo de armadura a otro.

En total tenemos **once ecuaciones y once incógnitas** $\theta$, $\varepsilon_x$, $\varepsilon_t$, $\varepsilon_1$, $\varepsilon_2$, $\sigma_1$, $\sigma_2$,$\sigma_{s;x1}$, $\sigma_{s;x2}$, $\sigma_{s;t}$, $f_{2,max}$ que pueden ser resueltas para un determinado valor de $V$ (cortante de agotamiento) y del parámetro de degradación de hormigón ($\kappa$).

Mediante sustitución y explicitación algebraica es posible reducir el sistema a un <u>sistema de dos ecuaciones</u> (que denotaremos $f$ y $g$) y dos incógnitas ($\varepsilon_1$, $\theta$); que tomarán distinta forma en función del régimen de comportamiento (elástico o plástico) que tomamos como supuesto para cada tipo de armadura. La función que define la primera ecuación se denota ***f*** y representa el equilibrio de las armaduras longitudinales superior e inferior; la función que define la segunda ecuación se denota ***g*** y corresponde al equilibrio de la armadura transversal.

Como hemos dicho anteriormente, las expresiones de estas funciones dependen de la hipótesis de comportamiento asumida (proceso recursivo), que posteriormente deberemos verificar en base a la solución general del sistema. En ambas ecuaciones, se denota mediante el subíndice "E" la hipótesis correspondiente al régimen elástico, y por "P" la hipótesis correspondiente al régimen plástico, de forma que para un espécimen concreto indicaremos una terna de tres letras "E" o "P" que se corresponden a la hipótesis asumida para las armaduras longitudinales superior e inferior, y trasversal.

$$f_{EE}\ [\theta,\varepsilon_1] = \frac{b_w f_{ct} z}{1+\sqrt{500\varepsilon_1}} - V_u \cot\theta + \left(A_{s;x1} + A_{s;x2}\right) E_s \Omega$$ (Ec. 12)

$$f_{EP}[\theta,\varepsilon_1] = \frac{b_w f_{ct} z}{1+\sqrt{500\varepsilon_1}} - V_u \cot\theta + A_{s;x1} E_s \Omega + A_{s;x2} f_{y;x2} - \frac{\kappa(\varepsilon_1) A_{c;x2} f_{ct}}{1+\sqrt{500\Omega}}$$ (Ec. 13)

$$f_{PP}[\theta,\varepsilon_1] = \frac{b_w f_{ct} z}{1+\sqrt{500\varepsilon_1}} - V_u \cot\theta + A_{s;x1} f_{y;x1} + A_{s;x2} f_{y;x2} - \frac{f_{ct}\kappa(\varepsilon_1)\left(A_{c;x1} + A_{c;x2}\right)}{1+\sqrt{500\Omega}}$$ (Ec. 14)

$$g_E\ [\theta,\varepsilon_1] = \frac{b_w f_{ct} s}{1+\sqrt{500\varepsilon_1}} - \frac{V_u \cdot s}{z} \tan\theta + A_{s;t} E_s \Psi$$ (Ec. 15)

$$g_P\ [\theta,\varepsilon_1] = \frac{b_w f_{ct} s}{1+\sqrt{500\varepsilon_1}} - \frac{V_u \cdot s}{z} \tan\theta + A_{s;t} f_{y;t} - \frac{\kappa(\varepsilon_1) A_{c;t} f_{ct}}{1+\sqrt{500\Psi}}$$ (Ec. 16)

Donde los factores *Ω* y *Ψ* son coeficientes asociados al tipo de armadura, y cuyas expresiones vienen dadas por:

$$\Omega = \frac{\varepsilon_1 \tan^2\theta + \lambda\varepsilon_c}{1+\tan^2\theta}$$ (behavior coefficient for the longitudinal reinforcement)

$$\Psi = \frac{\varepsilon_1 + \lambda\varepsilon_c \tan^2\theta}{1+\tan^2\theta}$$ (behavior coefficient for the transverse reinforcement) (Ec. 17a, 17b)

Estando sujetos a:

$$\lambda = 1 - \sqrt{1 - \frac{\dfrac{V(\cot\theta + \tan\theta)}{b_w z} - \dfrac{f_{ct}}{1+\sqrt{500\varepsilon_1}}}{\min\left\{f_c, \dfrac{f_c}{0.8+170\varepsilon_1}\right\}}}$$ (Ec. 18)

Una importante propiedad de este sistema reducido es que permite su representación gráfica, resultando así la denominada curva de solubilidad del sistema [105]. Esta curva se compone de un conjunto de puntos (*κ, ε1*) para los que el sistema reducido tiene una solución real.

| Author | Beam | $V_u$ | $\sigma_{st,exp}$ |
|---|---|---|---|
| Ahmad et al. (1995) [2] | NHW-3b | 122779 | 324.14 |
| Palaskas et al. (1981) [16] | A50 | 115426 | 492.41 |
| | A75 | 142203 | 420 |
| | C50 | 134107 | 507.59 |
| | C75 | 137977 | 444 |
| Kong et al. (1997) [13] | S1-4 | 277900 | 450 |
| | S2-3 | 253300 | 265.96 |
| | S4-6 | 202900 | 300 |
| | S7-4 | 273600 | 375 |
| Leonhardt et al. (1962) [14] | ET3 | 126248 | 313.92 |
| Moayer et al. (1974)[18] | P20 | 120096 | 310.28 |
| Soerensen (1974) [19] | T22 | 128987 | 399.27 |
| Bernhardt et al. (1986) [4] | S8 A | 125720 | 427 |
| Cladera et al. (2002) [6] | H 50/4 | 246340 | 540 |
| | H 75/4 | 255230 | 530 |
| | H 100/4 | 266530 | 540 |
| Levi et al. (1988) [15] | RC 30 A1 | 676000 | 480 |
| | RC 30 A2 | 688000 | 480 |
| | RC 60 A1 | 990000 | 480 |
| | RC 60 A2 | 938000 | 480 |
| | RC 60 B1 | 1181000 | 480 |
| | RC 60 B2 | 1239000 | 480 |
| | RC 70 B1 | 1330000 | 480 |
| Rosenbusch et al. (1999) [18] | MHB 2.5-25 | 98801 | 267.30 |
| Regan (1971) [17] | T3 | 105000 | 270 |
| | T4 | 110000 | 270 |
| | T6 | 205000 | 270 |
| | T7 | 109000 | 280 |
| | T8 | 124000 | 280 |
| | T9 | 154000 | 280 |
| | T13 | 90000 | 270 |
| | T15 | 104000 | 270 |
| | T17 | 134000 | 280 |
| | T19 | 106000 | 270 |
| | T20 | 138000 | 280 |
| | T26 | 179000 | 280 |
| | T32 | 216000 | 270 |
| | T34 | 112000 | 270 |
| | T35 | 115000 | 270 |
| | T37 | 209000 | 270 |
| | T38 | 238000 | 270 |

*Tabla 1. Listado de los 81 especímenes recopilados por Reineck et al. (2000).*

Ya establecida la formulación matemática del modelo, se utilizan los datos experimentales recogidos por Reineck et al. [106] (consistentes en una selección de un total de 81 vigas de hormigón armado ensayadas hasta su agotamiento a cortante) recogidos en la tabla 1 y correspondientes a diferentes campañas experimentales a fin de realizar la aproximación al parámetro buscado según un modelo teórico de ajuste. Para cada una de las vigas se midieron directamente, entre otros parámetros, los valores de cortante ($V_{exp}$) y tensión resultante ($\sigma_{st,exp}$) en la sección de agotamiento.

Una vez sustituidos los parámetros calculados en las ecuaciones reducidas del sistema, se obtiene, para cada una de las hipótesis de comportamiento ya definidas, un conjunto de dos funciones {*f,g*} parametrizadas en *V* y *κ*.

A medida que aumenta el valor de *κ*, se produce una evolución en la posición relativa de las curvas de ceros de las funciones *f* y *g*. Se dice que una hipótesis de comportamiento es soluble cuando ambas funciones tengan al menos un punto de intersección para algún valor de *κ*. Así mismo, diremos que una sección de hormigón armado es consistente a efectos del ajuste del parámetro κ cuando verifique para al menos una de las hipótesis de comportamiento analizadas bajo el valor experimental de V.

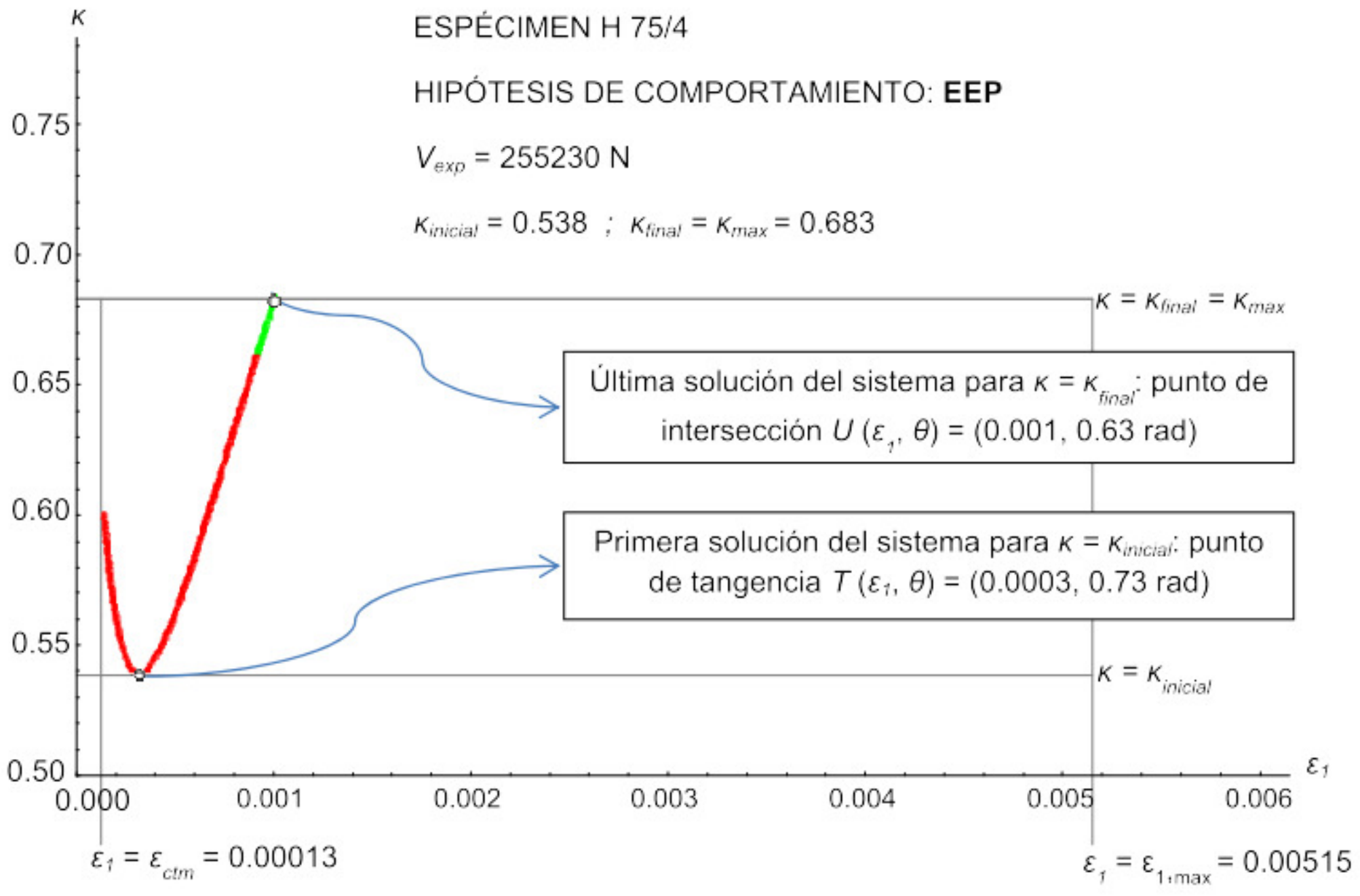

*Figura 7. Curva de solubilidad del espécimen H 75/4 correspondiente a la hipótesis EEP (Hernández-Díaz 2013).*

En la figura 7, se representa la curva de puntos (*κ, ε1*) del espécimen H 75/4 para la hipótesis EEP. De ellos solo una pequeña parte verifica la hipótesis de comportamiento asumida (porción en verde). A efectos del ajuste del parámetro κ, necesitamos especímenes que, además de solubles, sean consistentes. La función de parámetro buscado será aquel que logre intersectar el mayor número de segmentos consistentes para los **81 especímenes** tomados en el experimento. Por simplicidad, se considerarán funciones candidatas de forma polinómica $\{\kappa(\varepsilon_1) = a\varepsilon_1^3 + b\varepsilon_1^2 + c\varepsilon_1 + d;\ \ a,b,c,d \in \mathrm{R}\}$.

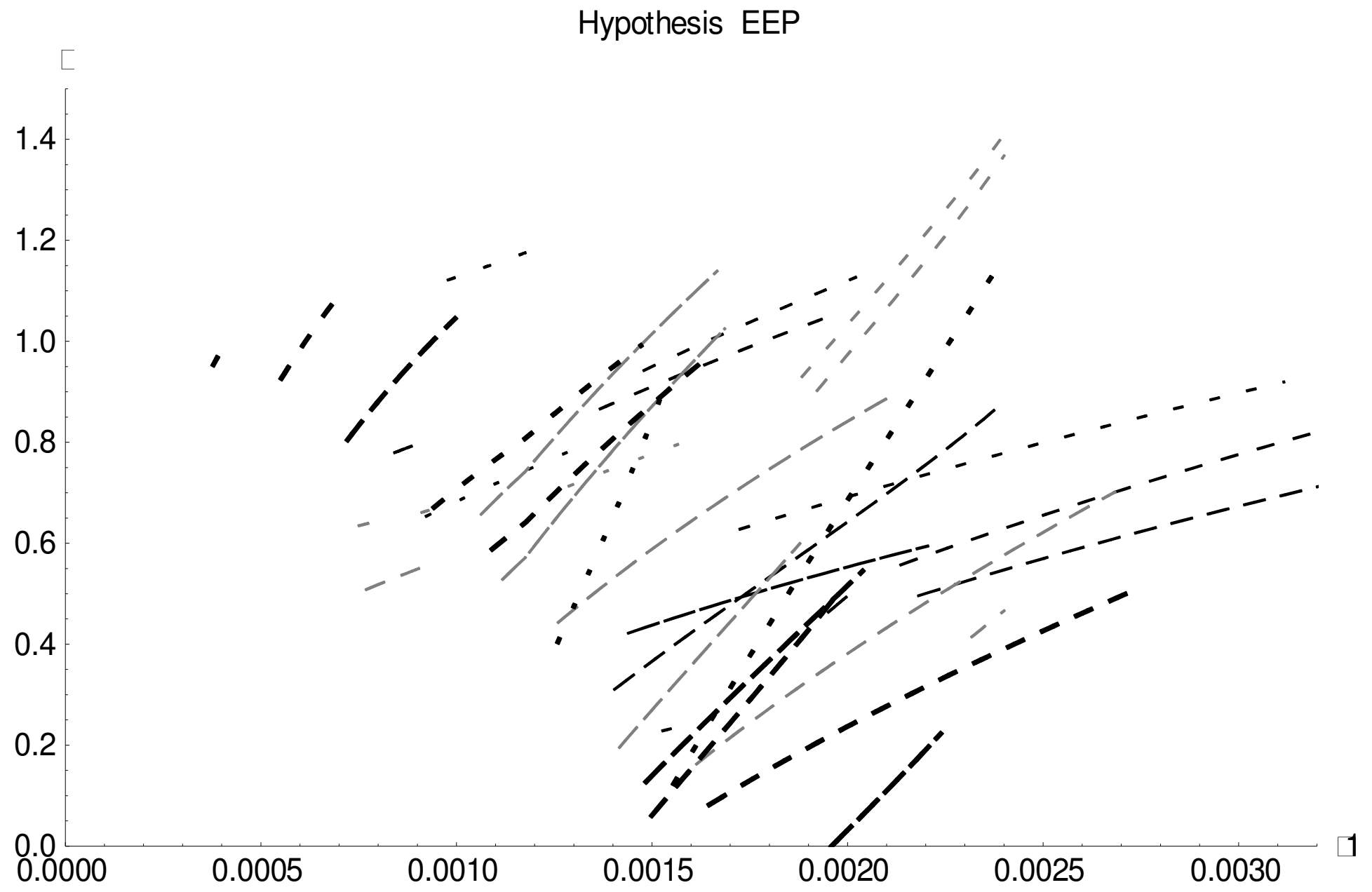

*Figura 8. Representación gráfica de todos los segmentos consistentes de la curvas de solubilidad de cada espécimen para la hipótesis Elástico-Elástico-Plástico.*

## 3.2 Propuestas de resolución mediante técnicas evolutivas

Hernández et al. proponen en [107] resolver el problema de la estimación del parámetro de degradación del hormigón armado mediante algoritmo de computación evolutiva derivado del *Little Genetic Algorithm* (LGA) propuesto por Coley et al. en el año 2000 para usos académicos e industriales [108]. En pocas palabras el método es una simplificación del esquema general de los algoritmos genéticos donde una población inicial de individuos evoluciona a través de dos mecanismos. Un mecanismo de selección basado en ruleta (proporcional al fitness) y un operador de cruce uniforme de 1 punto con una probabilidad de mutación aleatoria entre 0 y 1. La política de reemplazo realiza el reemplazo completo de la población en cada generación llevando a cabo una estrategia elitista (conservación del más apto individuo). Cada individuo en la población lleva asociado un *array R* de ***4*l* bits** (genoma) que codifica los cuatro coeficientes de una función cúbica $\kappa R(\varepsilon_1)$. El array R puede dividirse por tanto en cuatro secuencias de l bits *R1, R2, R3* y *R4*, que resultan de dividir los números enteros cuyas codificaciones en codificación gray son *R1, R2, R3* y *R4*, respectivamente, por *2l*, y, finalmente, transformar los números obtenidos en valores binarios.

Para calcular la función de aptitud (*fitness*) se hace uso de la siguiente función auxiliar $Fit(R)\in[0,1]$.

$$F(R)=\frac{\sum_{X\ \text{consistent}}\left(\sigma_{st,X,\exp}-\sigma_{st,X,R}\right)^2}{\#\{X;\ X\ \text{is consistent}\}} \qquad \text{(Ec. 19)}$$

Donde $\sigma_{st,X,exp}$ es la tensión experimental en cercos en el punto de ruptura ($\sigma_{st,exp}$) cuyo valor conocemos (tabla 1) y $\sigma_{st,X,R}$ es la tensión teórica obtenida por el modelo tomando la ecuación *κ=κ R(ε₁)* una vez deshechos los cambios de variable. Como podemos ver, la diferencia cuadrada ($\sigma st_{,X,exp}$ - $\sigma st_{X,R}$) ^ 2 actuará de error cuadrático para la obtención del fitness. En caso de que este valor no pueda calcularse se devolverá un valor de fitness penalizado lo suficientemente alto como para que se descarte la solución.

Para llevar a término la ejecución del algoritmo se conforma un *cluster* formado por 64 procesadores AMD *Opteron* de doble núcleo interconectados mediante red *GigaNet*. Aun así, el tiempo de ejecución reportado por los autores fue de algo más de 285 horas terminando su ejecución al alcanzar un límite de ejecuciones impuesto a priori partiendo de una población inicial donde ya ha sido incluido una solución aceptable (obtenida por los autores mediante regresión numérica).

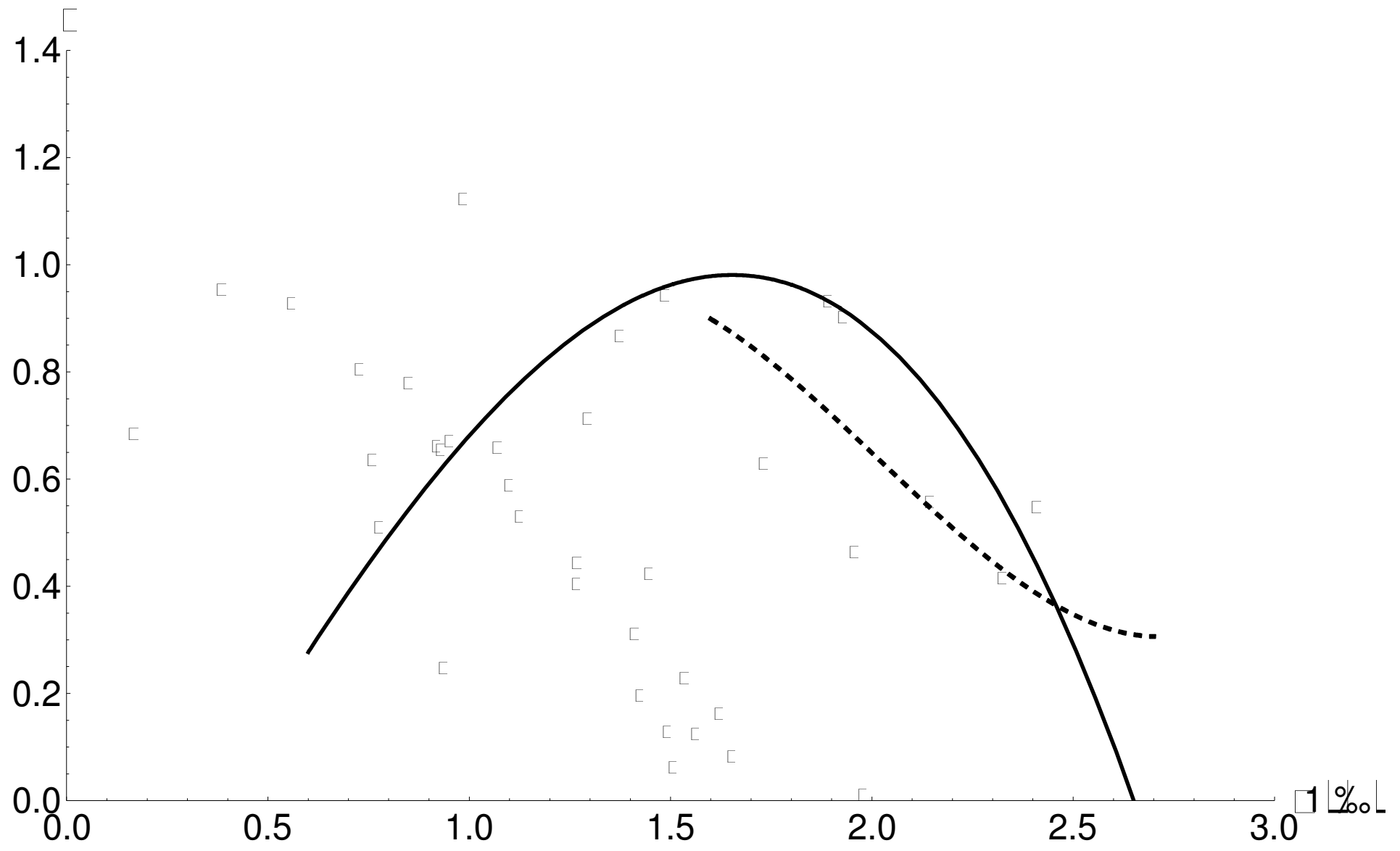

*Figura 9. Función de aproximación cúbica κ(ε1)=aε13+bε12+cε1+d con a=-0.1713, b=0.0346, c=1.2902 y d=-0.4725, obtenida como el individuo más apto después de 200 generaciones.*

Como consecuencia de haber añadido soluciones artesanales en la población inicial y la alta dispersión de los puntos (ε1, κopt) de las muestras de la base de datos, la aptitud máxima no aumenta mucho más allá del valor inicial máximo de 16,63 % en las primeras generaciones. La figura 9 muestra la representación gráfica del candidato más apto encontrado después de 200 generaciones. Los coeficientes correspondientes a esta solución para la función cúbica $\kappa_R(\varepsilon_1) = a\varepsilon_1^3 + b\varepsilon_1^2 + c\varepsilon_1 + d$ son a=-0.1713, b=0.0346, c=1.2902 y d=-0.4725, y el fitness conseguido 28.40% (figura 10).

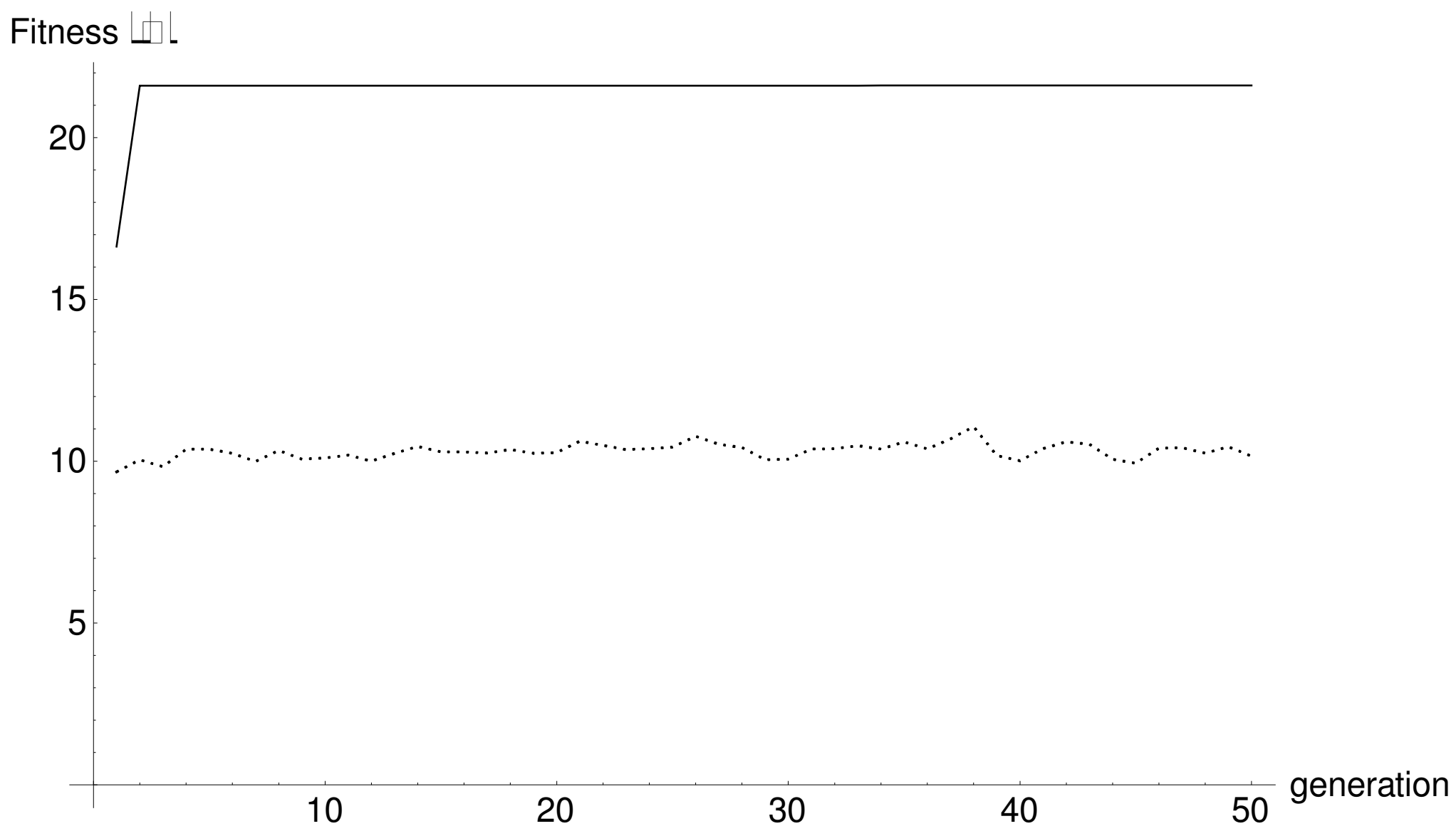

*Figura 10. Mejor fitness y el fitness medio para cada generación en las 50 primeras vueltas del algoritmo.*

La figura 10 muestra la evolución del fitness conseguido a lo largo de la ejecución del algoritmo. En ella podemos observar como existe una situación de estancamiento donde no existe una evolución real de la aptitud de los individuos de la población y donde la búsqueda ha degenerado en un proceso meramente aleatorio. La causa de este fenómeno posiblemente no puede ser explicada por un único factor sino por una conjunción de varios factores. El más visible de todos ellos es que la búsqueda se realiza en un vasto espacio de soluciones continuas (el espacio R de 4 dimensiones) y sin embargo el algoritmo empleado es enormemente genérico y no especializado en espacios continuos difíciles. También el hecho de introducir desde el principio soluciones aceptables, en combinación con la política elitista va a desembocar en un empobrecimiento muy rápido de la información genética contenida en la población. Tampoco ayuda la penalización introducida, que en la ejecución considerada era muy alta en comparación con la cota superior de soluciones posibles ($10^6$). Esta práctica se vuelve menos recomendable si consideramos que debido a la naturaleza del problema, puede darse el caso de que dos valores muy cercanos en el espacio de soluciones no comportan solubilidad, es decir, uno puede ser resoluble y otro no, y, al estar la función fitness indefinida en el caso de la no solubilidad, esta circunstancia lleva al extremo de que un avance en la dirección correcta pueda ser penalizado y por tanto descartado de la búsqueda. Podríamos considerar estos puntos no definidos como ruido en el espacio de soluciones. En este escenario, escapar de los mínimos locales (que serán abundantes) es crucial para poder alcanzar un mínimo global. Si esto es así, entonces la política elitista tendría que ser llevada a cabo con sumo cuidado ya que puede dificultar el que el algoritmo pueda escapar de estos puntos trampa.

Además de estas consideraciones, existen otras de menor gravedad que también deberían ser tenidas en cuenta. Por ejemplo, la forma de cruce (cruce binario uniforme en un punto) no necesariamente es coherente ni con la naturaleza de la solución, ni con la del dato mismo. Es decir, al partir una cadena binaria por la mitad podemos de hecho estar rompiendo un número real el dos mitades (una con la parte entera y otra con la parte decimal) o incluso, estar partiendo por la mitad una de las dos partes. Además cruzar los coeficientes *R1* y *R2* de una solución con los coeficientes *R3* y *R4* de otra solución podría no tener sentido. Así mismo, al realizarse la mutación al nivel binario, podemos encontrar que la matación de un bit produzca un salto enorme o diminuto en el valor de un coeficiente según el bit afectado.

### 3.2.1 Covariance Matrix Adaptation Evolution Strategy (CMA-ES)

Para tratar de solventar los problemas descritos en las secciones anteriores, se propone la implementación del problema utilizando un algoritmo evolutivo CMA-ES. Este algoritmo, perteneciente a la familia de las estrategias de evolución (ES), que en sus inicios fue concebido como método de búsqueda local [109] ofrece también un buen rendimiento y calidad en la solución cuando es aplicado en la búsqueda sobre espacios reales cuando el paisaje de búsqueda es complejo o contiene gran cantidad de ruido [110, 111].

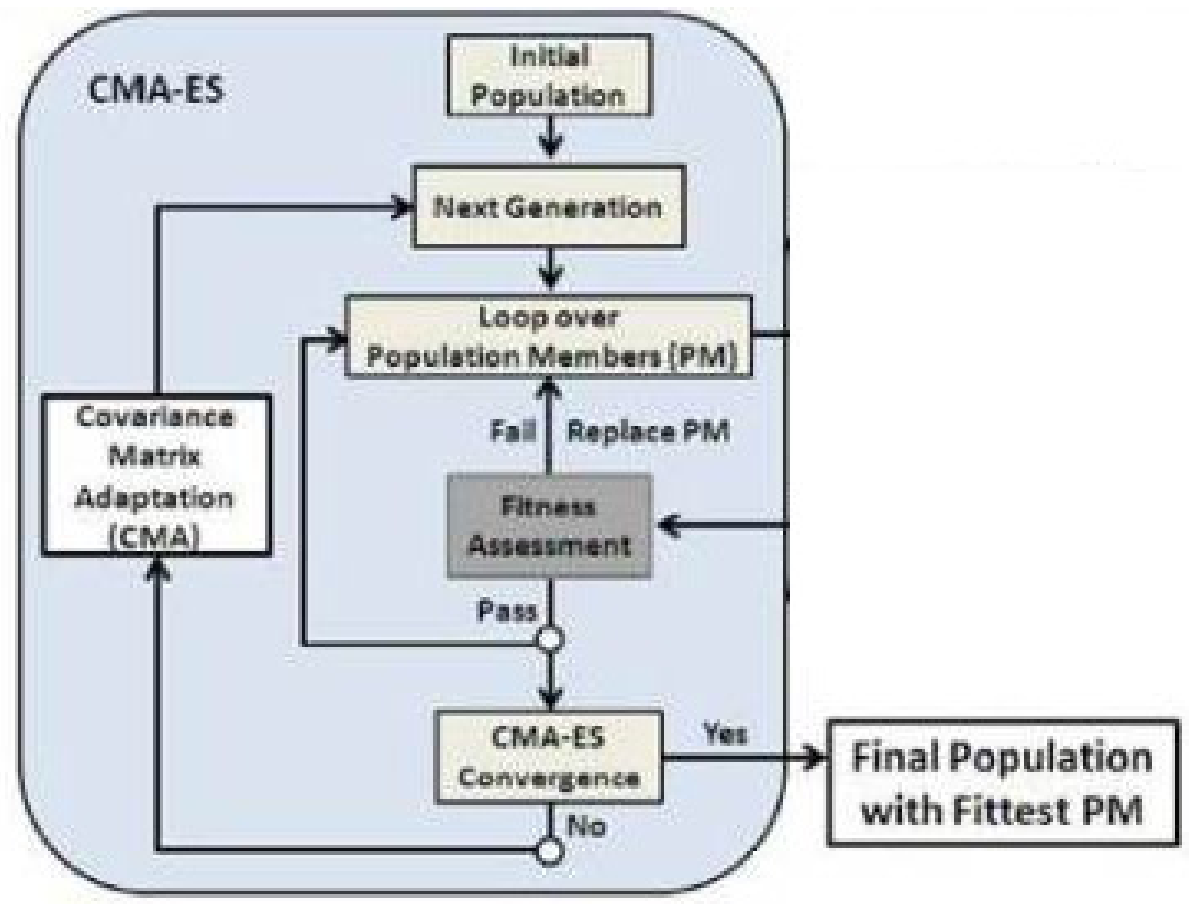

*Figura 11. Esquema general del algoritmo CMA (Schmitz, S. 2006)*

El algoritmo CMA-ES (siglas de estrategia de evolución des-aleatorizada con adaptación de matriz de covarianza) [90, 91, 92] es un método de búsqueda que basa su heurística en la sucesiva adaptación de la matriz de covarianza completa de una distribución normal (distribución gaussiana) de mutaciónes (figura 12).

El esquema general del algoritmo está recogido en la figura 11. En él puede verse como CMA-ES emplea una función gaussiana para generar una descendencia de *mu* soluciones para luego utilizar *lambda* mejores candidatos para refinar la propia función de distribución en un proceso iterativo. El bucle principal del algoritmo (ecuaciones 20.1, 20.2, 20.3) parte de una distribución gaussiana se generan los pesos, afectados por la raíz cuadrada de covarianzas ($\sqrt{\mathbf{C}}$) que actúa como matriz de transformación sobre los datos.

En CMA-ES, la distribución de probabilidad a ser estimada es una distribución normal multivariante N(m, $\delta^2$ C), cuyos parámetros son la media m y la matriz de covarianza $\delta^2$ C. La media representa la localización actual de búsqueda y se moverá en torno a mejores localizaciones a medida que la búsqueda progrese. A continuación, se utilizan esos pesos para perturbar a la nueve de puntos (1.2) y se evalúan según su aptitud (1.3).

(Ec. 20): $\forall i=1,\ldots,\lambda$: $\mathbf{w}_i \leftarrow \sigma \sqrt{\mathbf{C}}\, \mathbf{N}_i(\mathbf{0},\mathbf{1})$, (Ec. 20.1)

$\mathbf{y}_i \leftarrow \mathbf{y}+\mathbf{w}_i$, (L2) (Ec. 20.2)

$\mathbf{p}_i \leftarrow F(\mathbf{y}_i)$, (Ec 20.3)

```
Código Mathematica equivalente:

  OffspringPop = Table[   (* Próxima generación *)
     ( offspring[[4]] = Table[ Random[norm], {n}];
       offspring[[3]] = sigma*(SqrtCov.offspring[[4]]);(* Ec 20 .1 *)
       offspring[[2]] = yParent + offspring[[3]];  (* Ec 20 .2 *)
       offspring[[1]] = f[offspring[[2]]];  (* Ec 20 .3 *)
       offspring
    ), {lambda} ];
```

La matriz de covarianza controla las mutaciones y se usa para guiar la búsqueda. CMA-ES descompone la matriz de covarianza total en una matriz de covarianza C y la varianza global δ, llamado control de tamaño de paso (step-size control). Una vez terminado el proceso de reajuste se procede a realizar la selección (21.1) y el cruce (21.2) hasta alcanzar el tamaño apropiado indicado por lambda.

(Ec21): $\forall i=1,\ldots,mu$

$\mathbf{Sel} \leftarrow \mathrm{Sel} \cup \mathrm{Seleccionar}(\mathbf{p}_i)$ (Ec. 21.1)

$\mathbf{y}^* \leftarrow \mathrm{Recombinar}(\mathbf{sel})$ (Ec. 21.2)

```
Código Mathematica equivalente:

ParentPop = Take[Sort[OffspringPop], mu]; (* tomar los padres *)
Desc = Sum[ParentPop[[m]], {m, 1, mu}]/mu;(* el cruce *) (* Ec 21 *)
yParent = Desc[[2]]; (* el nuevo centro masas *)
```

Los diseñadores del algoritmo encontraron útil controlar el tamaño de los pasos de la mutación $\delta^2$ independientemente de la dirección de búsqueda C. La razón es que estimar C requiere muchas muestras y, por tanto, lleva más tiempo ser adaptada, mientras $\delta^2$ puede ser adaptada en marcos de tiempo más cortos. Lo que se persigue con esto es intentar reducir el número de generaciones para llegar a buenos individuos, mientras se evita caer en mínimos locales. Puede observarse como el vector obtenido en Ec. 20.1 conecta el Yparent de dos generaciones. Posteriormente se procede a auto-adaptar los valores de la propia búsqueda. La ecuación 22, ajusta el vector de dirección de la búsqueda. Aquí el termino (1-(1/*tau*)) es un término de memoria (*cumulacion*). Y decrece conforme a la convergencia.

(Ec. 22): S- VECTOR DE **DIRECCION**

$$\mathbf{s} \leftarrow \left(1 - \frac{1}{\tau}\right)\mathbf{s} + \sqrt{\frac{\mu}{\tau}\left(2 - \frac{1}{\tau}\right)}\,\frac{\langle \mathbf{w} \rangle}{\sigma},$$

```
Código Mathematica equivalente:

  s = (1-1/tau)*s+Sqrt[mu/tau*(2-1/tau)]*Desc[[3]]/sigma;(* Ec 22 *)
```

Posteriormente, el vector de dirección calculado es usado para actualizar la matriz C (ecuación 23). Como se ha dicho anteriormente *tauC* es una ponderación de tiempo en función de la generación actual, como este va decreciendo el elipsoide va colapsando en cada generación.

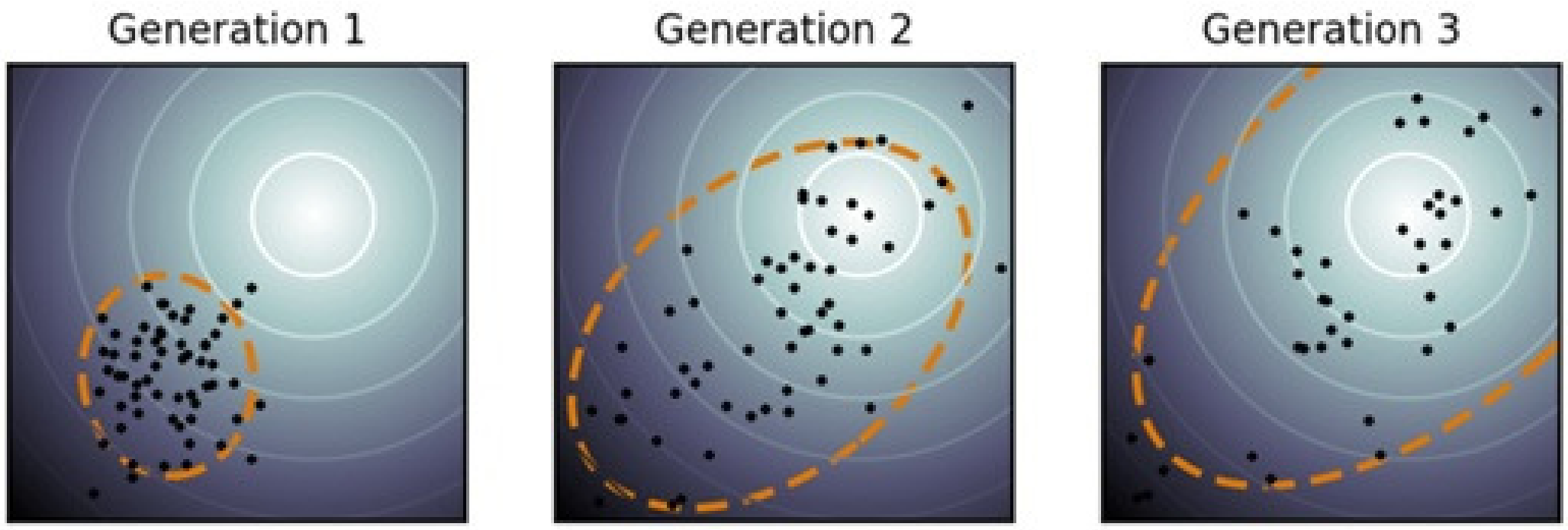

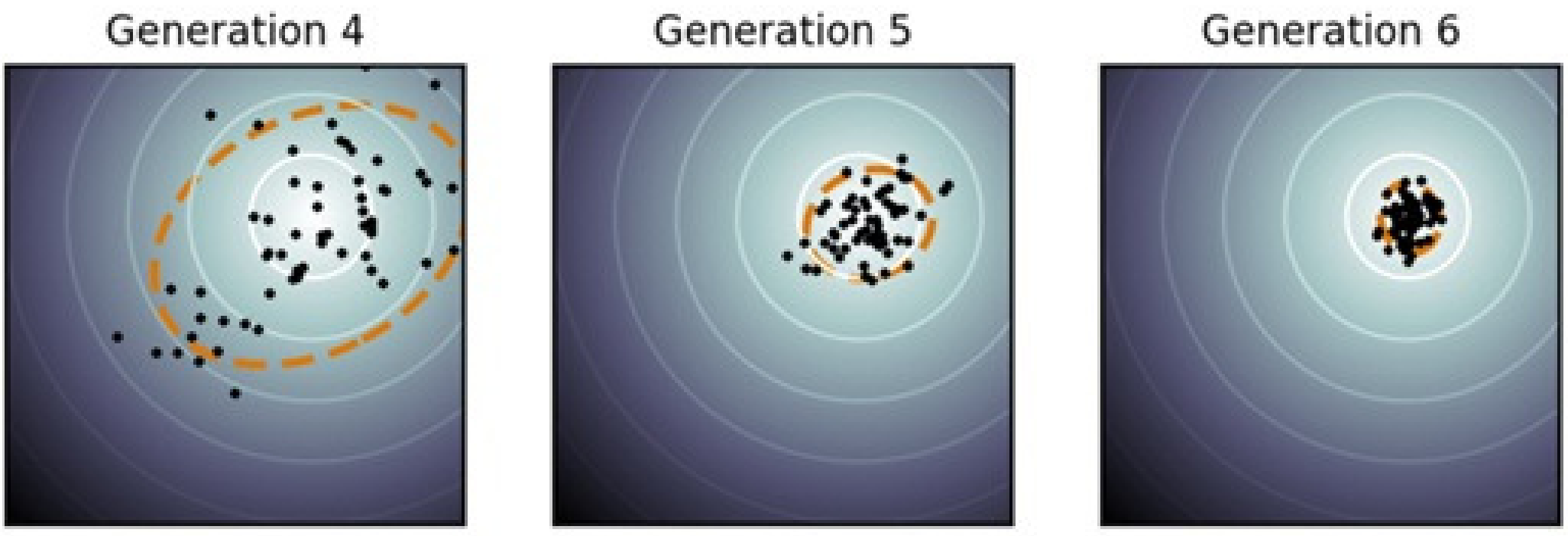

*Figura 12. Ejemplo de búsqueda adaptativa (Fuente: Wikimedia Commons 2013)*

CMA-es controla el tamaño de los pasos comparándolo con el que tendrían si se eligiese un camino aleatorio. Si el camino es más corto significa que los pasos son muy largos y se producen ciclos, en cambio si el camino es más largo significa que se dan varios pasos en la misma dirección, por lo que podrían reducirse si los pasos fuesen más largos. La media, el *step-size* y la matriz de covarianza se actualizan en cada generación t.

(Ec. 23): EL VECTOR DE DIRECCION ES USADO PARA ACTUALIZAR C

$$\mathbf{C} \leftarrow \left(1 - \frac{1}{\tau_c}\right)\mathbf{C} + \frac{1}{\tau_c}\mathbf{s}\mathbf{s}^T,$$

```
Código Mathematica equivalente:

 Cov = (1-1/tauC)*Cov + Outer[Times, (s/tauC), s];      (* Ec 23 *)
```

Por último, las ecuaciones 24.1 y 24.2 se encargan de actualizar los parámetros de la distribución de forma que se produce un mejor ajuste como consecuencia del mayor conocimiento del área explorada.

(Ec. 24):

$$\mathbf{s}_\sigma \leftarrow \left(1 - \frac{1}{\tau_\sigma}\right)\mathbf{s}_\sigma + \sqrt{\frac{\mu}{\tau_\sigma}\left(2 - \frac{1}{\tau_\sigma}\right)}\langle \mathbf{N}(\mathbf{0}, \mathbf{1})\rangle,$$

(Ec. 24.1)

$$\sigma \leftarrow \sigma \exp\left[\frac{\|\mathbf{s}_\sigma\|^2 - n}{2n\sqrt{n}}\right]$$

(Ec. 24.2)

```
Código Mathematica equivalente:

 sSigma = (1-1/tauSigma)*sSigma +
            Sqrt[mu/tauSigma*(2-1/tauSigma)] * Desc[[4]];(* Ec 5 *)

sigma = sigma*Exp[(sSigma.sSigma - n)/(2*n*Sqrt[n])];    (* Ec 6 *)
```

# Capítulo 7.

# Experimentación y análisis de los resultados obtenidos.

En las secciones anteriores, hemos profundizado en la complejidad del problema de estimación del parámetro de degradación del hormigón armado sometido a cortante y repasado su expresión analítica y las técnicas que se han utilizado tradicionalmente para su abordaje. Así mismo, hemos realizado una introducción a las técnicas de computación evolutiva y descrito en detalle el funcionamiento del subgrupo de estas técnicas conocido como estrategias de evolución (ES). En el presente capítulo, utilizaremos una de esas técnicas (descritas en el apartado anterior) para tratar de conseguir el mejor ajuste para el parámetro de una función polinómica de grado 3 que cumpla los requisitos impuestos por el problema. El problema a resolver puede entonces enunciarse como un problema de optimización donde deberemos encontrar los cuatro coeficientes para el polinomio interpolador del parámetro que consiguen un mejor ajuste. Un vector *R* candidato a solución vendrá por tanto formado por cuatro números reales **{*R1, R2, R3, R4*}** que representan los cuatro coeficientes del polinomio buscado.

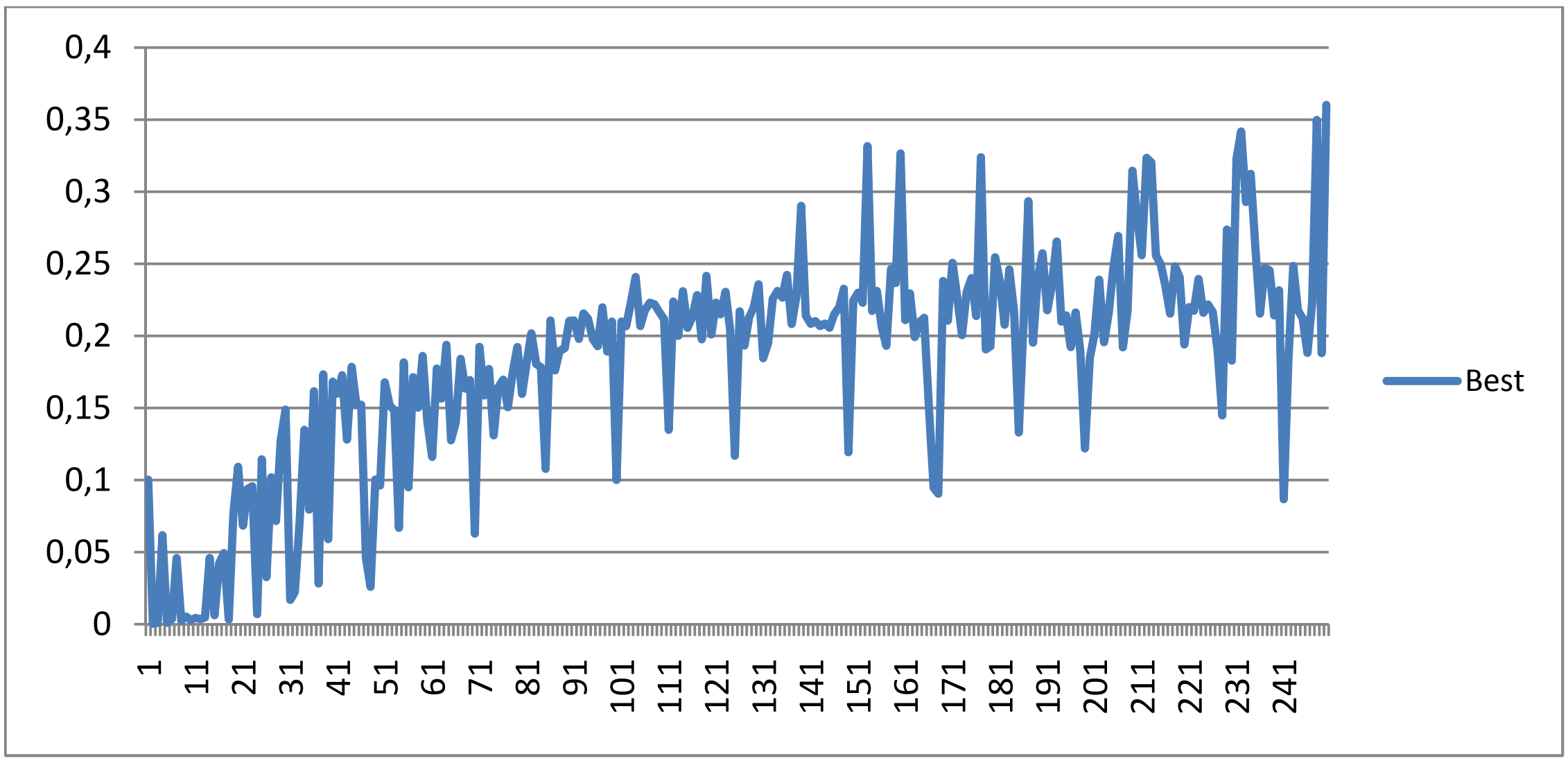

*Figura 23. Evolución del fitness para la implementación básica de cma-es. Pueden observarse como aparecen reiteradas generaciones perdidas.*

El algoritmo evolutivo utilizado será una versión adaptada y mejorada de la utilizada por los autores en los paper de Beyer et al. de 2001 [29b] y Hans-Georg Beyer de 2007 respectivamente.

Al tratarse de un problema especialmente duro computacionalmente y ser muy costoso el cálculo de  la función de aptitud, el algoritmo CMA-es aparece como una buena opción por ser un algoritmo que no necesita de un gran número de evaluaciones y que funciona relativamente bien con poblaciones pequeñas.

En la figura 23 observamos los valores de fitness obtenidos para la mejor de tres ejecuciones. En ella se puede ver como el valor de aptitud crece rápidamente en las primeras generaciones pero se estanca después de la generación 100 oscilando alrededor del 22%. En la figura 24 están graficadas las mejores soluciones encontradas por el algoritmo (versión 1) durante la misma ejecución para 100, 150 y 200 generaciones.

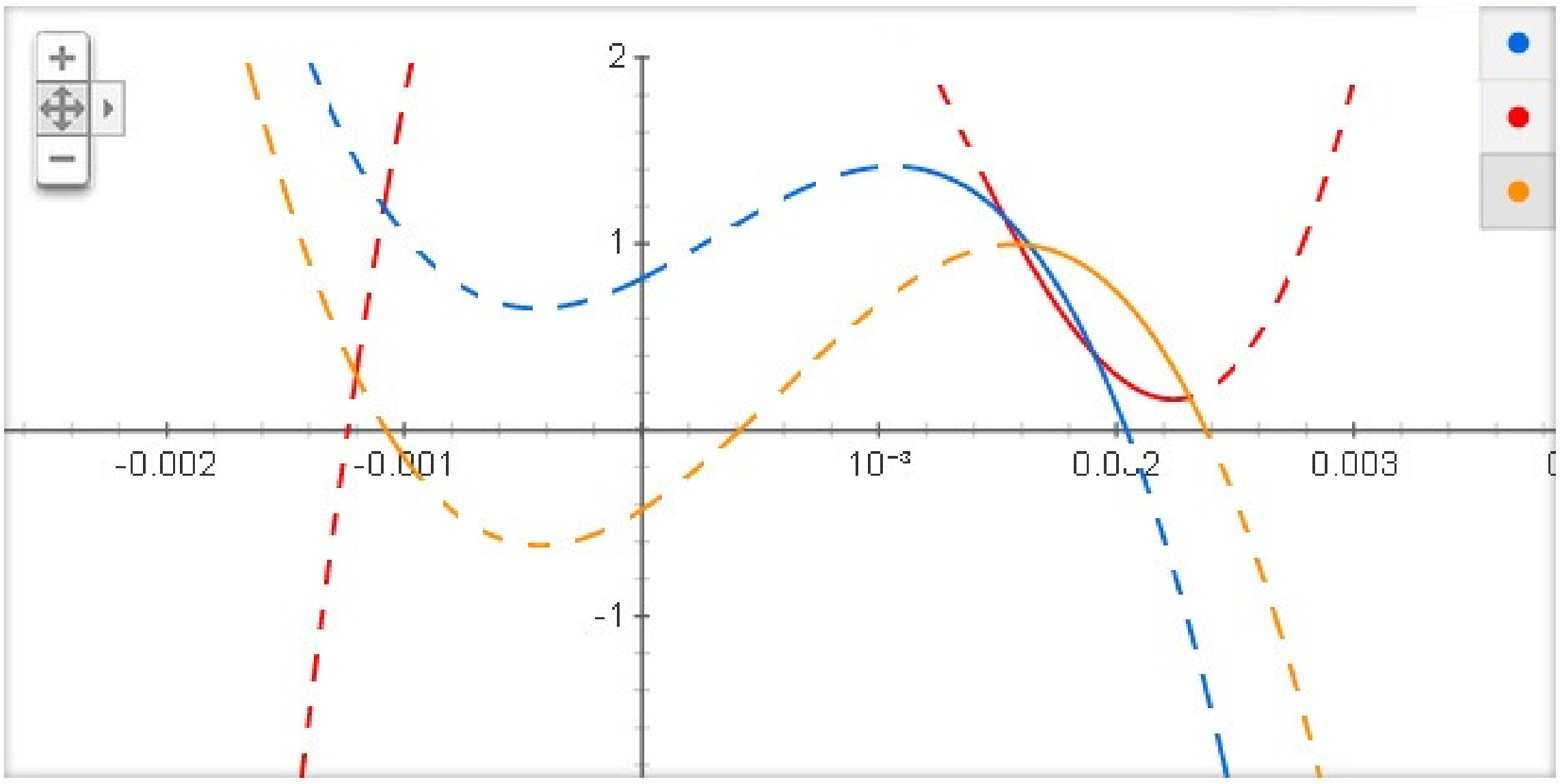

*Figura 24. Detalle de resultados obtenidos por CMA-es (versión 1) para: naranja (generación 100), azul (generación 150) y roja (generación 200).*

Si restringimos la gráfica eliminando valores imposibles (segmentos del primer cuadrante con pendiente ascendente) obtenemos el detalle de la figura 25.

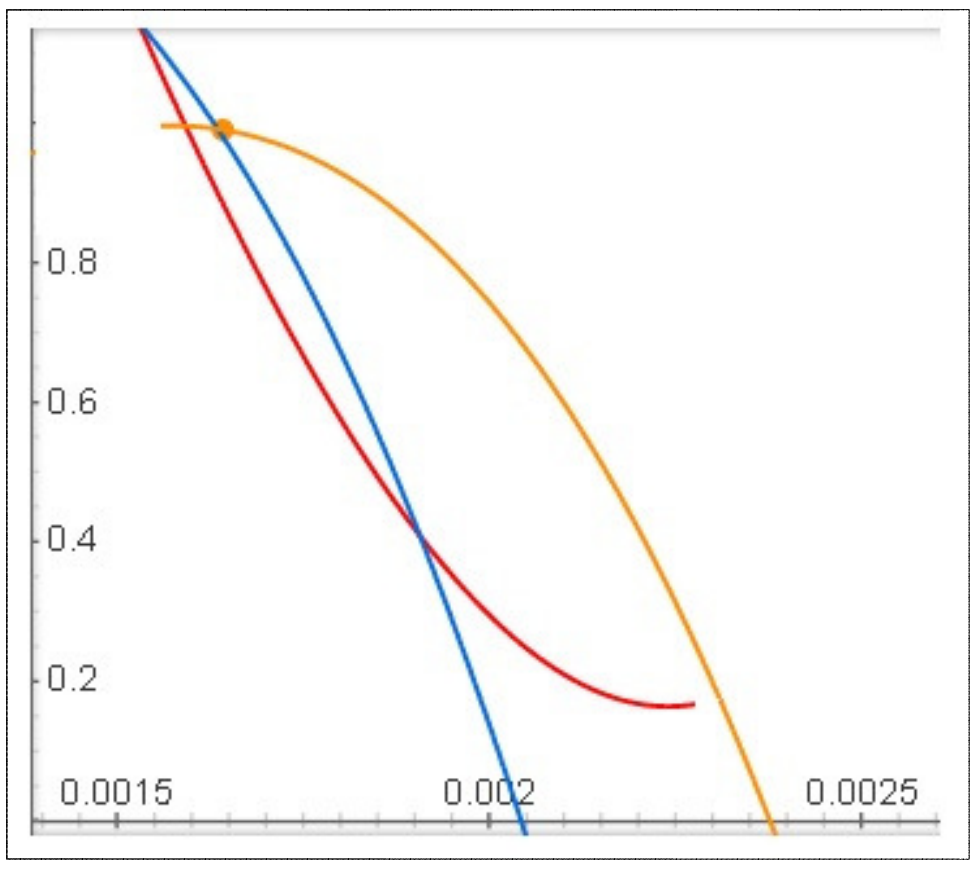

*iones*

Sin embargo los resultados resultan pobres y la mejora conseguida es muy modesta. En la figura 23, se aprecian varias caídas bruscas que seguro se corresponden con soluciones inválidas que aparecen incluso en estadios avanzados de la búsqueda. Asociamos dichos saltos a valores donde la función fitness no está definida. Una posible causa de esto podría ser el que para la función de aptitud con la que trabajamos, dos polinomios que están muy próximos pueden tener puntuaciones de aptitud radicalmente distintos ya un pequeño cambio en los coeficientes puede hacer que el sistema de ecuaciones no tenga solución (ver capítulo 4). Para esquivar estos problemas, se decide recuperar el valor de penalización que proponían los autores en el estudio original [101]. Este valor (debe ser un número grande en comparación con el rango donde se mueve la función de aptitud pero no excesivamente grande para no causar descarte en cadena), hará que se descarten las soluciones invalidas y que se minimice la posibilidad de que estas sean seleccionadas. También se decide fijar el número de padres considerados a dos ya que de esta forma, un descendiente será combinación lineal de los coeficientes de únicamente dos padres lo que debería minimizar el riesgo de dar saltos bruscos.

Hechos estos cambios volvemos a ejecutar el algoritmo (versión 2) obtenido en este caso una gráfica de fitness donde se observa un valor más estable y una tendencia más clara al alza (figura 26), aunque los resultados siguen sin ser los esperados.

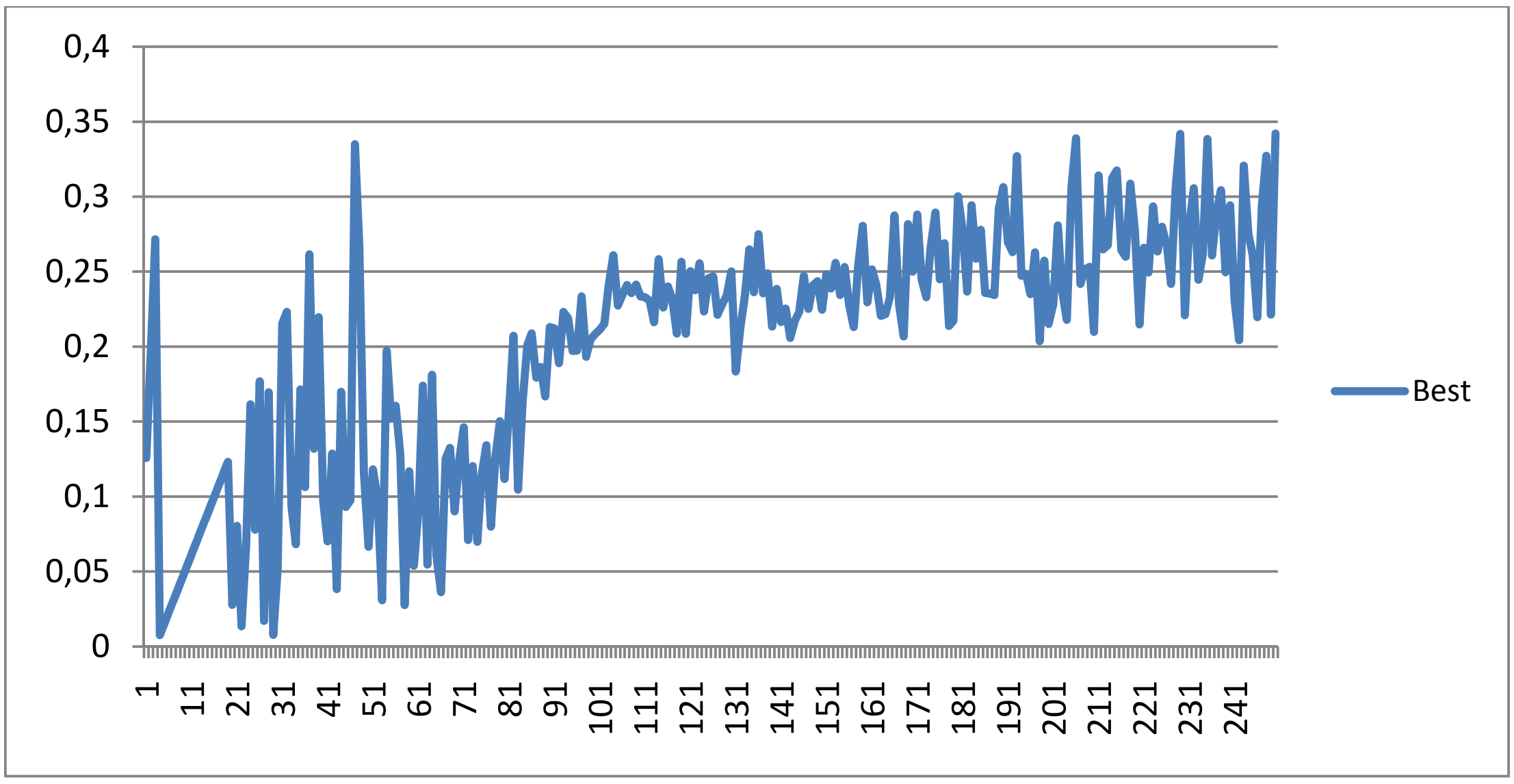

*Figura 26. Evolución del fitness en la ejecución de CMA-es con las modificaciones referidas (versión 2).*

A la vista de la nueva grafica observamos que aparecen esporádicamente, incluso en generaciones muy tempranas, soluciones medianamente buenas con valores de aptitud por encima del 25% (recordemos que el polinomio obtenido por LGA tenía un ajuste del 16,6%), características que sin embargo, parecen no heredarse a sus descendientes.

Esto nos mueve a replantear de nuevo el algoritmo y se decide probar una nueva modificación del algoritmo base buscando producir un comportamiento elitista mediante la incorporación de una memoria global de soluciones, que hará las veces de archivo genético regenerando con periodicidad semi-aleatoria la variabilidad genética de la población, evitando de esta forma, la convergencia prematura y orientando la búsqueda hacia zonas que resultaron ser prometedoras (versión 3).

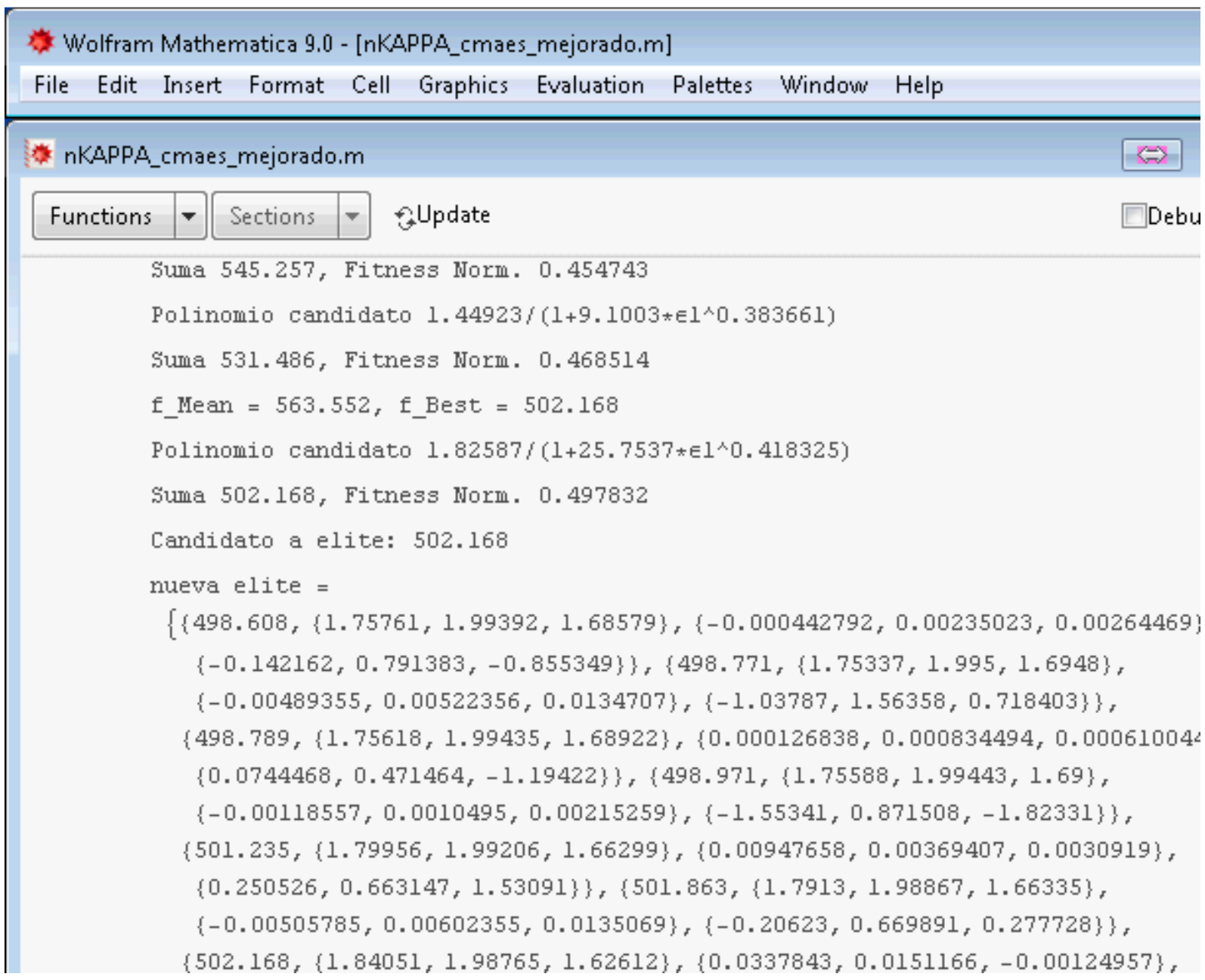

*Figure 27. Fragmento de la consola de mensajes durante la ejecución del código.*

Así mismo, y para minimizar el número de evaluaciones, se decide hacer variar el número de descendientes (lambda) de manera que el tamaño de la descendencia se ajuste según sea la diferencia entre el máximo fitness y el fitness promedio. El mecanismo descrito, puede observarse en el fragmento de la salida de mensajes que se aprecia en la captura (figura 27). En ella vemos como la población se retroalimenta en ciertas iteraciones de individuos que habían quedado apartados durante un tiempo del flujo evolutivo.

Para confirmar el correcto funcionamiento de la modificación, se realizan una nueva serie de ejecuciones y graficamos de nuevo tomando la evolución de la mejor aptitud para cada generación. En la nueva grafica (figura 28), observamos un crecimiento más suave y progresivo así como una recuperación más rápida cuando el algoritmo avanza en una dirección errónea.

Con todo, los resultados son mejores en las nuevas pruebas pero no satisfacen las expectativas del experimento por lo que decidimos sustituir la función buscada (hasta ahora veníamos trabajando con funciones polinómicas de grado 3) por una función de la forma **κ(ε1)=a/(1+b*(ε1)^c),** ya que creemos que la forma de esta función podría ajustarse mejor a nuestros datos experimentales (versión 4).

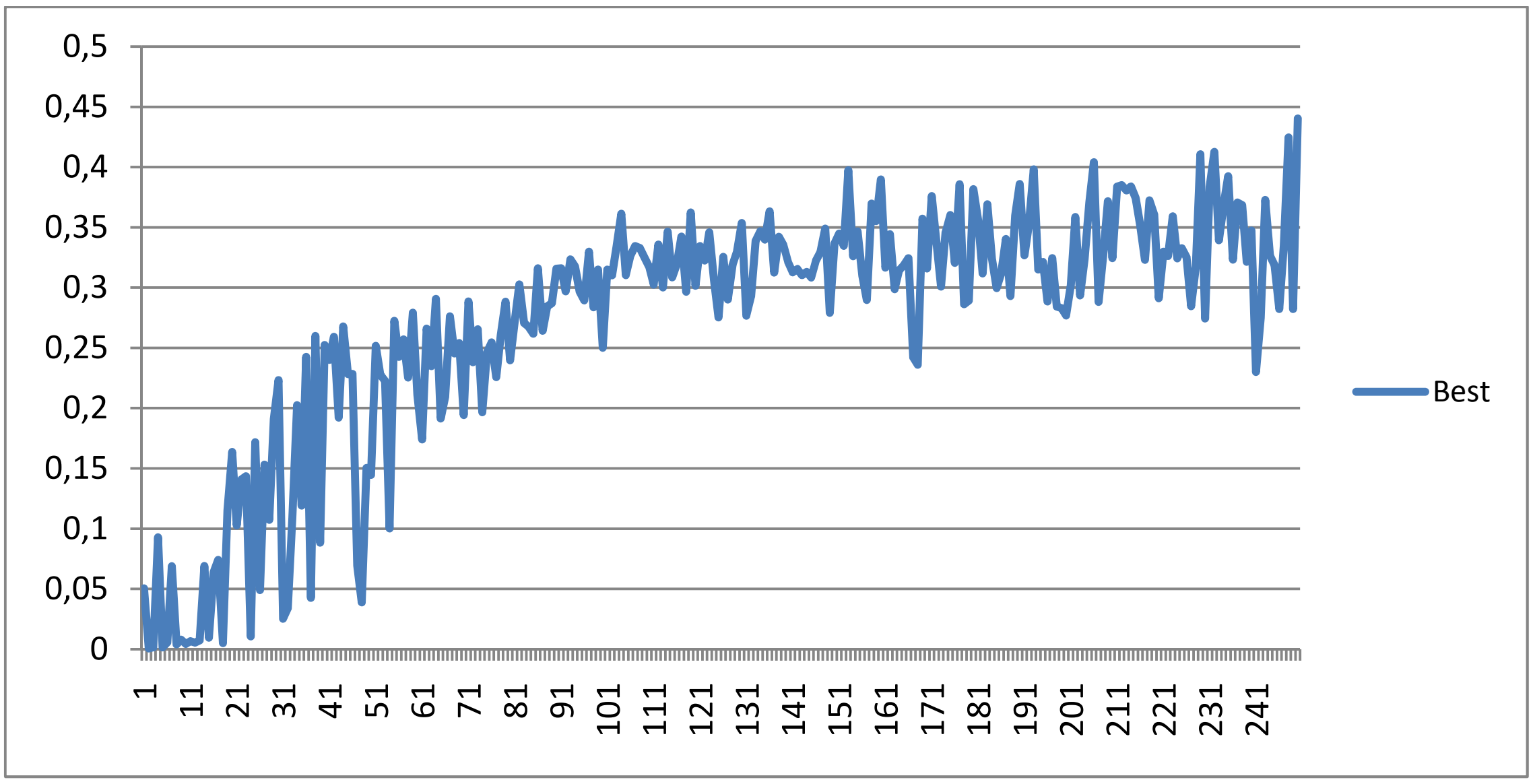

*Figure 28. Evolución del fitness en la ejecución de algoritmo CMA-es con modificaciones referidas (versión 3)*

Al realizar nuevamente la batería de pruebas nos sorprende gratamente observar como mejoran sensiblemente la correspondencia obteniendo valores de aptitud en el entorno de 50-53%.

Los datos obtenidos son bastante mejores que en la anterior ocasión con una convergencia clara que por encima de la generación 150 (esta tendencia puede observarse en la figura 29) e incluye ya hasta el tercer decimal. Parece que la nueva función favorece la búsqueda disminuyendo la inter-dependencia al estar los parámetros del problema menos acoplados. Por otra parte, el mejor ajuste podría favorecer el que el algoritmo pase más tiempo en el espacio de soluciones aceptables.

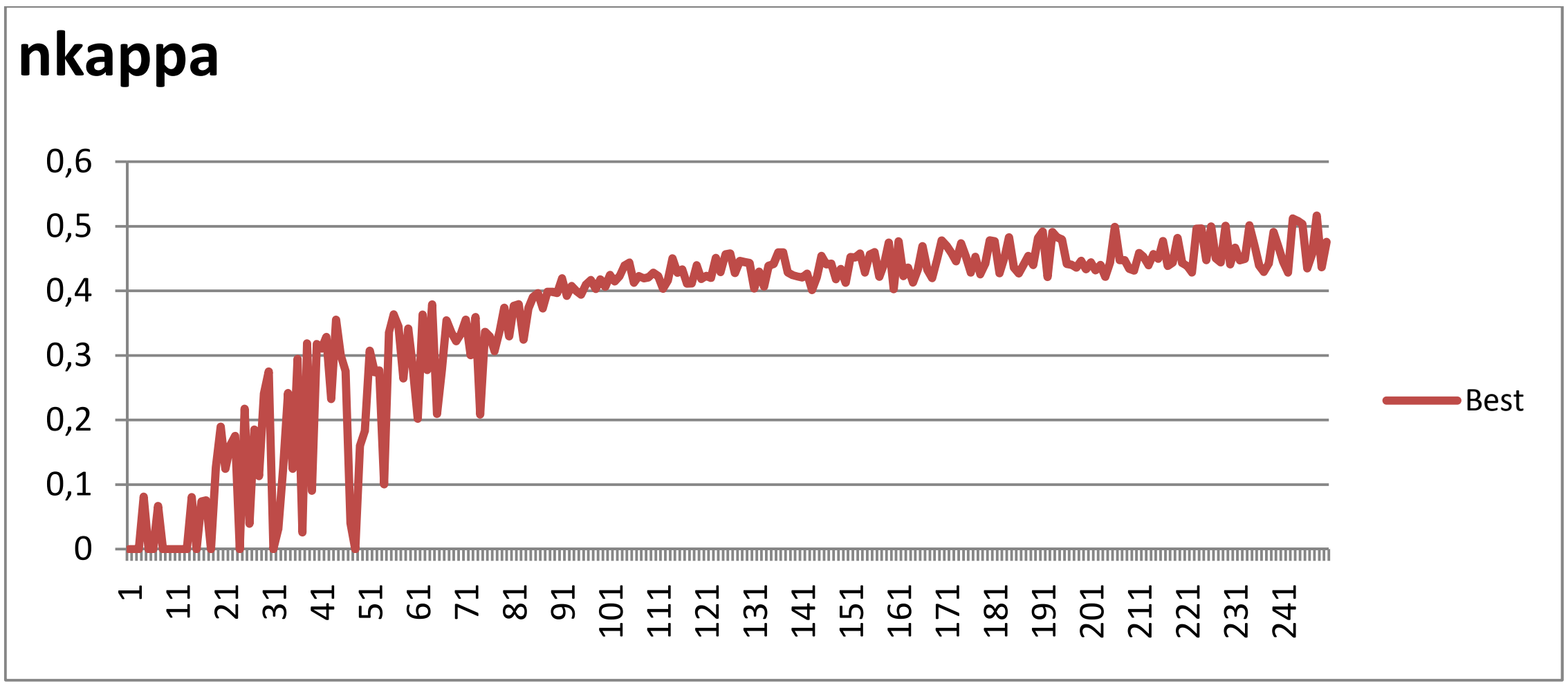

*Figura 29. Evolución del fitness en la ejecución de algoritmo CMA-es con modificaciones referidas (versión 4)*

Tomemos por ejemplo el caso de la función polinómica; En él, el parámetro *R1* se correspondía con el coeficiente del término de orden 3, así, que una variación en este coeficiente alteraba considerablemente el resultado, mientras, que un cambio similar en el coeficiente de orden 1 alteraría en pequeña medida el resultado.

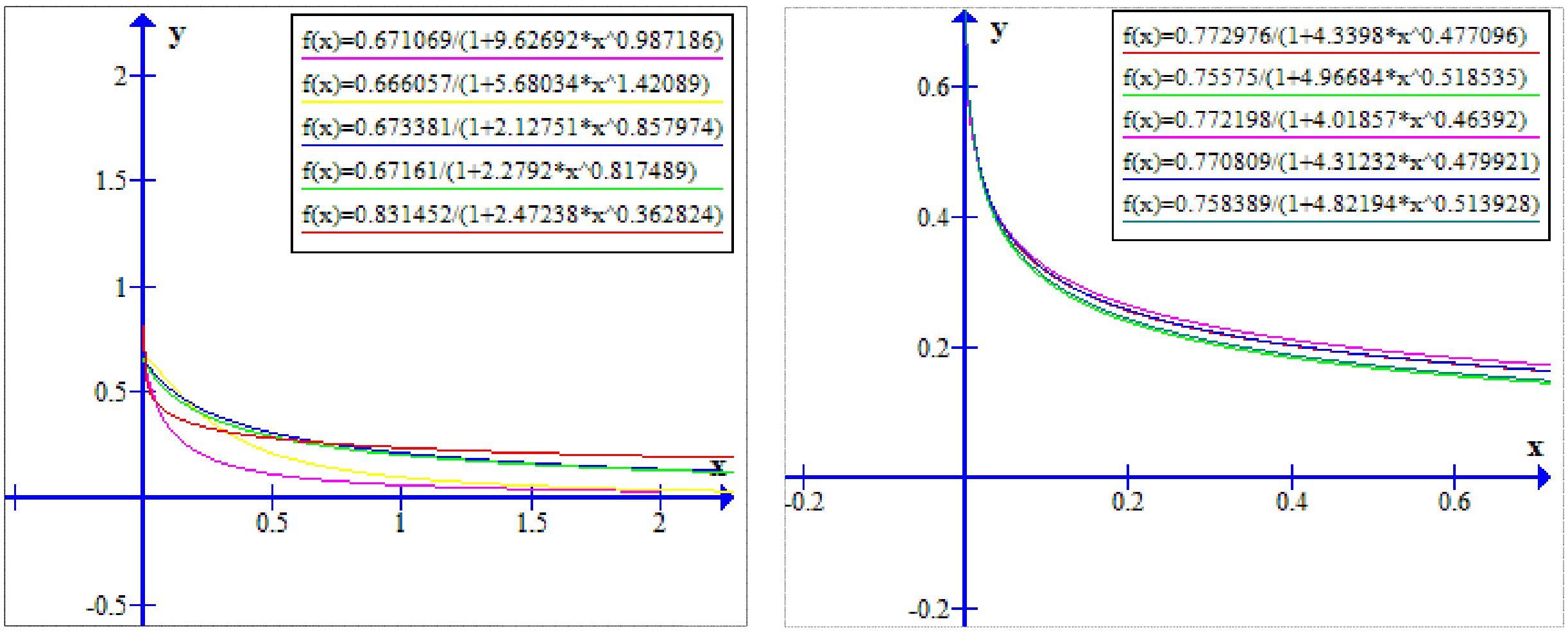

*Figure 30. IZQ. Detalle de resultados obtenidos por CMA-es (versión 4). Mejores para generación 25 (morado), 50 (rojo), 75 (amarillo), 100 (verde), y 150 (azul). A partir de ahí las gráficas se solapan y es difícil graficar. DER. Los cinco mejores resultados obtenidos. El mejor, representado en color azul verdoso obtuvo un 51% de ajuste.*

La figura 30 muestra la evolución del mejor candidato con un lapso de 25 generaciones. En ella puede observarse gráficamente cómo evolucionan los parámetros de entrada para ajustar la función (izquierda) y como se lleva a cabo el ajuste fino local (derecha).

La convergencia ahora puede apreciarse claramente en la fase de ajuste local cuando el tamaño del paso tiende a hacerse muy pequeño, y es por eso, de que a pesar de que no tenemos un valor optimo global creemos bastante probable que lo solución ofrecida sea una función muy próxima a los datos y aproxime por tanto de forma muy aceptable a la verdadera función y demuestra la aplicabilidad de este tipo de técnicas de optimización para el problema considerado por lo que damos nuestros resultados por buenos a la espera de una posterior confirmación experimental de los mismos en el laboratorio.

# Capítulo 8.

# Conclusiones y posibles trabajos futuros.

Los resultados obtenidos demuestran que las estrategias de evolución pueden mejorar notablemente los resultados obtenidos mediante métodos de regresión y tienen una aplicación clara en problemas donde el paisaje de búsqueda no se conoce o es imposible de calcular.

En el problema en cuestión que ha sido tratado, son varios los inconvenientes con los que nos hemos encontrado y que hemos tenido que lidiar. El primero de ellos es la costosa evaluación de candidatos y la imposibilidad de aproximar su aptitud mediante alguna directriz heurística.

La representación de los parámetros del problema (que en nuestro caso se correspondían con los coeficientes del polinomio interpolador) promovía el solapamiento de las variables, de forma que, por ejemplo, un pequeño cambio en el coeficiente de orden 3 podía alterar enormemente el resultado mientras que un cambio similar en el término independiente solo ocasionaba un cambio pequeño. Para mayor dificultad, encontramos que la función de aptitud no estaba definida en todos los valores de forma que era enormemente difícil para el algoritmo aprender acerca del espacio de soluciones y un camino evolutivo prometedor podía acabar degenerando en soluciones inválidas.

Como línea de investigación futura debería proponerse una reformulación de la representación de los candidatos que permitiese una mayor libertad al método genético, a la vez que aprovechase mejor sus capacidades y limitase la degeneración de la población. Por ejemplo, podría considerarse permitir al algoritmo elegir el tipo de función interpoladora o incluso construir la suya propia a partir de bloques constructivos básicos.

También, y debido a las capacidades implícitas de paralelización de los AE en futuras ampliaciones podrían considerarse estrategias de evolución en implementación paralela para explotar al máximo las posibilidades de ejecución concurrente en multi-cpu, computación GPU, o ejecución en un Grid dedicado de alto rendimiento.

# Bibliografía

[1] MOSER, Martin; JOKANOVIC, Dusan P.; SHIRATORI, Norio. An algorithm for the multidimensional multiple-choice knapsack problem. IEICE transactions on fundamentals of electronics, communications and computer sciences, 1997, vol. 80, no 3, p. 582-589..

[2] TAMIR, Arie. New pseudopolynomial complexity bounds for the bounded and other integer Knapsack related problems. Operations Research Letters, 2009, vol. 37, no 5, p. 303-306.

[3 TAVAKKOLI-MOGHADDAM, Reza; RAHIMI-VAHED, Alireza; MIRZAEI, Ali Hossein. A hybrid multi-objective immune algorithm for a flow shop scheduling problem with bi-objectives: weighted mean completion time and weighted mean tardiness. Information Sciences, 2007, vol. 177, no 22, p. 5072-5090.

[4] VEGA GARCIA, C., et al. Applying neural network technology to human-caused wildfire occurrence prediction. AI applications, 1996, vol. 10..

[5] SUD, Avneesh, et al. Real-time path planning for virtual agents in dynamic environments. En ACM SIGGRAPH 2008 classes. ACM, 2008. p. 55.

[6] SUD, Avneesh, et al. Real-time navigation of independent agents using adaptive roadmaps. En Proceedings of the 2007 ACM symposium on Virtual reality software and technology. ACM, 2007. p. 99-106..

[7] STENTZ, Anthony. Optimal and efficient path planning for partially-known environments. En Robotics and Automation, 1994. Proceedings., 1994 IEEE International Conference on. IEEE, 1994. p. 3310-3317.

[8] LI, Yi; GUPTA, Kamal. Motion planning of multiple agents in virtual environments on parallel architectures. En Robotics and Automation, 2007 IEEE International Conference on. IEEE, 2007. p. 1009-1014.

[9] SANCHES, Carlos Alberto Alonso; SOMA, Nei Yoshihiro; YANASSE, Horacio Hideki. An optimal and scalable parallelization of the< i> two-list</i> algorithm for the subset-sum problem. European Journal of Operational Research, 2007, vol. 176, no 2, p. 870-879.

[10] PLAZA, Antonio; VALENCIA, David; PLAZA, Javier. An experimental comparison of parallel algorithms for hyperspectral analysis using heterogeneous and homogeneous networks of workstations. Parallel Computing, 2008, vol. 34, no 2, p. 92-114..

[11] FINK, Andreas; VOB, Stefan. Generic metaheuristics application to industrial engineering problems. Computers & Industrial Engineering, 1999, vol. 37, no 1, p. 281-284.

[12] GLOVER, Fred W.; KOCHENBERGER, Gary A. Handbook of metaheuristics (International series in operations research & management science). 2003.

[21] FOGEL, David B. Artificial Intelligence Through Simulated Evolution. Wiley-IEEE Press, 1967. New York: Wiley Publishing

[22] RECHENBERG, Ingo. Cybernetic solution path of an experimental problem. 1965.

[23] RECHENBERG, Ingo, I.: Evolutionsstrategie : Optimierung technischer Systeme nach Prinzipien der biologischen Evolution. 15. Stuttgart-Bad Cannstatt : Frommann-Holzboog, 1973.

[24] SCHWEFEL, H.P. Numerische Optimierunguon Computer-Modellenmittels der Ezdutionsstrategie, 1977, Volume 26 of Interdisciplinary systems research. Basel Birkhauser.

[25] HOLLAND, John H. Outline for a logical theory of adaptive systems. Journal of the ACM (JACM), 1962, vol. 9, no 3, p. 297-314..

[26] HOLLAND, John H. Adaptation in natural and artificial systems: An introductory analysis with applications to biology, control, and artificial intelligence. U Michigan Press, 1975.

[26b] JOHN, Holland. Holland, Adaptation in Natural and Artificial Systems: An Introductory Analysis with Applications to Biology, Control and Artificial Intelligence. 1992.

[27] TURCK-CHIÈZE, Sylvaine. The solar interior. Physics reports, 1993, vol. 230, no 2, p. 57-235..

[28] EIBEN, Agosten E.; SMITH, James E. Introduction to evolutionary computing. Berlin: Springer, 2010.

[29] MICHALEWICZ, Zbigniew. Genetic algorithms + data structures = evolution programs. springer, 1996..

[29b] BEYER, Hans-Georg. The theory of evolution strategies. Springer, 2001.

[30] ZAFRA, Amelia; GIBAJA, Eva L.; VENTURA, Sebastián. Multiple instance learning with multiple objective genetic programming for web mining.Applied Soft Computing, 2011, vol. 11, no 1, p. 93-102

[31] AKBARZADEH-T, M. R., et al. Soft computing paradigms for hybrid fuzzy controllers: experiments and applications. En Fuzzy Systems Proceedings, 1998. IEEE World Congress on Computational Intelligence., The 1998 IEEE International Conference on. IEEE, 1998. p. 1200-1205..

[32] QIN, Hao; YANG, Simon X. Adaptive neuro-fuzzy inference systems based approach to nonlinear noise cancellation for images. fuzzy sets and systems, 2007, vol. 158, no 10, p. 1036-1063..

[33] CHARBONNEAU, Paul. Genetic algorithms in astronomy and astrophysics.The Astrophysical Journal Supplement Series, 1995, vol. 101, p. 309.

[34] TANG, Kit-Sang, et al. Genetic algorithms and their applications. Signal Processing Magazine, IEEE, 1996, vol. 13, no 6, p. 22-37.

[35] KEANE, A. J. The design of a satellite beam with enhanced vibration performance using genetic algorithm techniques. The Journal of the Acoustical Society of America, 1996, vol. 99, no 4, p. 2599-2603..

[36] ALTSHULER, Edward E.; LINDEN, Derek S. Wire-antenna designs using genetic algorithms. Antennas and Propagation Magazine, IEEE, 1997, vol. 39, no 2, p. 33-43..

[37] HUGHES, Evan J.; LEYLAND, Maurice. Using multiple genetic algorithms to generate radar point-scatterer models. Evolutionary Computation, IEEE Transactions on, 2000, vol. 4, no 2, p. 147-163.

[38] VASILE, Massimiliano. Hybrid behavioral-based multiobjective space trajectory optimization. En Multi-Objective Memetic Algorithms. Springer Berlin Heidelberg, 2009. p. 231-253.

[39] METCALFE, Travis S.; CHARBONNEAU, Paul. Stellar structure modeling using a parallel genetic algorithm for objective global optimization. Journal of Computational Physics, 2003, vol. 185, no 1, p. 176-193.

[40] GIRO, R., M. Cyrillo y D.S. Galvão. Designing conducting polymers using genetic algorithms.Chemical Physics Letters, vol.366, no.1-2, p.170-175, 2002.

[41] KROO, I. Aeronautical Applications of Evolutionary Design. VKI lecture series on Optimization Methods & Tools for Multicriteria/Multidisciplinary Design, 2004.

[42] HANSEN, Nikolaus; OSTERMEIER, Andreas. Adapting arbitrary normal mutation distributions in evolution strategies: The covariance matrix adaptation. En Evolutionary Computation, 1996., Proceedings of IEEE International Conference on. IEEE, 1996. p. 312-317.

[43] HANSEN, Nikolaus; MÜLLER, Sibylle D.; KOUMOUTSAKOS, Petros. Reducing the time complexity of the derandomized evolution strategy with covariance matrix adaptation (CMA-ES). Evolutionary Computation, 2003, vol. 11, no 1, p. 1-18.

[44] OYAMA, Akira; LIOU, Meng-Sing; OBAYASHI, Shigeru. Transonic axial-flow blade shape optimization using evolutionary algorithm and three-dimensional Navier-Stokes solver. En 9th AIAA/ISSMO Symposium and Exhibit on Multidisciplinary Analysis and Optimization, Atlanta, GA. 2002.

[45] LIAN, Yongsheng; OYAMA, Akira; LIOU, Meng-Sing. Progress in design optimization using evolutionary algorithms for aerodynamic problems.Progress in Aerospace Sciences, 2010, vol. 46, no 5, p. 199-223.

[46] LIAN, Yongsheng; LIOU, Meng-Sing. Multiobjective Optimization Using Coupled Response Surface Model and Evolutionary Algorithm. AIAA journal, 2005, vol. 43, no 6, p. 1316-1325.

[47] LIAN, Yongsheng; LIOU, Meng-Sing; OYAMA, Akira. An enhanced evolutionary algorithm with a surrogate model. En Proceedings of genetic and evolutionary computation conference, Seattle, WA. 2004.

[48] HWATAL, Andreas M.; RAIDL, Günther R. Determining orbital elements of extrasolar planets by evolution strategies. En Computer Aided Systems Theory–EUROCAST 2007. Springer Berlin Heidelberg, 2007. p. 870-877.

[49] PENIAK, Martin; MAROCCO, Davide; CANGELOSI, Angelo. Autonomous robot exploration of unknown terrain: A preliminary model of mars rover robot. En Proceedings of 10th ESA Workshop on Advanced Space Technologies for Robotics and Automation. 2008.

[50] OYAMA, Akira; LIOU, Meng-Sing. Multiobjective optimization of rocket engine pumps using evolutionary algorithm. Journal of Propulsion and Power, 2002, vol. 18, no 3, p. 528-535.

[51] SCHÜTZE, Oliver, et al. Designing optimal low-thrust gravity-assist trajectories using space pruning and a multi-objective approach.Engineering Optimization, 2009, vol. 41, no 2, p. 155-181.

[52] KANG, Shin-Jin; KIM, YongO; KIM, Chang-Hun. Live path: adaptive agent navigation in the interactivevirtual world. The Visual Computer, 2010, vol. 26, no 6-8, p. 467-476.

[53] GOSSELIN, Louis; TYE-GINGRAS, Maxime; MATHIEU-POTVIN, François. Review of utilization of genetic algorithms in heat transfer problems.International Journal of Heat and Mass Transfer, 2009, vol. 52, no 9, p. 2169-2188.

[54] COELLO, CA Coello; CHRISTIANSEN, Alan D.; HERNÁNDEZ, F. Santos. A simple genetic algorithm for the design of reinforced concrete beams.Engineering with Computers, 1997, vol. 13, no 4, p. 185-196.

[54b] COELLO, Carlos Coello; HERNÁNDEZ, Filiberto Santos; FARRERA, Francisco Alonso. Optimal design of reinforced concrete beams using genetic algorithms. Expert systems with Applications, 1997, vol. 12, no 1, p. 101-108.

[55] RAFIQ, Mohammad Y.; SOUTHCOMBE, Colin. Genetic algorithms in optimal design and detailing of reinforced concrete biaxial columns supported by a declarative approach for capacity checking. Computers&structures, 1998, vol. 69, no 4, p. 443-457.

[56] MARROQUÍN, José Luis; BOTELLO RIONDA, Salvador; OÑATE, Eugenio. Un modelo de optimización estocástica aplicado a la optimización de estructuras de barras prismáticas. Revista internacional de métodos numéricos para cálculo y diseño en ingeniería, 1999, vol. 15, no 4, p. 425-434.

[57] KOUMOUSIS, Vlasis K.; ARSENIS, S. J. Genetic algorithms in optimal detailed design of reinforced concrete members. Computer□Aided Civil and Infrastructure Engineering, 1998, vol. 13, no 1, p. 43-52.

[57b] GOVINDARAJ, V.; RAMASAMY, J. V. Optimum detailed design of reinforced concrete continuous beams using genetic algorithms. Computers & structures, 2005, vol. 84, no 1, p. 34-48.

[58] CHAU, K. W.; ALBERMANI, F. Knowledge-based system on optimum design of liquid retaining structures with genetic algorithms. Journal of structural engineering, 2003, vol. 129, no 10, p. 1312-1321.

[59] LEPŠ, Matěj; ŠEJNOHA, Michal. New approach to optimization of reinforced concrete beams. Computers & structures, 2003, vol. 81, no 18, p. 1957-1966.

[60] LEE, C.; AHN, J. Flexural design of reinforced concrete frames by genetic algorithm. Journal of structural engineering, 2003, vol. 129, no 6, p. 762-774.

[61] FAIRBAIRN, Eduardo MR, et al. Optimization of mass concrete construction using genetic algorithms. Computers & structures, 2004, vol. 82, no 2, p. 281-299.

[62] LIM, Chul-Hyun; YOON, Young-Soo; KIM, Joong-Hoon. Genetic algorithm in mix proportioning of high-performance concrete. Cement and Concrete Research, 2004, vol. 34, no 3, p. 409-420.

[63] AMIRJANOV, Adil; SOBOLEV, Konstantin. Optimal proportioning of concrete aggregates using a self-adaptive genetic algorithm. Computers and Concrete, 2005, vol. 2, no 5, p. 411-421.

[64] QIN, A. Kai; SUGANTHAN, Ponnuthurai N. Self-adaptive differential evolution algorithm for numerical optimization. En Evolutionary Computation, 2005. The 2005 IEEE Congress on. IEEE, 2005. p. 1785-1791.

[65] GERO, Mª Belén Prendes, et al. APLICACIÓN DE UN ALGORITMO GENÉTICO ELITISTA EN LA OPTIMIZACIÓN DE EDIFICIOS METÁLICOS..

[66] DEKA, Dhyanjyoti. Crystal plasticity modeling of deformation and creep in polycrystalline, Metallurgical and Materials Transactions A, 2006, vol. 37, no 5, p. 1371-1388.

[67] NEHDI M, El Chabib H, Said A. Genetic algorithm model for shear capacity of RC beams reinforced with externally bonded FRP. Materials and Structures 44:1249–1258, 2011

[68] NEHDI M, El Chabib H, Said A. Proposed shear design equations for FRP-reinforced concrete beams based on genetic algorithms approach. ASCE J Mater CivEng 19:1033–1042, 2007.

[69] BONISSONE, Piero P. Soft computing: the convergence of emerging reasoning technologies. Soft computing, 1997, vol. 1, no 1, p. 6-18.

[70] KECMAN, Vojislav. Learning and soft computing: support vector machines, neural networks, and fuzzy logic models. MIT press, 2001.

[71] MITRA, Sushmita; PAL, Sankar K.; MITRA, Pabitra. Data mining in soft computing framework: A survey. IEEE transactions on neural networks, 2002, vol. 13, no 1, p. 3-14.

[72] SANCHEZ, Elie; SHIBATA, Takanori; ZADEH, Lotfi Asker (ed.). Genetic algorithms and fuzzy logic systems: Soft computing perspectives. World Scientific, 1997.

[73] GONZÁLEZ, Javier Serrano, et al. Optimization of wind farm turbines layout using an evolutive algorithm. Renewable Energy, 2010, vol. 35, no 8, p. 1671-1681.

[74] MAULIK, Ujjwal; BANDYOPADHYAY, Sanghamitra. Genetic algorithm-based clustering technique. Pattern recognition, 2000, vol. 33, no 9, p. 1455-1465.

[75] DEMIRIZ, Ayhan; BENNETT, Kristin P.; EMBRECHTS, Mark J. Semi-supervised clustering using genetic algorithms. Artificial neural networks in engineering (ANNIE-99), 1999, p. 809-814.

[76] ONG, Yew-Soon; NAIR, Prasanth B.; LUM, Kai Yew. Max-min surrogate-assisted evolutionary algorithm for robust design. Evolutionary Computation, IEEE Transactions on, 2006, vol. 10, no 4, p. 392-404.

[77] GUTIÉRREZ, José A. García; COTTA, Carlos; LEIVA, Antonio Fernández. Design of emergent and adaptive virtual players in a war RTS game. Foundations on Natural and Artificial Computation. Springer Berlin Heidelberg, 2011. p. 372-382.

[78] FLOREANO, Dario; MONDADA, Francesco. Evolution of homing navigation in a real mobile robot. Systems, Man, and Cybernetics, Part B: Cybernetics, IEEE Transactions on, 1996, vol. 26, no 3, p. 396-407.

[79] JANG, Jyh-Shing Roger; SUN, Chuen-Tsai; MIZUTANI, Eiji. Neuro-fuzzy and soft computing-a computational approach to learning and machine intelligence [Book Review]. Automatic Control, IEEE Transactions on, 1997, vol. 42, no 10, p. 1482-1484.

[80] JANG, Jyh-Shing Roger; SUN, Chuen-Tsai; MIZUTANI, Eiji. Neuro-fuzzy and soft computing-a computational approach to learning and machine intelligence [Book Review]. Automatic Control, IEEE Transactions on, 1997, vol. 42, no 10, p. 1482-1484.

[81] AKBARZADEH-T, M.-R., et al. Soft computing for autonomous robotic systems. Computers & Electrical Engineering, 2000, vol. 26, no 1, p. 5-32

[82] ZADEH, Lotfi A. Some reflections on soft computing, granular computing and their roles in the conception, design and utilization of information/intelligent systems. Soft Computing-A fusion of foundations, methodologies and applications, 1998, vol. 2, no 1, p. 23-25.

[83] DASGUPTA, Dipankar; JI, Zhou; GONZALEZ, Fabio. Artificial immune system (AIS) research in the last five years. En Evolutionary Computation, 2003. CEC'03. The 2003 Congress on. IEEE, 2003. p. 123-130.

[84] ZHANG, Zheng, et al. A greedy algorithm for aligning DNA sequences.Journal of Computational biology, 2000, vol. 7, no 1-2, p. 203-214.

[85] WOLSEY, Laurence A. An analysis of the greedy algorithm for the submodular set covering problem. Combinatorica, 1982, vol. 2, no 4, p. 385-393.

[86] QUINTERO, Luis Vicente Santana. Un Algoritmo Basado en Evolución Diferencial para Resolver Problemas Multiobjetivo. 2004. Tesis Doctoral. Tesis de Maestría CINVESTAV-IPN.

[87] RICE, Kenneth V.; STORN, Rainer M.; LAMPINEN, Jouni A. Differential evolution a practical approach to global optimization. 2005.

[88] SUN, Jianyong; ZHANG, Qingfu; TSANG, Edward PK. DE/EDA: A new evolutionary algorithm for global optimization. Information Sciences, 2005, vol. 169, no 3, p. 249-262.

[89] QIN, A. Kai; SUGANTHAN, Ponnuthurai N. Self-adaptive differential evolution algorithm for numerical optimization. En Evolutionary Computation, 2005. The 2005 IEEE Congress on. IEEE, 2005. p. 1785-1791.

[90] HANSEN, Nikolaus; KERN, Stefan. Evaluating the CMA evolution strategy on multimodal test functions. En Parallel Problem Solving from Nature-PPSN VIII. Springer Berlin Heidelberg, 2004. p. 282-291.

[91] ROS, Raymond; HANSEN, Nikolaus. A simple modification in CMA-ES achieving linear time and space complexity. En Parallel Problem Solving from Nature–PPSN X. Springer Berlin Heidelberg, 2008. p. 296-305.

[92] AUGER, Anne; HANSEN, Nikolaus. A restart CMA evolution strategy with increasing population size. En Evolutionary Computation, 2005. The 2005 IEEE Congress on. IEEE, 2005. p. 1769-1776.

[93] SAVIC, Dragan A.; WALTERS, Godfrey A. Genetic algorithms for least-cost design of water distribution networks. Journal of Water Resources Planning and Management, 1997, vol. 123, no 2, p. 67-77.

[94] BAKIRTZIS, Anastasios G., et al. Optimal power flow by enhanced genetic algorithm. Power Systems, IEEE Transactions on, 2002, vol. 17, no 2, p. 229-236.

[95] DA SILVA, Edson Luiz; GIL, Hugo Alejandro; AREIZA, Jorge Mauricio. Transmission network expansion planning under an improved genetic algorithm. En Power Industry Computer Applications, 1999. PICA'99. Proceedings of the 21st 1999 IEEE International Conference. IEEE, 1999. p. 315-321.

[96] DASGUPTA, Dipankar. Computational Intelligence in Cyber Security. EnComputational Intelligence for Homeland Security and Personal Safety, Proceedings of the 2006 IEEE International Conference on. IEEE, 2006. p. 2-3.

[97] VLAHOPOULOS, N.; HART, C. G. A Multidisciplinary design optimization approach to relating affordability and performance in a conceptual submarine design. Journal of Ship Production and Design, 2010, vol. 26, no 4, p. 273-289.

[98] DE JONG, Kenneth A. Evolutionary computation: a unified approach. Cambridge: MIT press, 2006

[99] SIVAKUMAR, Raghupathy; SINHA, Prasun; BHARGHAVAN, Vaduvur. CEDAR: a core-extraction distributed ad hoc routing algorithm. Selected Areas in Communications, IEEE Journal on, 1999, vol. 17, no 8, p. 1454-1465.

[100] SHEN, Chien-Chung; JAIKAEO, Chaiporn. Ad hoc multicast routing algorithm with swarm intelligence. Mobile Networks and Applications, 2005, vol. 10, no 1-2, p. 47-59.

[101] HERNÁNDEZ-DÍAZ, A. M. Revisión de las teorías de campo de compresiones en hormigón estructural. 2013.

[102] DE JONG, Kenneth A. Evolutionary computation: a unified approach. Cambridge: MIT press, 2006.

[103] EBELING, Werner. Applications of evolutionary strategies. Systems Analysis Modelling Simulation, 1990, vol. 7, no 1, p. 3-16.

[104] COLLINS, Michael P.; MITCHELL, Denis. Prestressed concrete structures. Englewood Cliffs: Prentice Hall, 1991.

[105] HERNÁNDEZ-DÍAZ A M; GARCÍA-ROMÁN, M.D; Gil-Martín L.M.; Hernández-Montes. Why is not always solvable the shear model for reinforced concrete beams proposed by the Compression Field Theories, 2012.

[106] REINECK, K.; Kuchma, D.; Fitik, B., Formelsammlumg für die Datenerhebungs dateil Stahlbetonbalken mit Bügel, 2011.

[107] HERNÁNDEZ-DÍAZ, A.M; GARCÍA-ROMÁN, M.D. Introducing a Degradation Parameter in the Shear Design of Reinforced Concrete Beams, 2013.

[108] COLEY, David A. An introduction to genetic algorithms for scientists and engineers. World Scientific, 1999.

[109] HANSEN, Nikolaus; OSTERMEIER, Andreas. Adapting arbitrary normal mutation distributions in evolution strategies: The covariance matrix adaptation. En Evolutionary Computation, 1996., Proceedings of IEEE International Conference on. IEEE, 1996. p. 312-317.

[110] MÜLLER, Sibylle D.; HANSEN, Nikolaus; KOUMOUTSAKOS, Petros. Increasing the serial and the parallel performance of the CMA-evolution strategy with large populations. En Parallel Problem Solving from Nature—PPSN VII. Springer Berlin Heidelberg, 2002. p. 422-431.

[111] JASTREBSKI, Grahame A.; ARNOLD, Dirk V. Improving evolution strategies through active covariance matrix adaptation. En Evolutionary Computation, 2006. CEC 2006. IEEE Congress on. IEEE, 2006. p. 2814-2821.

[112] Código modelo CEB-FIP 1990 para hormigón estructural. Colegio de Ingenieros de Caminos, Canales y Puertos. Grupo Español del Hormigón y Asociación Técnica Española del Pretensado.

[113] GIL-MARTIN, L.M.; HERNÁNDEZ-MONTES, E. Refinements to Compression Field Theory, with Application to Wall-Type Structures. American Concrete Institute, ACI Special Publication (265 SP), pp. 123-142.

[114] VAN LAARHOVEN, Peter JM; AARTS, Emile HL. Simulated annealing. Springer Netherlands, 1987.

[115] KIRKPATRICK, Scott. Optimization by simulated annealing: Quantitative studies. Journal of statistical physics, 1984, vol. 34, no 5-6, p. 975-986.

[116] AUGER, Anne; HANSEN, Nikolaus. A restart CMA evolution strategy with increasing population size. En Evolutionary Computation, 2005. The 2005 IEEE Congress on. IEEE, 2005. p. 1769-1776.

[117] HOUCK, Christopher R.; JOINES, Jeffrey A.; KAY, Michael G. Comparison of genetic algorithms, random restart and two-opt switching for solving large location-allocation problems. Computers & Operations Research, 1996, vol. 23, no 6, p. 587-596.

[118] CASILLAS, J.; BENÍTEZ, A. D. Optimización Continua Multimodal mediante Evolución de Funciones de Densidad de Probabilidad.

[119] GESTAL POSE, Marcos. Computación evolutiva para el proceso de selección de variables en espacios de búsqueda multimodales. 2009.

[120] BECK K., Extreme Programming Explained: Embrace Change, Addison-Wesley, Reading, Mass., 1999.

[121] HIGHHSMITH, Jim; COCKBURN, Alistair. Agile software development: The business of innovation. Computer, 2001, vol. 34, no 9, p. 120-127.

[122] MARTIN, Robert Cecil. Agile software development: principles, patterns, and practices. Prentice Hall PTR, 2003.

[123] ASTELS, Dave. Test driven development: A practical guide. Prentice Hall Professional Technical Reference, 2003.

[124] JANZEN, David S.; SAIEDIAN, Hossein. Test-driven development: Concepts, taxonomy, future direction. Computer Science and Software Engineering, 2005, p. 33.

[125] MUMMOLO, Giovanni. Measuring uncertainty and criticality in network planning by PERT-path technique. International journal of project Management, 1997, vol. 15, no 6, p. 377-387

[126] MAYLOR, Harvey. Beyond the Gantt chart:: Project management moving on. European Management Journal, 2001, vol. 19, no 1, p. 92-100.

[127] KAROLAK, Dale Walter; KAROLAK, N. Software Engineering Risk Management: A Just-in-Time Approach. IEEE Computer Society Press, 1995.

[128] KELLNER, Marc I. Software process modeling support for management planning and control. En Software Process, 1991. Proceedings. First International Conference on the. IEEE, 1991. p. 8-28.

[129] MARTIN, Robert Cecil. Agile software development: principles, patterns, and practices. Prentice Hall PTR, 2003.

[130] BECK, Kent, et al. Manifesto for agile software development. 2001.

[131] COCKBURN, Alistair. Agile software development. Boston: Addison-Wesley, 2002.

[132] WOLFRAM, Stephen. The mathematica book. Champaign, Illinois: Wolfram Media, 1996

[133] WOLFRAM, Stephen. Mathematica: a system for doing mathematics by computer. Redwood City. CA: Addison–Wesley. 0, 1991, vol. 20, p. 40-60.

[134] STEINDL, Christoph; SCRUMMASTER, Certified. Test-Driven Development. 2005.

[135] MADEYSKI, Lech. Test-Driven Development. Springer, 2010.

# Anexo I.

# Código fuente de los algoritmos implementados.

```
(* ::Package:: *)

(* basado en los fuentes que acompañan al paper Beyer et al. at 2001 y Hans-Georg Beyer de 2007 *)


(* ***********************************************************************

 ENTRADA & SALIDA

*********************************************************************** *)

messages=OpenWrite[path<>"messages.txt"];

inputdata=path<>"DataBase_Kuchma_et_al_nonprestressed_mathematica.xls";

validhyp=path<>"Critical_Outlet_kappa_refined.xls";

initpop=path<>"initial_population.csv";

thetaseeds=path<>"Strut-angle_seeds.xls";

outputxls=path<>"Salida.csv";

outputpng=path<>"Salida.png";

mfit=OpenWrite[path<>"mfitness.txt"];

initpop=path<>"initial_population.csv";


 (* ********************************************************************** *)

PARAMETROS PROBLEMA

** ********************************************************************** *)

epsilon1seeds=3; (* Number of equispaced seeds for epsilon1 to try *)

specimens={"NHW-3b","A50","A75","C50","C75","S1-4","S2-3","S4-6","ET3","P20","T-22","S8 A","H 50/4","H 75/4","H 100/4","RC 30 A1","RC 30 A2","RC 60 B1","MHB 2.5-25","T3","T4","T6","T7","T8","T9","T13","T15","T17","T19","T20","T26","T32","T34","T35","T37","T38"};

penal=10^5;
```

```
(* ************************************************************************* **
Parametros de estrategia
** ************************************************************************* *)
mu = 2;                      (* 2 numero de padres considerados *)
maxlambda = 20;                  (* tamaño de la descendencia *)
minlambda = 12;                   (* tamaño de la descendencia *)
yInit = Table[2, {3}];          (* vector inicial *)
sigmaInit = 1.2;                  (* initial global mutation strength sigma *)
sigmaMin = 10^-8;               (* Criterio de parada sigma < sigmaMin *)
maxgens=300;

(* ************************************************************************* **
 Inicialización de la distribución de arranque
** ************************************************************************* *)
norm = NormalDistribution[0, 1];  (* inicializamos distribucion normal *)
n = Dimensions[yInit][[1]];      (* dimensiones del espacio de busqueda *)
tau = Sqrt[n]; tauC = n^2; tauSigma = Sqrt[n];
Cov = IdentityMatrix[n];         (* matriz covarianzas inicialmente igual a identidad *)
sigma = sigmaInit; s = Table[0, {n}]; sSigma = Table[0, {n}];

(* Inicializacion Memoria *)
Individual = {f[yInit], yInit, yInit, yInit};
ParentPop = Table[Individual, {mu}];
yParent = yInit;   descendencia = {{}, {}, {}, {}};     (* reservamos mem para la pob *)
elite = {{}, {}, {}, {}};    (* reservamos mem para la pob *)

(* Definicion de Funciones *)
ftest[x_] := Module[ {n},  n = Dimensions[x][[1]];
```

```
                Sum[ i*x[[i]]^3, {i, 1, n}] ]

f[R_]:=Module[{i,j,T,fit,Es,EqSystem,suma,\[Alpha],\[Alpha]1,\[Alpha]2,\[Alpha]t,z,bw,Asx1,Asx2,Ast,s,fyx1,fyx2,fyt,fc,\[Epsilon]c,fctm,\[Epsilon]ctm,Acx1,Acx2,Act,Ec,V,\[Sigma]stexp,sol,\[Kappa]limx1,\[Kappa]limx2,\[Kappa]limt,\[Kappa]lim2x1,\[Kappa]lim2x2,\[Kappa]lim2t,\[Kappa]lim1,\[Kappa]lim2,epsilonx,epsilont,\[Epsilon]yx1,\[Epsilon]yx2,\[Epsilon]yt,check,hypx1,hypx2,hypt,emaxx1,emaxx2,emaxt,solx1,solx2,solt,iter,y0,x0,x1,x2,a,b,c,d,H,spec,line,hyp,h,diff,specdiff,seedvector,S,lineseed,Sol},


If [R[[1]]==Null, return penal];If [R[[2]]==Null, return penal]; (* casos imposibles *)

If [R[[3]]==Null, return penal]; (* casos imposibles *)

a=Coef[R[[1]]]; b=Coef[R[[2]]]; c=Coef[R[[3]]]; (*d=Coef[R[[4]]];*)

Print["Polinomio candidato ",a,"/(1+",b,"*\[Epsilon]1^",c,")"];

T=Import[inputdata]; Es=200000; suma=0;fit=0;


(* PARA CADA ESPECIMEN *)

For[spec=1,spec<=Length[specimens],spec++,

        line=0;

        For[i=1,i<=Length[T[[1]]] && line==0, i++,

                If[T[[1,i,1]]==specimens[[spec]],line=i];

        ];

        If[line==0 ,

                Write[messages,"Specimen "<>ToString[specimens[[spec]]]<>" is not in DataBase. "];

          Exit[];

        ];

        \[Alpha]=T[[1,line,7]]; [Alpha]1=T[[1,line,4]]; \[Alpha]2=T[[1,line,5]]; \[Alpha]t=T[[1,line,6]];

        z=T[[1,line,8]]; bw=T[[1,line,9]]; Asx1=T[[1,line,10]]; Asx2=T[[1,line,11]];
        Ast=T[[1,line,12]]; s=T[[1,line,13]]; fyx1=T[[1,line,14]]; fyx2=T[[1,line,15]];

        fyt=T[[1,line,16]]; fc=T[[1,line,17]]; Ec=8500Power[fc+8, (3)^-1]; \[Epsilon]c=T[[1,line,18]];

        fctm=T[[1,line,19]]; \[Epsilon]ctm=fctm/Ec; Acx1=T[[1,line,20]]; Acx2=T[[1,line,21]];

        Act=T[[1,line,22]]; Es=200000; V=T[[1,line,23]]; \[Epsilon]yx1= fyx1/Es;
        \[Epsilon]yx2=fyx2/Es; \[Epsilon]yt=fyt/Es;


        (* ESTE ES EL VALOR EXPERIMENTAL *)
```

```
\[Sigma]stexp=T[[1,line,2]];

(* Limits to 'Degradation Parameter' for each type of reinforcement *)

 (* valores de frontera *)

\[Kappa]limx1=Asx1*fyx1/(\[Alpha]1*Acx1*fctm);

If[\[Alpha]2!=0,\[Kappa]limx2=Asx2*fyx2/(\[Alpha]2*Acx2*fctm)];

\[Kappa]limt=Ast*fyt/(\[Alpha]t*Act*fctm);

\[Kappa]lim1=Min[\[Kappa]limt,If[\[Alpha]2!=0,Min[\[Kappa]limx1,\[Kappa]limx2],\[Kappa]limx
1]];

\[Kappa]lim2x1=\[Kappa]limx1*(4500*\[Epsilon]yx1+(1+1500*\[Epsilon]yx1)^(3/2)-
1)/(6750*\[Epsilon]yx1);

If[\[Alpha]2!=0,\[Kappa]lim2x2=\[Kappa]limx2*(4500*\[Epsilon]yx2+(1+1500*\[Epsilon]yx2)^(3
/2)-1)/(6750*\[Epsilon]yx2)];

\[Kappa]lim2t=\[Kappa]limt*(4500*\[Epsilon]yt+(1+1500*\[Epsilon]yt)^(3/2)-
1)/(6750*\[Epsilon]yt);

\[Kappa]lim2=Min[\[Kappa]lim2t,If[\[Alpha]2!=0,Min[\[Kappa]lim2x1,\[Kappa]lim2x2],\[Kappa]l
im2x1]];

(* Seeds for theta*)

(* son los valores iniciales del parametro theta para el newton-rap *)

S=Import[thetaseeds];

lineseed=0;

For[i=1,i<=Length[S[[1]]] && lineseed==0, i++,

        If[S[[1,i,1]]==specimens[[spec]],

        lineseed=i; seedvector={};

        ];

];

If[lineseed==0 , Write[messages,"Specimen "<>ToString[specimens[[spec]]]<>" has no theta
seeds. Check if 'path' for theta seeds file is right."];

        seedvector={30 Degree,30 Degree,30 Degree,30 Degree,30 Degree};  (* caso default *)

];


(* Valid behaviour hypotheses for reinforcement  *)

(* contiene las hipotesis validas para cada especimen. Al menos una de ellas es cierta  *)
```

```
H=Import[validhyp];

line=0;   For[i=1,i<=Length[H[[1]]] && line==0, i++,

            If[H[[1,i,1]]==specimens[[spec]],line=i];

];

If[line==0 , Write[messages,"Specimen "<>specimens[[spec]]<>" has no valid behaviour hypotheses. Check if 'path' for hypotheses file is right."];

            Exit[];

];

hyp={};

For[h=0,h<=4,h++,

            If[H[[1,line,2+9*h]]!=""  &&  H[[1,line,2+9*h]]!="Null",

                        hyp=Append[hyp,Table[StringTake[H[[1,line,2+9*h]],{k}],{k,1,3}]];

                        If[lineseed!=0,seedvector=Append[seedvector,S[[1,lineseed,2+h]]]];

            ];

];

(*Print["Specimen "<>specimens[[spec]]," Hypothesis ",hyp,"   Semillas ",seedvector];*)

diff={};

For[i=1,i<=Length[hyp],i++,

            j=1; specdiff={};

            While[j<=epsilon1seeds, (*** Search for solutions with several seeds for epsilon1 ***)

            (* Print[" ",hyp[[i]]," - j:",j," \[Epsilon]1seed=",\[Epsilon]ctm+(Min[\[Epsilon]yx1,\[Epsilon]yx2]+\[Epsilon]yt-\[Epsilon]ctm)*(j-1)/(epsilon1seeds-1),"  \[Theta]seed=",seedvector[[i]]]; *)

            Clear[\[Sigma]1,\[Sigma]2,\[Sigma]sx1,\[Sigma]sx2,\[Sigma]st,\[Epsilon]1,\[Epsilon]2,\[Epsilon]x,\[Epsilon]t,\[Theta],f2max,\[Kappa]];

                        (* sustituciones de variables *)

                        \[Sigma]1=\[Alpha]*fctm/(1+Sqrt[500*\[Epsilon]1]);

                        \[Sigma]2=(Tan[\[Theta]]+1/Tan[\[Theta]])*V/(z*bw)-\[Sigma]1;

                        f2max=Min[fc,fc/(0.8+170*\[Epsilon]1)];

                        \[Epsilon]2=\[Epsilon]c*(1-Sqrt[1-\[Sigma]2/f2max]);

                        \[Epsilon]t=(\[Epsilon]2*Tan[\[Theta]]^2+\[Epsilon]1)/(Tan[\[Theta]]^2+1);
```

```
\[Epsilon]x=\[Epsilon]1+\[Epsilon]2-\[Epsilon]t;

\[Kappa]=a/(1+b*(\[Epsilon]1)^c);

If [a/(1+b*1^c)>=1, Return [penal];]; (* Caso imposible *)


(*\[Kappa]=a*(\[Epsilon]1*1000)^3+b*(\[Epsilon]1*1000)^2+c*\[Epsilon]1*1000+d ;*)

\[Sigma]sx1=If[hyp[[i,1]]=="E",Es*\[Epsilon]x,fyx1-[Kappa]*(Acx1/Asx1)*\[Alpha]1*fctm/(1+Sqrt[500*\[Epsilon]x])];

\[Sigma]sx2=If[\[Alpha]2==0,0,If[hyp[[i,2]]=="E",Es*\[Epsilon]x,fyx2-\[Kappa]*(Acx2/Asx2)*\[Alpha]2*fctm/(1+Sqrt[500*\[Epsilon]x])]];

\[Sigma]st=If[hyp[[i,3]]=="E",Es*\[Epsilon]t,fyt-\[Kappa]*(Act/Ast)*\[Alpha]t*fctm/(1+Sqrt[500*\[Epsilon]t])];


(* soluciona el sistema mediante N-R *)

Eq1[\[Epsilon]1_,\[Theta]_]:=Evaluate[Asx1*\[Sigma]sx1+Asx2*\[Sigma]sx2-V/Tan[\[Theta]]+\[Sigma]1*bw*z];

Eq2[\[Epsilon]1_,\[Theta]_]:=Evaluate[\[Sigma]st*Ast-(\[Sigma]2*Sin[\[Theta]]^2-\[Sigma]1*Cos[\[Theta]]^2)*bw*s];

Quiet[Sol=Check[FindRoot[{Eq1[\[Epsilon]1,\[Theta]],Eq2[\[Epsilon]1,\[Theta]]},{\[Epsilon]1,\[Epsilon]ctm+(Min[\[Epsilon]yx1,\[Epsilon]yx2]+\[Epsilon]yt-\[Epsilon]ctm)*(j-1)/(epsilon1seeds-1)},{\[Theta],seedvector[[i]]}, MaxIterations->10000],0]];

(* hay solucion pero esta fuera de rango *)

If[Length[Sol]!=0,

        \[Epsilon]1=\[Epsilon]1/.Sol[[1]];

        (* POLINOMIO INTERPOLADOR para el valor k *)

        \[Kappa]=a/(1+b*(\[Epsilon]1)^c);

        If[\[Epsilon]1<\[Epsilon]ctm || \[Kappa]>\[Kappa]lim2,

                Sol=0; (* Descarto solucion *)

        ];

];

(* hay solucion. eq1 y eq2 se interceptan *)

If[Length[Sol]!=0,
```

```
(* deshacer el cambio de variables *)

        \[Theta]=\[Theta]/.Sol[[2]];

        \[Sigma]1=\[Alpha]*fctm/(1+Sqrt[500*\[Epsilon]1]);

        \[Sigma]2=(Tan[\[Theta]]+1/Tan[\[Theta]])*V/(z*bw)-\[Sigma]1;

        f2max=Min[fc,fc/(0.8+170*\[Epsilon]1)];

        \[Epsilon]2=\[Epsilon]c*(1-Sqrt[1-\[Sigma]2/f2max]);

        \[Epsilon]t=(\[Epsilon]2*Tan[\[Theta]]^2+\[Epsilon]1)/(Tan[\[Theta]]^
        2+1); (* la solucion contradice la hipotesis inicial?. *)

        \[Epsilon]x=\[Epsilon]1+\[Epsilon]2-\[Epsilon]t;(* la solucion
        contradice la hipotesis inicial?. *)



        (* ESTE ES EL VALOR TEORICO *)

        \[Sigma]st=If[hyp[[i,3]]=="E",Es*\[Epsilon]t,fyt-
        \[Kappa]*(Act/Ast)*\[Alpha]t*fctm/(1+Sqrt[500*\[Epsilon]t])];

        (* Print["\[Sigma]st=",\[Sigma]st,", \[Sigma]stexp=",\[Sigma]stexp]; *)

        solx1=FindRoot[Es*(\[Epsilon]yx1-x)==
        \[Kappa]*(Acx1/Asx1)*\[Alpha]1*fctm/(1+Sqrt[500*x]),{x,\[Epsilon]yx
        1}];

        emaxx1=x/.solx1[[1]];

        If[\[Alpha]2!=0,

                solx2=FindRoot[Es*(\[Epsilon]yx2-x)==
                \[Kappa]*(Acx2/Asx2)*\[Alpha]2*fctm/(1+Sqrt[500*x]),{x,\[E
                psilon]yx2}];

                emaxx2=x/.solx2[[1]];

        ];

        solt=FindRoot[Es*(\[Epsilon]yt-x)== \[Kappa]*(Act/Ast)*\[Alpha]t
        *fctm/(1+Sqrt[500*x]),{x,\[Epsilon]yt}];

        emaxt=x/.solt[[1]];

        (* Calculo las diferencias cuadradas *)

        specdiff=If[((\[Epsilon]x<=emaxx1 &&
        hyp[[i,1]]=="E")||(\[Epsilon]x>=emaxx1 && hyp[[i,1]]=="P"))
        &&(\[Alpha]2==0 ||( \[Epsilon]x<=emaxx2 &&
        hyp[[i,2]]=="E")||(\[Epsilon]x>=emaxx2 && hyp[[i,2]]=="P")) &&
        ((\[Epsilon]t<=emaxt && hyp[[i,3]]=="E")||(\[Epsilon]t>=emaxt &&
```

```
                    hyp[[i,3]]=="P")), Append[specdiff,(\[Sigma]st-
                    \[Sigma]stexp)^2],Append[specdiff,penal]];

                    (* Print["Diferencia acumulada: ", specdiff]; *)

                ];

                (* NO hay solucion. eq1 y eq2 no se interceptan *)

                If[Length[Sol]==0,

                    specdiff=Append[specdiff,penal];

                ]; j++;

            ]; (* fin for por cada especimen. *)

            (* Print[specdiff]; *)

            diff=Append[diff,Min[specdiff]];

        ]; (* fin for por cada hipotesis. siguiente hipotesis posible *)

(* Print ["Suma parcial: ",suma," diff^2 de hipotesis: ",diff]; *)

suma=suma+Min[diff];

]; (* fin for por cada especimen *)

suma=suma/(Length[specimens]*10^2); fit=Max[0,1-(suma/1000)]//N;

If [suma<1000,Print ["Suma ",suma, ", Fitness Norm. ", fit]; ,Print ["No se alcanza
solucion."];Return[penal];];

Return[suma];

];

Coef[x_]:=Tan[\[Pi]*x-\[Pi]/2];

Coord[y_]:=ArcTan[y]/\[Pi]+1/2;


(* ********************************************************************** *)

  RUTINA PRINCIPAL

(* ********************************************************************** *)

curr = yInit;

While[ True,

 generacion=generacion+1;
```

```
rSqrtCov=SqrtCov; errorTry[SqrtCov = Transpose[CholeskyDecomposition[Cov]];,SqrtCov=rSqrtCov;]; (* matriz transformacion *)

lambda=Max[minlambda,(maxlambda-generacion)] //N; (* intensidad de la busqueda *)

Print["Generacion ",generacion," Dp=", Dp, ", mu=", mu,", lambda=", lambda];

oldfmed=fmed; fmed=0;   (* generacion de descendencia *)

descendenciaPop = Table[

 (   descendencia[[4]] = Table[ Random[norm], {n}];

     descendencia[[3]] = sigma*(SqrtCov.descendencia[[4]]);

     descendencia[[2]] = yParent + descendencia[[3]];

     descendencia[[1]] = f[descendencia[[2]]]; fmed = (fmed + descendencia[[1]])/2;

     descendencia

 ), {lambda} ];


(* memoria generacional (elitismo) *)

If [Element[generacion,Primes],descendenciaPop = Union[descendenciaPop,Take[Sort[elite], Min[5,Length[elite]-1]]];];  (* elitismo *)

ParentPop = Take[Sort[descendenciaPop], mu]; fmed=fmed//N; (* tomamos los padres *)

tfmean=ParentPop[[1, 1]]; (* mejor fitness *)

If [tfmean<10^7,Print[" f_mean = ", tfmean];];

(*If [generacion>5,Print["Candidato a elite: ", f[ParentPop[[1, 2]]]];];  mejor individuo *)


If [Mod[generacion,10]==0, Print["Best = ", ParentPop[[1]]]; Delete[elite,1]; ];


(* Descomentar solo UNA de las dos líneas siguientes *)

(* If [generacion==1, elite=Import[melite,"CSV"];,If [f[ParentPop[[1, 2]]]!=penal, elite=Union[elite,Take[Sort[descendenciaPop], 1]];]; ]; elitismo *)

If [generacion==1, elite=Take[Sort[descendenciaPop], 1];,If [f[ParentPop[[1, 2]]]!=penal, elite=Union[elite,Take[Sort[descendenciaPop], 1]];]; ]; (* elitismo *)

If [Length[elite]>1,Print["nueva elite = ", elite];];


(* cruce y semilla descendencia *)
```

```
  Recombinant = Sum[ParentPop[[m]], {m, 1, mu}]/mu; (* el cruce *)
  yParent = Recombinant[[2]]; (* el nuevo centro *)
  If [ParentPop[[1, 1]]<penal,curr = yParent;];

  (* autoadaptacion *)
  s = (1-1/tau)*s + Sqrt[mu/tau*(2-1/tau)]*Recombinant[[3]]/sigma;
  Cov = (1-1/tauC)*Cov + Outer[Times, (s/tauC), s];
  Cov = (Cov + Transpose[Cov])/2; (* forzamos la simetria *)
  sSigma = (1-1/tauSigma)*sSigma + Sqrt[mu/tauSigma*(2-1/tauSigma)]*Recombinant[[4]];
  sigma = sigma*Exp[(sSigma.sSigma - n)/(2*n*Sqrt[n])]; (* E6 *)

  (* extincion y reinicio *)
  If [oldfmed==fmed, Print["Reinicio.."]; yParent =  yInit*RandomReal[{0.5,1}]; lambda=maxlambda; tau = Sqrt[n]; tauC = n^2; tauSigma = Sqrt[n]; Cov = IdentityMatrix[n]; sigma = sigmaInit; ];

  (* Criterios de parada *)
  If[ sigma < sigmaMin, Break[] ]
  If[ generacion > maxgens, Break[] ]
](* Fin while, avanzamos a generacion siguiente *)
Export[melite,elite,"CSV"];
Close[messages]; Close[mfit]; Close[melite];
```

Anexo II.

# Autorización para la defensa.

Dr. D. José María Cecilia Canales profesor de la UCAM.

CERTIFICA: que el Trabajo Fin de Grado titulado "ANÁLISIS E IMPLEMENTACIÓN DE ALGORITMOS EVOLUTIVOS PARA LA OPTIMIZACIÓN DE SIMULACIONES EN INGENIERÍA CIVIL." que presenta D. José Alberto García Gutiérrez, para optar al título oficial de Grado en Ingeniería informática mención en Ingeniería del Software, ha sido realizado bajo su dirección.

A su juicio reúne las condiciones necesarias para ser presentado en la Universidad Católica San Antonio de Murcia y ser juzgado por el tribunal correspondiente.

Murcia, a 17 de Junio de 2014